\documentclass[journal]{IEEEtran}
\usepackage{times}
\usepackage{amsmath,amsthm,amssymb,mathrsfs,upgreek,dsfont}
\usepackage{tensor}
\usepackage{graphicx}
\usepackage[font=footnotesize]{caption,subcaption}
\usepackage{overpic}
\graphicspath{{./figs/}}
\usepackage{rotating}
\usepackage{verbatim}
\usepackage{enumitem}
\usepackage{color}
\usepackage{soul}
\usepackage{bm}
\usepackage{float}
\usepackage{todonotes}
\usepackage[ruled,vlined]{algorithm2e}
\usepackage{ifpdf}
\usepackage{cite}
\usepackage{booktabs}
\usepackage{hyperref}
\usepackage{tikz}
\usetikzlibrary{decorations.pathreplacing,calc}
\usepackage[multiple]{footmisc}
\usepackage{xurl}
\usepackage{multirow}
\usepackage{diagbox}

\DeclareCaptionSubType*[Alph]{table}
\DeclareCaptionLabelFormat{mystyle}{Table~\bothIfFirst{#1}{ ̃}#2}
\DeclareMathOperator*{\argmin}{arg\!\min}
\DeclareMathOperator*{\argmax}{arg\!\max}
\newcommand*\diff{\mathop{}\!\mathrm{d}}
\newcommand*\eig{\mathop{}\!\mathrm{eig}}

\newtheorem{thm}{Theorem}
\newtheorem{lemma}{Lemma}
\newtheorem{corol}{Corollary}

\def\code#1{\texttt{#1}}

\begin{document}
%
\title{Adaptive Gaussian Process based Stochastic Trajectory Optimization for Motion Planning}
%
%
%

\author{Feng Yichang,
        Zhang~Haiyun, 
        Wang~Jin, 
        and Lu~Guodong, 
\thanks{Wang Jin, corresponding author,  State Key Laboratory of Fluid Power and Mechatronic Systems, School of Mechanical Engineering, Zhejiang University, Hangzhou, Zhejiang, China.
        {\tt\small dwjcom@zju.edu.cn}}
\thanks{Manuscript received Xxx xx, 20xx; revised Xxx xx, 20xx.}}

%
%

\markboth{Journal of \LaTeX\ Class Files,~Vol.~xx, No.~x, Xxx~xxxx}%
{Shell \MakeLowercase{\textit{et al.}}: Bare Demo of IEEEtran.cls for IEEE Journals}
%



\maketitle

\begin{abstract}

We propose a new formulation of optimal motion planning (OMP) algorithm for robots operating in a hazardous environment, called adaptive Gaussian-process based stochastic trajectory optimization (AGP-STO). It first restarts the accelerated gradient descent with the reestimated Lipschitz constant ($\bm{\mathcal{L}}$-reAGD) to improve the computation efficiency, only requiring 1st-order momentum. However, it still cannot infer a global optimum of the nonconvex problem, informed by the prior information of Gaussian-process (GP) and obstacles. So it then integrates the adaptive stochastic trajectory optimization (ASTO) in the $\bm{\mathcal{L}}$-reestimation process to learn the GP-prior rewarded by the important samples via accelerated moving averaging (AMA). Moreover, we introduce the incremental optimal motion planning (iOMP) to upgrade AGP-STO to iAGP-STO. It interpolates the trajectory incrementally among the previously optimized waypoints to ensure time-continuous safety. Finally, we benchmark iAGP-STO against the numerical (CHOMP, TrajOpt, GPMP) and sampling (STOMP, RRT-Connect) methods and conduct the tuning experiment of key parameters to show how the integration of $\bm{\mathcal{L}}$-reAGD, ASTO, and iOMP elevates computation efficiency and reliability. Moreover, the implementation of iAGP-STO on LBR-iiwa, multi-AGV, and rethink-Baxter demonstrates its application in manipulation, collaboration, and assistance. 

\end{abstract}

%
\IEEEpeerreviewmaketitle

\section{Introduction}
%
%
%
%
\IEEEPARstart{I}ndustrial robot plays a vital role in manufactory benefited from their high flexibility, efficiency, and robustness compared to manual labor in some repetitive works. However, most robotization achieves via manual programming or teaching. It is time-consuming to assign every support waypoint motion, resulting in low flexibility and adaptation facing the space or time-varying of the working condition. Motion planning is crucial for robots to safely execute complex tasks in a harsh workspace (W-space). 

Optimal motion planning (OMP), a.k.a. trajectory optimization, tends to find a series of collision-free waypoints satisfying the task requirements. Some former studies search for an optimal solution via numerical optimization~\cite{Zucker2013CHOMP, Schulman2014SCO, Mukadam2018GPMP} or probabilistic sampling~\cite{Lavalle1998RRT, Kuffner2000RRT-connect, Karaman2011RRT*, Kalakrishnan2011STOMP} in configuration space (C-space) or W-space. 
The numerical method can rapidly approach minima by serial descent steps and adapt to a dynamic environment, while the minima’s feasibility highly depends on the initial point. The sampling-based method can overcome this problem regardless of the high computational cost of high-dimensional searching space with narrow feasible subspace. 
\begin{enumerate}[leftmargin = 0pt, itemindent = 2\parindent, label=(\roman*)]
	\item Though some numerical methods~\cite{Schulman2014SCO,Mukadam2018GPMP} adopt the trust-region or line search methods for solution search, the trust-region adjustment or quasi-Hessian matrix estimation will consume the extra computation resource and finally approach the local minimum. 
	\item The graph searching methods~\cite{Kuffner2000RRT-connect, Karaman2011RRT*,Kavraki1994PRM} construct a probabilistic roadmap in C-space with random vertex sampling and shortest branch connection. However, they require massive computational resources for repeated collision-check and roadmap storage. The trajectory sampling methods~\cite{Kalakrishnan2011STOMP, Zucker2013CHOMP} flatten this issue via resampling a limited number of noisy trajectories regardless of the lower success rate. 
	\item Though some fusion methods~\cite{Zucker2013CHOMP, Berenson2011TSR-RRT,Stouraitis2020OHMP-DcM} integrate the numerical and probabilistic methods for path smoothing and success rate improvement, the limited searching process restrains the environmental information gathering for OMP. 
\end{enumerate}

To solve the above concerns, iAGP-STO makes an incremental probabilistic inference of an optimal trajectory based on the adaptive Gaussian process (AGP), integrating stochastic learning and numerical trajectory optimization (STO).  
\begin{enumerate}[leftmargin = 0pt, itemindent = 2\parindent, label=(\roman*)]
	\item Considering the high convexity requirement~\cite{Schulman2014SCO} and the extra computation resource~\cite{Mukadam2018GPMP}, AGP-STO adopts \ul{accelerated gradient descent (AGD)}~\cite{Ghadimi2016AG-NLP} and \ul{restarts it with the reestimated Lipschitz constant ($\mathcal{L}$-reAGD)}. It also adopts the penalty method~\cite{Nocedal2006NumericalOpt} to gather the GP prior and obstacle information to maximize a posterior. To elevate the efficiency further, we \ul{incrementally optimize the motion planning (iOMP)} by constructing a Bayes tree consisting of factors informed by GP prior, obstacle, and other motion constraints and optimize the sub-trajectories containing significant factors. 
	\item Like other numerical methods, $\mathcal{L}$-reAGD easily encounters the local minima when a dense obstruction is in the W-space. Therefore, we judge whether a minimum approaches during $\mathcal{L}$-reestimation. When it approaches, \ul{adaptive stochastic trajectory optimization (ASTO)} will resample noise trajectories, gain the reward via expectation-maximization and \ul{learn the GP-policy via accelerated moving averaging (AMA)} to infer a trajectory located in a convex subspace. 
	\item Unlike CHOMP~\cite{Zucker2013CHOMP} alternating between resampling and leapfrog, AGP-STO \ul{inserts the ASTO in the  $\mathcal{L}$-reestimation} depending on whether a minimum exists and then \ul{infers an optimal policy learned from ASTO} via $\mathcal{L}$-reAGD until convergence.  Besides, it utilizes $\mathcal{L}$-reAGD to restrain the searching space of ASTO, unlike CBiRRT~\cite{Berenson2011TSR-RRT}, reducing the C-space dimension by finding the null-space of constraints.
\end{enumerate}
\begin{figure}[htbp]
		\begin{subfigure}[b]{0.24\textwidth}		
			\centering
			\includegraphics[width=1\linewidth]{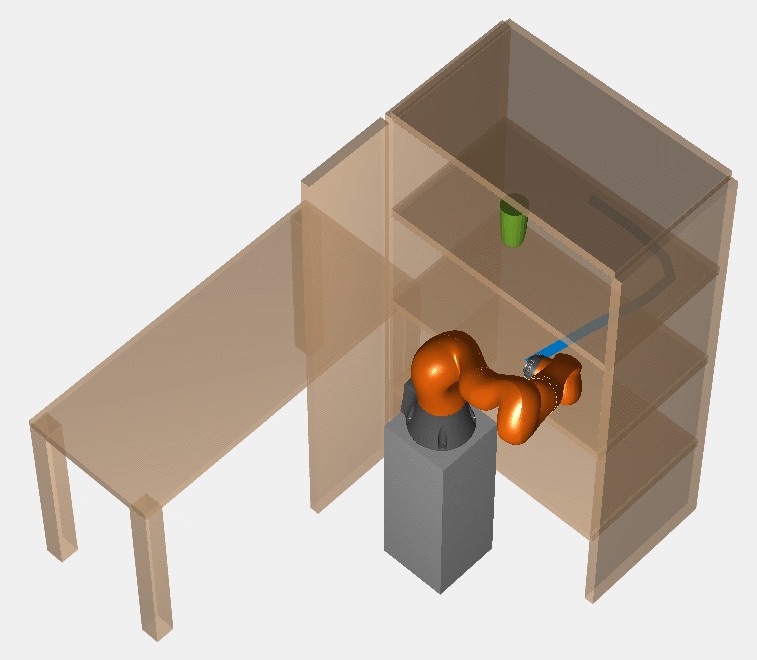}
			\caption{LBR-iiwa manipulates in a shelf.}
		\end{subfigure}	
		\hfill
		\begin{subfigure}[b]{0.24\textwidth}		
			\centering
			\includegraphics[width=1\linewidth]{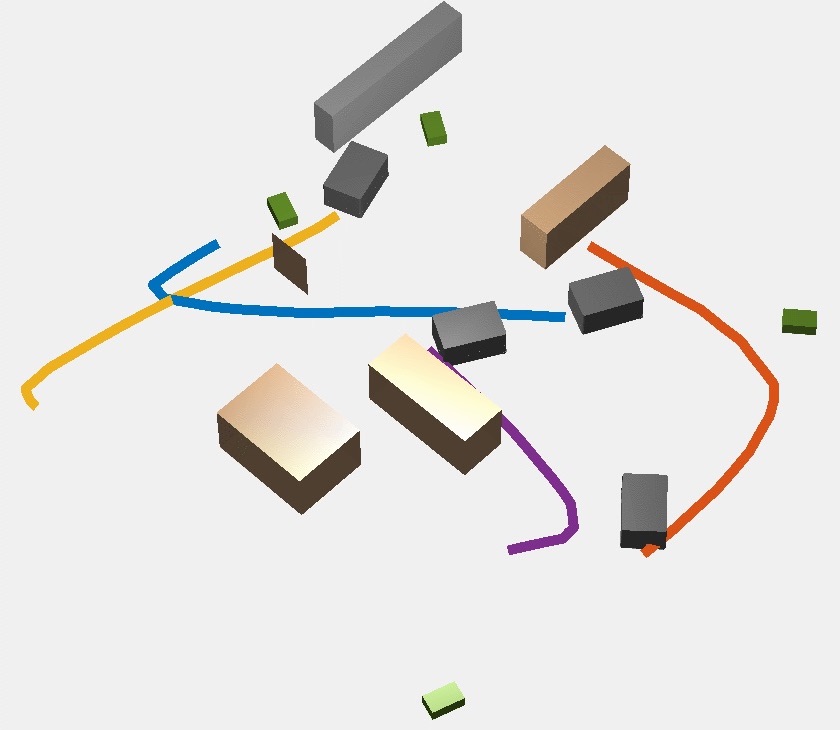}
			\caption{4 AGVs collaborate in a warehouse. }
		\end{subfigure}
		\\[4pt]
		\begin{subfigure}[b]{0.24\textwidth}
			\centering
			\includegraphics[width=1\linewidth]{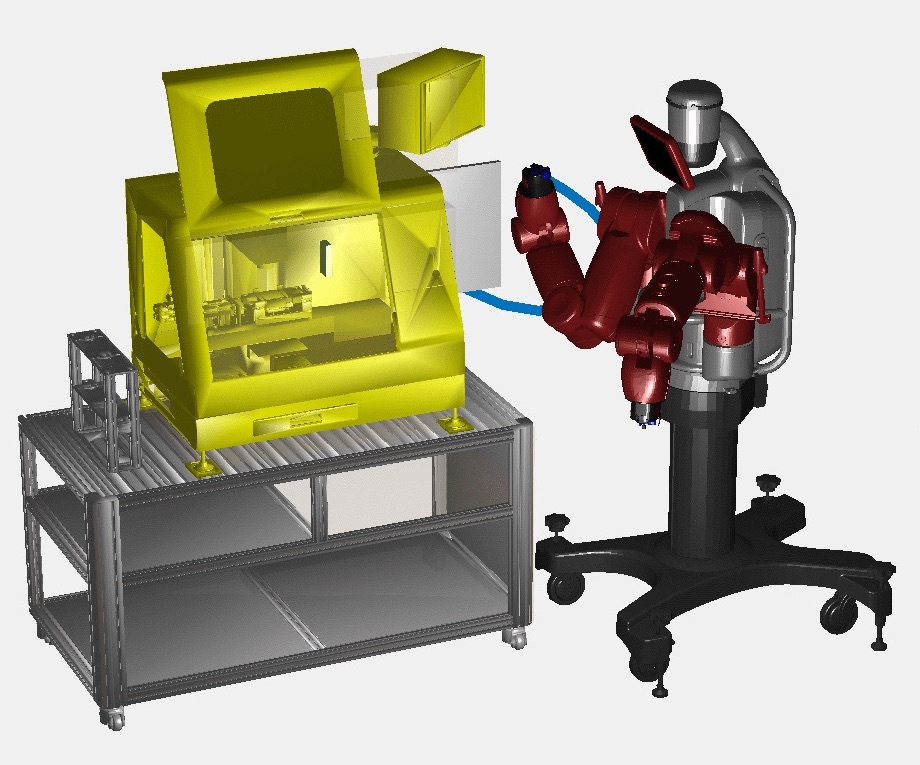}
			\caption{rethink-Baxter assists at a machine. }
		\end{subfigure}
		\hfill
		\begin{subfigure}[b]{0.24\textwidth}
			\centering
			\includegraphics[width=1\linewidth]{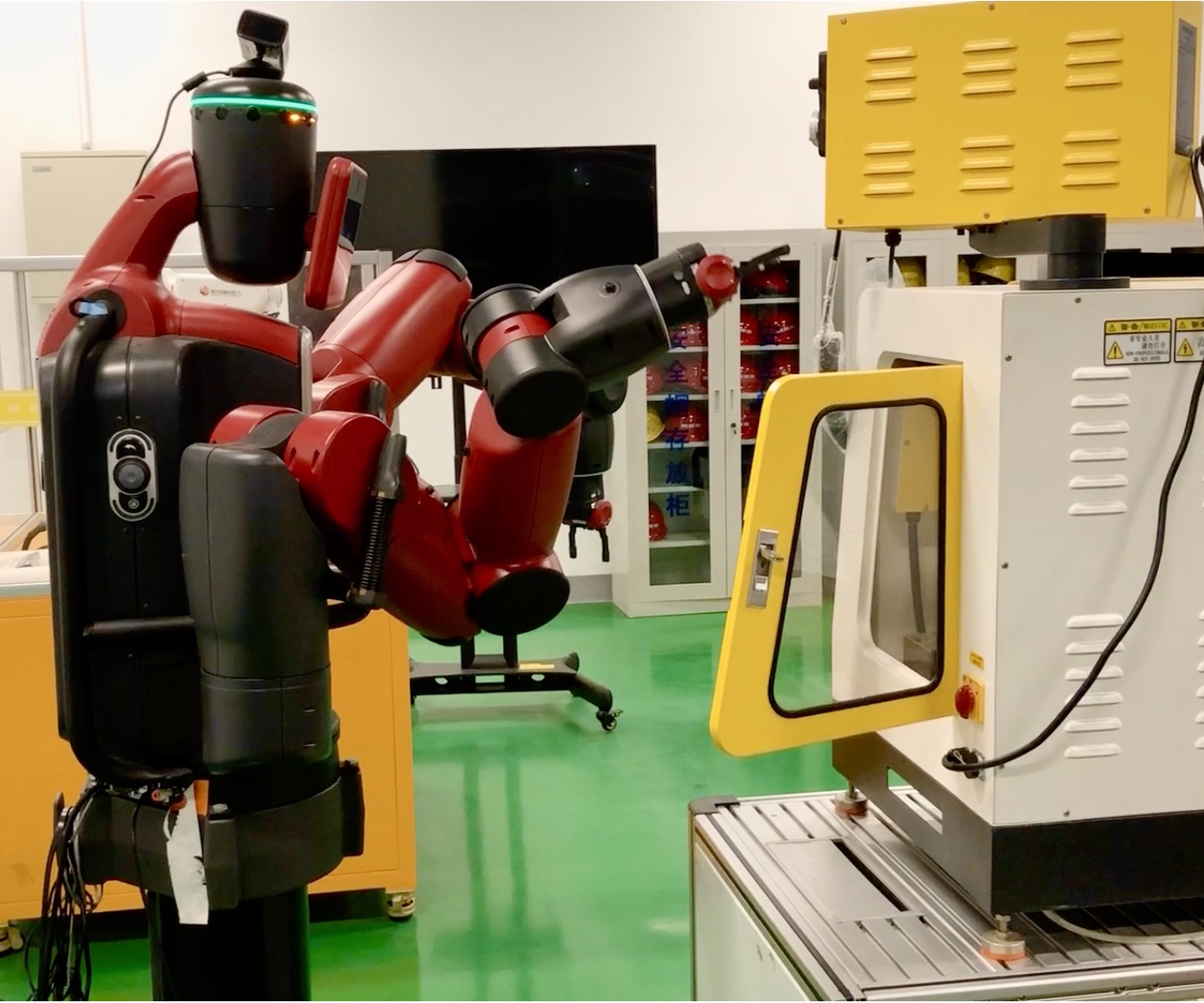}
			\caption{The real experiment of rethink-Baxter. }
		\end{subfigure}
\protect\caption{ Different problems of optimal motion planning (OMP).
\label{fig:OMP}}
\end{figure}
\begin{table*}[hbt]
\caption{Terms of Abbreviation}
\label{tab:abbr}
\centering
\scalebox{1}{
\begin{tabular}{cc|ccccccccc}
\toprule 
abbr. & meaning & abbr. & meaning  \\ 
\midrule
AGP-STO & adaptive gaussian process based stochastic trajectory optimization & 
iAGP-STO & incremental AGP-STO \\
AGD \textbar $\mathcal{L}$-reAGD & accelerated gradient descent \textbar $\mathcal{L}$ipschitz restarted  AGD & 
HMC & Hamiltonian Monte-carlo \\
ASTO & adaptive stochastic trajectory optimization & 
AMA & accelerated moving averaging \\
GP \textbar $\mathcal{GP}$ & Gaussian process \textbar GP distribution or model & 
CCB & collision-check box or ball \\
BT \textbar BTI & Bayes tree \textbar BT inference & 
MAP & maximize a posterior \\
W-\textbar C-\textbar T-space & work\textbar configuration\textbar trajectory -space & 
OMP & optimal motion planning\\
\bottomrule
\end{tabular}}
\end{table*} 

Finally, we have successfully implemented iAGP-STO on LBR-iiwa (Figure~\ref{fig:OMP}) in 25 tasks with different initial and goal states. All tasks are categorized into three classes (A, B, C) to benchmark iAGP-STO against $\mathcal{L}$-reAGD, AGP-STO, and other algorithms, such as CHOMP~\cite{Zucker2013CHOMP}, GPMP~\cite{Mukadam2018GPMP}, TrajOpt~\cite{Schulman2013SCO}, STOMP~\cite{Kalakrishnan2011STOMP} and RRT-Connect~\cite{Kuffner2000RRT-connect}. The benchmark results show that AGP-STO has an eminently higher success rate, especially in the Class C task. However, it takes more time for iAGP-STO to generate the trajectory than the numerical methods. Besides, the results of tuning and cross-validation experiments show how $\mathcal{L}$-reAGD, ASTO, and iOMP unite to achieve highly reliable and efficient planning. Furthermore, we conduct the simulation experiments on a multi-AGV system and the real experiments on a single arm of rethink-Baxter, as shown in Figure~\ref{fig:OMP}. Table~\ref{tab:abbr} provides the terms of abbreviation.  

\section{Related Works}\label{sec:relateWorks}


\subsection{Formulation of Optimal Motion Planning}

Since OMP optimizes the objective functional meeting the motion constraints to gain an optimal trajectory, it is essential to formulate the constrained optimization problem (COP). Recent studies can be divided into whole trajectory optimization and local waypoints adaptation.  

\subsubsection{Whole Trajectory Optimization}
The former studies of whole trajectory optimization, such as CHOMP~\cite{Zucker2013CHOMP}, ITOMP~\cite{Park2012ITOMP}, TrajOpt (a.k.a. SCO)~\cite{Schulman2013SCO, Schulman2014SCO} and GPMP~\cite{Mukadam2018GPMP}, all adopt an objective functional consisting of smooth term and obstacle constrained terms. In the smooth term, CHOMP, ITOMP, and TrajOpt adopt the kinetic energy form, while GPMP uses the GP model. As for the obstacle term, all of them except for TrajOpt employ the body-obstacle kinetic energy~\cite{Quinlan1993CollisionCheck, Quinlan1995CollisionCheck, Zucker2013CHOMP}. TrajOpt uses a fixed convex combination of constraint gradients in adjacent time to ensure the time-continuous safety, while GPMP uses an up-sampling method. 
Moreover, the penalty factor~\cite{Nocedal2006NumericalOpt} is adopted to optimize the objective under collision-free constraints. Most of the recent studies~\cite{Zucker2013CHOMP, Mukadam2018GPMP} transform the SDF~\cite{Osher2003SDF} into a non-negative obstacle cost function and use constant trade-off parameter, while TrajOpt uses the $\ell_1$-penalty method~\cite{Nocedal2006NumericalOpt}. 

\subsubsection{Local Waypoints Adaptation}
As for the local waypoints adaptation method, ITOMP~\cite{Park2012ITOMP} interleaves of planning and execution using the prediction of environment change for real-time replanning. The algorithmic safety measure~\cite{Liu2016SafetyMeasure} uses the model-predictive control and proposes a time continuously differentiable obstacle cost function for real-time collision avoidance. GPMP2~\cite{Dong2016GPMP2} adopts iSAM2~\cite{Kaess2012iSAM2} and applies BTI~\cite{Toussaint2010BTI} for sparse nonlinear incremental optimization. Note that a wide range of recent studies has applied Bayes inference in simultaneous localization and mapping~\cite{Paull2014SLAM-ACP, Paull2018SLAM-CACP}, pattern recognition~\cite{Kapoor2010GPOC, Rodner2017GPIPR} and OMP~\cite{Dong2016GPMP2, Mukadam2018GPMP, Huang2017Graph-GPMP} for the convenience to apply quasi-Newton method. 



\subsection{Numerical Optimization}

As the main branch of OMP study, numerical optimization searches for the optimum on the manifold of COP in the trajectory space. Its studies could be divided into Quasi-Newton, gradient descent, and stochastic optimization method according to the estimation method of descent step. 

\subsubsection{Quasi-Newton Method}
The Newton method~\cite{Nocedal2006NumericalOpt} solves linear systems to achieve a superlinear convergence. 
However, it consumes numerous resources for a Hessian matrix and its inverse. 
So the quasi-Newton method always approximates Hessian via LM~\cite{Levenberg1944LM, Sugihara2011LMforIK, Dong2016GPMP2}, BFGS~\cite{Badreddine2014BFGS, Curtis2017BFGS} et al. to gain the same rate. GPMP2~\cite{Mukadam2018GPMP} recently adopts the LM method to solve the MAP problem given the prior information. Besides, machining learning and OMP~\cite{Lo2002BFGSforHMP} apply the limited-memory BFGS (L-BFGS)~\cite{Liu1989L-BFGS} and damped BFGS (dBFGS)~\cite{Zhu1997BFGS} algorithm for high dimensional optimization and optimal control, respectively. Because L-BFGS and dBFGS only require 1st order information for superlinear convergence.

\subsubsection{Gradient Descent Method}
In contrast to the Newton or quasi-Newton method, gradient descent needs to estimate the step size rather than Hessian. It contains two main methods: the line search~\cite{Al-baali1985LineSearch} and the trust-region method~\cite{Conn2000TrustRegion, Schulman2015TrustRegion}. In practice, CHOMP~\cite{Zucker2013CHOMP} and GPMP~\cite{Mukadam2018GPMP} both use a fixed step, while TrajOpt~\cite{Schulman2013SCO} adopts the trust region method. 

\subsubsection{Stochastic Optimization}
Unlike the exact numerical optimization method, the stochastic gradient descent (SGD)~\cite{Bottou1999onlineSGD} methods, such as Adam~\cite{Kingma2014Adam}, RMSProp~\cite{Hinton2012RMSProp}, and AdaGrad~\cite{Duchi2011Ada}, always program the object composed of a sum of weakly coupling sub-functions and generates the sub-gradients via the random selections of sub-functions. Our recent work~\cite{Feng2021iSAGO} proposes iSAGO based on SGD to overcome the local minima brought by the body-obstacle stuck case. 
During SGD, they all adopt Nesterov's accelerated gradient descent (AGD)~\cite{Nesterov1983AG, Ghadimi2016AG-NLP}, which forms a projection from a current step into its near future. Moreover, the AGD steps, containing the stochastic gradients, optimize machine learning progress~\cite{Ji2009AGDforML, Hu2009AGDforML, Jin2017AGDforML}, whose convergence rate correlates with the order of the Lipschitz constant.

\subsection{Probabilistic Sampling}
The sampling-based method samples waypoints or trajectories dynamically in the configuration space (C-space) or working space (W-space). 

\subsubsection{Roadmap Construction}
Though A*~\cite{Hart1968A*} introduces a heuristic function for the forward cost estimation, its dynamic programming takes numerous computation resources for OMP. Rapid random tree (RRT)~\cite{Lavalle1998RRT} treats OMP as a tree search problem to solve this concern. As an upgrade of RRT, RRT* \cite{Karaman2011RRT*} appends the original construction of RRT with a rewiring process to shorten the path. RRT-connect~\cite{Kuffner2000RRT-connect} introduces the bidirectional searching method to improve efficiency. The law of the above algorithms is abstracted by the random nearest-neighbor-type graph~\cite{Wade2007kNN-RRG}. LQR-RRT*~\cite{Perez2012LQR-RRT*} combines RRT* with the linear–quadratic regulator for optimal manipulation control. Unlike RRT methods, PRMs~\cite{Kavraki1994PRM, Kavraki1996PRMs, Kavraki1998PRMs} randomly generate and connect numerous collision-free configurations to construct a probabilistic road map. However, the generation of several collision-free waypoints always consumes more time with less optimality than RRT and RRT*. FFRob~\cite{Garrett2018FFRob} introduces extended action specifications to make a general representation of planning problems and adopts RRT-connect and sPRM to construct a conditional reachability graph.

\subsubsection{Trajectory Sampling}

Despite the above sampling methods, STOMP~\cite{Kalakrishnan2011STOMP} adopts the objective functional form of \cite{Zucker2013CHOMP} and optimizes the trajectory via noise trajectory resampling obeying Gaussian distribution renewed by the selected samplings. \cite{Petrovic2019HGP-STO} introduces the GP model into the stochastic process of STOMP to improve the searching efficiency. Recently, SMTO~\cite{Osa2020SMTO} upgraded STOMP’s sampling model by applying Monte-Carlo optimization for multi-solution solving and using the numerical method for policy refinement. 

\section{Background}\label{sec:backGround}

\subsection{Accelerated Gradient Descent}\label{sec:AGD}

Nesterov's AGD~\cite{Nesterov1983AG} makes faster nonlinear programming than gradient descent~\cite{Al-baali1985LineSearch} yet takes more steps than the quasi-Newton method~\cite{Armand2017QuadLagrange}, benefiting from Hessian estimation. 
This none Hessian method performs efficiently~\cite{Priyank2021NAGD-COP} in some high-dimensional, even nonconvex problems. Section~\ref{sec:L-reAGD} adopts AGD for GP inference of trajectory given the obstacle information, while Section~\ref{sec:ASTO} adopts it for adaptive GP-learning during the trajectory resampling. For generalization, this section abstracts them as 
\begin{equation}\label{eq:opt_prob}
	\arg \min_{\bm{x}\in \mathds{R}^{n}} \Psi(\bm{x})
\end{equation} 
and assumes 
\begin{equation} \label{eq:Lips_continue}
	\|\nabla \Psi(\bm{y})-\nabla \Psi(\bm{x})\| \leq \mathcal{L}_{\Psi}\|\bm{y}-\bm{x}\| \quad \forall \bm{x}, \bm{y} \in \bm{\mathcal{X}}, 
\end{equation}
where $\bm{\mathcal{X}} \subseteq \mathds{R}^{n}$ is a locally convex subspace with Lipschitz ($\mathcal{L}_{\Psi}$) continuous gradient $\nabla \Psi$. 

\begin{lemma}\label{lemma:Gamma}
Let $\{{\alpha}_{k}\}$ be a series of forgetting coefficients, each of whom selects a point between $\bm{x}^{ag}$ and $\bm{x}$, and ${\gamma}_{k}$ be a sequence satisfying
\begin{equation}\nonumber
	\gamma_{k} \leq\left(1-\alpha_{k}\right) \gamma_{k-1}+\eta_{k}, \quad k=1,2, \ldots, N
\end{equation}
where we define another sequence
\begin{equation}\label{eq:Gamma}
	\Gamma_{k}:=\left\{\begin{array}{ll}1, & k=1 \\ 
	\left(1-\alpha_{k}\right) \Gamma_{k-1}, & k \geq 2. \end{array}\right.
\end{equation}
Then we have
\begin{equation}\nonumber 
	\gamma_{k} \leq \Gamma_{k} \sum_{i=1}^{k}\left(\eta_{i} / \Gamma_{i}\right), \forall k > 1. 
\end{equation}
\end{lemma}

Unlike the traditional methods~\cite{Al-baali1985LineSearch, Shi2005LineSearch} performing steepest descent in each step, 
Ghadimi's AGD~\cite{Ghadimi2016AG-NLP} introduces $\lambda_{k}$ to release the primary steps from steepest descent and $\beta_{k}$ to ensure a robust convergence in some nonconvex cases. 

\begin{algorithm}[]
\caption{AGD}\label{alg:AGD}
\DontPrintSemicolon
\LinesNumbered
\SetKwInOut{Input}{Input}
\SetKwInOut{Output}{Output}
\SetKwFunction{Union}{Union}
\SetKw{Or}{or}

\Input{initial $\bm{x}_{0}$, objective function $\Psi$, Lipschitz constant $\mathcal{L}_{\Psi}$, update rule $\{\alpha_k, \beta_k, \lambda_k \}$, and termination condition \code{AccBreak()}.}
\textbf{Initialize: } $\left\{\bm{x}^{md}_{0},\bm{x}^{ag}_{0}\right\} \leftarrow \bm{x}_{0}$; \;
\For {$\textit{AgdIter}: k = 1\dots N_{ag}$}{
 	$\bm{x}_{k}^{md}=\left(1-\alpha_{k}\right) \bm{x}_{k-1}^{ag}+\alpha_{k} \bm{x}_{k-1}$; \;
	$\textit{agdBrk} \leftarrow $ \code{AccBreak($\{\bm{x}_k, \bm{x}^{ag}_k, \bm{x}^{md}_k\}$,$\Psi$,$\mathcal{L}_{\Psi}$)};\label{alg:AccBrk} \;
	\lIf{\textit{agdBrk} \Or $k \geq N_{ag}$ \Or converge to $\mathcal{F}$tol,$\theta$tol}{
		\Return $\bm{x}_{k}^{md}$. 
	}
 	Compute $\nabla{\Psi}(\boldsymbol{x}_{k}^{md})$ and update $\{\alpha_{k}$, $\beta_{k}$, $\lambda_{k}\}$;  \;
 	$\bm{x}_{k}=\bm{x}_{k-1}-\lambda_{k} \nabla{\Psi}(\boldsymbol{x}_{k}^{md})$; \;
 	$\bm{x}_{k}^{ag}= \bm{x}_{k}^{md}-\beta_{k} \nabla{\Psi}(\boldsymbol{x}_{k}^{md})$; \;
} 
\end{algorithm}
\begin{thm}\label{thm:conditionAGparas}
Following Lemma~\ref{lemma:Gamma} and Algorithm~\ref{alg:AGD},
we get condition $\{\alpha_{k}, \beta_{k}, \lambda_{k}\}$ needs to satisfy according to \cite{Ghadimi2016AG-NLP}:
\begin{equation}\label{eq:conditionAGparas}
C_{k}=1-\mathcal{L}_{\Psi} \lambda_{k}-\frac{\mathcal{L}_{\Psi}\left(\lambda_{k}-\beta_{k}\right)^{2}}{2 \alpha_{k} \Gamma_{k} \lambda_{k}}\left(\sum_{\tau=k}^{N} \Gamma_{\tau}\right)>0,
\end{equation}
to approach the convergence of $\| \nabla\Psi \|$: 
\begin{equation} \label{eq:converge01}
	\min _{k=1, \ldots, N_{ag}}\left\| \nabla \Psi \left(\bm{x}_{k}^{md}\right)\right\|^{2} \leq \frac{\Psi\left(\bm{x}_{0}\right)-\Psi^{*}}{\sum_{k=1}^{N_{ag}} \lambda_{k} C_{k}}.
\end{equation}
\end{thm}

Unlike Ghadimi's method using a  constant $\mathcal{L}_{\Psi}$, our AGD method inserts \code{AccBreak()} in line~\ref{alg:AccBrk} of Algorithm~\ref{alg:AGD} to adaptively terminate the AGD process. Moreover, Section~\ref{sec:L-reAGD} will detail the technologies of \code{AccBreak()} to restart AGD with the reestimated $\mathcal{L}_{\mathcal{F}}$, while Section~\ref{sec:ASTO} will reformulate AGD to resample the trajectory adaptively. 


\section{Problem formulation}\label{sec:formulation}

As we treat the robotic motion as a GP, we first review the Gauss-Markov process, and the derivation of the GP prior. Then we illustrate the posterior of the GP prior based on the GP theorem in Section~\ref{sec:gpModel}. Finally, Section~\ref{sec:costs} reviews how to construct an objective functional informed by the GP and obstacle factors. 

\subsection{A Gauss-Markov model}\label{sec:ltvsde}

Given the initial state, goal state, and environment information, the motion planning problem is a probabilistic inference. Its prior probabilistic model is formed by a linear time-varying stochastic differential equation (LTV-SDE):
\begin{equation}\label{eq:LTV-SDE}
	\dot{{{\theta}}}(t) = \mathbf{A}(t){\theta}(t) + \mathbf{u}(t) + \mathbf{F}(t)\bm{w}(t),
\end{equation}
where $\mathbf{A}$ and $\mathbf{F}$ are the matrices of linear system and stochastic process correspondingly, and $\mathbf{u}(t)$ is a system input. The white noise process $\bm{w}(t)$ itself is a GP with zero expectation: 
\begin{equation}\label{eq:GaussianNoise}
	\bm{w}(t) \sim \mathcal{GP} (\boldsymbol{0}, \mathbf{Q}_c(t) \bm{\delta} (t - t^{\prime})), 
\end{equation}
where $\mathbf{Q}_c(t)$ is an isotropic power-spectral density matrix. Then we could gain the mean ${{\widetilde{\mu}}} (t)$ and covariance ${{\widetilde{\mathcal{K}}}}(t, t^{\prime})$~\footnote{This paper uses the bold font, such as $\bm{\theta}_{t,t'}, \bm{\mu}_{t,t'}, \bm{\mathcal{K}}_{t,t'}$, for a point in the trajectory space of the time interval $(t,t')$, while uses the normal font, such as ${\theta}_{t}, {\mu}_{t}, {\mathcal{K}}_{t}$, for a point in configuration space at time $t$. }\footnote{This paper uses $\|\bm{p}\|_\mathbf{S}$ to measure the Mahalanobis distance of $\bm{p}$ with the covariance $\mathbf{S}$: $\sqrt{\bm{p}^\mathrm{T}\mathbf{S}^{-1}\bm{p}}$. } of GP generated by LTV-SDE~\eqref{eq:LTV-SDE} according to~\cite{Mukadam2018GPMP}:   
%
\begin{gather}
	\label{eq:prior_x}
	\widetilde{\mu} (t) = \boldsymbol{\Phi}(t, t_0) \mu_0 + \int_{t_0}^{\widetilde{t}} \boldsymbol{\Phi}(t, s) \mathbf{u}(s) \diff s,\\
	\label{eq:prior_P}
	\widetilde{\mathcal{K}}(t, t^{\prime}) =  \|\boldsymbol{\Phi}(t, t_0)\|_{\mathcal{K}_0^\text{-1}}^2  +  \int_{\text{t}_0}^{\widetilde{t}} \|\bm{\Phi}(t, s) \mathbf{F}(s)\|_{\mathbf{Q}_{c}^{\text{-}1}(s)}^2 \diff s, 
\end{gather}
where $\widetilde{t} = \min(t,t^{\prime})$, ${{\mu}}_0$ and ${{\mathcal{K}}}_0$ are the mean and covariance of the initial state, and $\boldsymbol{\Phi} (t, s)$ is the state transition matrix. 

\subsection{The GP model for motion planning}\label{sec:gpModel}

After gaining a GP prior via \eqref{eq:prior_x} and \eqref{eq:prior_P}, we now construct the observation system of the state
\begin{equation}
	\bm{\theta}' = \mathbf{C}\bm{\theta} + \bm\varepsilon_{\bm{t}'},~  \bm\varepsilon_{\bm{t}'} \backsim \mathcal{N}(\mathbf{0}, \bm{\mathcal{K}}'), 
\end{equation}
where the state 
\begin{equation} \nonumber
\bm{\theta} = [\theta_{t}^{\mathrm{T}}, \theta_{t+1}^{\mathrm{T}}, \dots, \theta_{t'}^{\mathrm{T}}]^{\mathrm{T}} \backsim \mathcal{N}(\widetilde{\bm{\mu}}, \widetilde{\bm{\mathcal{K}}}),
\end{equation}
 $\bm{t}' \subseteq \bm{t} = \{ t, t+1, \dots, t' \}$, and $\mathbf{C}$ denotes the corresponding observation matrix of $\bm{t}'$. Unlike GPMP~\cite{Mukadam2018GPMP} arbitrarily setting $\{\theta_t, \theta_{t'}\} = \{\theta_0, \theta_g\}$ and $\mathbf{C} = [\mathbf{0}_{t_0}, \dots, \mathbf{0}_{t_{N-1}}, \mathbf{I}]$ with $\bm{t}' = t_N$, we alternate the setting of $\mathbf{C}$ and $\{t,t'\}$ according to the prerequisites, especially in the BTI case (Section~\ref{sec:iOMP}). Based on that, we define the joint distribution as
\begin{equation}\nonumber
\hspace{-1mm}
	\left(\begin{array}{l}
		\bm{\theta}\\ \bm{\theta}'
	\end{array}\right)
	\backsim 
	\mathcal{N}\left(\left[\begin{array}{c}
		\widetilde{\bm\mu} \\ 
		\mathbf{C} \widetilde{\bm\mu} 
	\end{array}\right],
	\left[\begin{array}{cc}
		\widetilde{\bm{\mathcal{K}}} & (\mathbf{C}\widetilde{\bm{\mathcal{K}}})^{\mathrm{T}} \\ 
		\mathbf{C}\widetilde{\bm{\mathcal{K}}} & \mathbf{C}\widetilde{\bm{\mathcal{K}}}\mathbf{C}^{\mathrm{T}} +\bm{\mathcal{\mathcal{K}}}'  
	\end{array}\right]\right). 
\end{equation}
%

So we formulate the posterior of $\bm{\theta}$ given the observation $\bm{\theta}'$ 
and simplify its expression $\bm{\mu}_{\bm{\theta}|\bm{\theta}'}$, $\bm{\mathcal{K}}_{\bm{\theta}|\bm{\theta}'}$ by $\bm{\mu}$, $\bm{\mathcal{K}}$: 
\begin{gather}
	\label{eq:prior_mean}
	\bm{\mu} = \widetilde{\bm{\mu}} + (\mathbf{C}\widetilde{\bm{\mathcal{K}}})^{\mathrm{T}} (\mathbf{C}\widetilde{\bm{\mathcal{K}}}\mathbf{C}^{\mathrm{T}} + \bm{\mathcal{K}}' )^{-1} (\bm{\theta}' - \mathbf{C} \widetilde{\bm{\mu}}), \\
	\label{eq:prior_k}
	\bm{\mathcal{K}} = \widetilde{\bm{\mathcal{K}}} - \widetilde{\bm{\mathcal{K}}}^{\mathrm{T}} \mathbf{C}^{\mathrm{T}} (\mathbf{C}\widetilde{\bm{\mathcal{K}}}\mathbf{C}^{\mathrm{T}} + \bm{\mathcal{K}}' )^{-1} \mathbf{C}\widetilde{\bm{\mathcal{K}}}. 
\end{gather}

\if
This particular construction of the prior leads to a Gauss-Markov model that generates a GP with an exactly sparse tridiagonal precision matrix (inverse kernel) that can be factored as
\begin{equation} \label{eq:Kinv}
	\bm{\mathcal{K}}^{-1} = \mathbf B^\top \mathbf Q^{-1} \mathbf B
\end{equation}
with
\begin{equation}\label{eq:Ainv}
\small
	\mathbf B = \left[ \begin{matrix}
	\mathbf{I} & \bm 0 & \dots & \bm 0 & \bm 0\\
	-\mathbf\Phi_{t+1,t} & \mathbf{I} & \dots & \bm 0 & \bm 0\\
	\bm 0 & -\mathbf\Phi_{t+2,t} & \ddots & \vdots & \vdots \\
	\vdots & \vdots & \ddots & \mathbf{I} & \bm 0 \\
	\bm 0 & \bm 0 & \dots &-\mathbf\Phi_{t',t'-1} & \mathbf{I} \\
	\bm 0 & \bm 0 & \dots & \bm 0 & \mathbf{I} \end{matrix} \right],
\end{equation}
which has a band diagonal structure and $\mathbf Q^{-1}$ is block diagonal such that
\begin{align}
	\mathbf Q^{-1} &= \operatorname{diag}({\mathcal{K}}_t^{-1}, \mathbf Q_{t,t+1}^{-1}, \dots , \mathbf Q_{t'-1,t'}^{-1}, {\mathcal{K}}_{t'}^{-1}), \label{eq:Q}\\
	\mathbf Q_{a,b} &= \int_{t_a}^{t_b} \mathbf \Phi(b,s) \mathbf F(s) \mathbf{Q}_c \mathbf F(s)^\top \mathbf \Phi(b,s)^\top \diff s. \label{eq:Qab}
\end{align}
This sparse structure is useful for incremental trajectory interpolation (Section \ref{sec:ITI}) and efficient optimization (Section~\ref{sec:reTraj-BTI}).
\fi

\subsection{Cost functionals}\label{sec:costs}
This paper adopts the objective functional proposed in~\cite{Zucker2013CHOMP} for generating an optimal trajectory with low energy cost without collision.  In this way, the problem for collision-free optimal motion planning can be formularized as 
\begin{equation}\label{eq:optimization}
	\begin{array}{cl}
		{\operatorname{minimize}} & {\mathcal{F}_{gp}(\bm{\theta})} \\ 
		{\text { subject to }} & {\mathcal{G}_{i}(\bm{\theta}) \leq 0, \quad i=1, \ldots, m_{\textit{ieq}}} \\ 
		{} & {\mathcal{H}_{i}(\bm{\theta})=0, \quad i=1, \ldots, m_{\textit{eq}}, }
	\end{array}
\end{equation}
where $\mathcal{F}_{gp}$ is an objective functional for trajectory smoothing, $\mathcal{G}_i$ is an inequality constraint for collision avoidance, $\mathcal{H}_i$ is an equality constraints for task requirements. 

Since AGP-STO mainly focuses on trajectory optimization whose initial value always dissatisfies $\mathcal{G}$, we design the objective functional as
\begin{equation} \label{eq:chomp_cost}
	\mathcal{F}(\bm{\theta}) = \mathcal{F}_{\textit{obs}}(\bm{\theta}) + \mathcal{F}_{gp}(\bm{\theta}), 
\end{equation}
where the GP prior $\mathcal{GP}(\bm{\mu},\bm{\mathcal{K}})$ forms the smooth part 
\begin{equation} \label{eq:gp_cost}
	\mathcal{F}_{gp}(\bm{\theta}) = \frac{1}{2}\| \bm{\theta} - \bm\mu \|^2_{\bm{\mathcal{K}}}, 
\end{equation}
and the obstacle part $\mathcal{F}_{\textit{obs}}$~\cite{Zucker2013CHOMP, Mukadam2018GPMP} computes the integration of the body-point kinetic costs in W-space:
\begin{equation}\label{eq:obs_cost}
	\mathcal{F}_\textit{obs}(\bm{\theta}) = \int_{t_0}^{t_{N}} \varrho(t) \int_\mathcal{B} c\left[\bm{x}(u,t)\right] \| \dot{\bm{x}}(u,t) \| \diff u \diff t, 
\end{equation}
where factor $\varrho(t)$ penalizes the collision-free constraints $\mathcal{G}_{i}$, $c(\cdot):\mathbb{R}^3 \to \mathbb{R}$ gathers the obstacle information of the set of points $\mathcal{B} \subset \mathbb{R}^3$ on the robot body in W-space, and $x$ maps from C-space to W-space via the forward kinematics. 

Now some readers might query about the relation between the functional $\mathcal{F}(\bm{\theta})$ and the probabilistic inference of $\bm{\theta}$. Here we define a posterior $p(\bm{\theta} | \bm{\theta}', \mathcal{G})$ informed by the obstacle  $\mathcal{G}$ and the GP prior formulated in Section~\ref{sec:gpModel}: 
\begin{equation}\nonumber
	p(\bm{\theta} | \bm{\theta}', \mathcal{G}) = \frac{1}{\mathcal{P}} \exp\left\{-\frac{1}{2}\|\boldsymbol{\theta}-\boldsymbol{\mu}\|_{\bm{\mathcal{K}}}^{2} - \mathcal{F}_{\textit{obs}}(\bm{\theta})\right\}, 
\end{equation}
where $\mathcal{P} = \int_{\bm{\theta}\in\bm{\Uptheta}} e^{-\mathcal{F}(\bm{\theta})} \diff \bm{\theta}$~\footnote{$\bm{\Uptheta}$ denotes a feasible set of $\bm{\theta}$ under the motion constraint~\eqref{eq:motionConstraintCost}. }, and $\varrho(t)$ adjusted by the penalty method (Section~\ref{sec:AGP-STO}) gathers the GP prior and obstacle information. Then we could gain an optimum 
\begin{equation}\label{eq:MAP}
	\bm{\theta}^{*} = \argmax_{\bm{\theta}\in\bm{\Uptheta}} p(\bm{\theta} | \bm{\theta}', \mathcal{G}) 
\end{equation}
via MAP and obtain $\mathcal{F}(\bm{\theta})$ via $-\log p(\bm{\theta} | \bm{\theta}', \mathcal{G})$. 

\if
In practice, the continuous cost functional can be approximated by the discretization of support way points of the trajectory i.e. $\mathcal{F}_{obs}[\bm{\theta}(t)] =\sum_{i=1}^{N} \mathcal{F}_{obs}[{\theta_i}]$. In order to improve the efficiency and feasibility of  ASOMP, it is important to decompose the discretized objective functioncal
 \begin{align}\label{eq:local_cost}
	\mathcal{F}[\bm{\theta}_{t,t'}] & = \lambda \mathcal{F}_{gp}[\bm{\theta}_{t,t'}] + \mathcal{F}_{obs}[\bm{\theta}_{t,t'}] \nonumber \\
	& = \frac{\lambda}{2} \|\bm{\theta}_{t,t'} - \bm{\mu}_{t,t'}\|_{\bm{\mathcal{K}}_{t, t'}} \nonumber \\
	& +\sum_{\tau\in(t,t')}\sum_{j \in \mathcal{B}}c(x_{\tau,j})\|\dot{x}_\tau\|, 
\end{align}
where $\bm{\mathcal{K}}_{t,t'}$ is the localized covariance matrix gained by Eq.~\eqref{eq:Kinv}, $\bm{\theta}_{t,t'} = [{\theta}_t^{\text{T}},\dots,{\theta}_{t'}^{\text{T}}]^{\text{T}}$  and  $\bm{\mu}_{t,t'} = [{\mu}_t^{\text{T}},\dots,{\mu}_{t'}^{\text{T}} ]^{\text{T}}$ are the current and expected discretized local trajectory from $t$ to $t'$ correspondingly.

The problem described above aims to find a time-discrete trajectory in low energy cost meeting all inequality and equality constraints. However, it is hard to guarantee the collision-free constraint of the interval waypoints when all the support waypoints are collision-free. In this way, we introduce a interval obstacle functional $\mathcal{F}_\textit{iObs}$ ensuring the continuous-time safety
\begin{equation}\label{eq:intv_obs}	
	\mathcal{F}_\textit{iObs}[{\theta}_{i}] = \mathcal{F}_{obs}\left[{\theta}_{i}\right]+\lambda_{\textit{intv}} \mathcal{F}_{obs}\left[{\theta}_{\textit{intv}}(\alpha_{m})\right], 
\end{equation}
where ${\theta}_\textit{intv}(\alpha): \mathbb{R}\mapsto\mathbb{R}^{N_{\textit{q}_\textit{dim}}}$ is designed to generate the interval waypoint between $t_{i-1}$ and $t_{i}$ as shown in Eq.~\eqref{eq:internWaypoint}, $\lambda_\textit{intv}$ is another trade-off parameter between support waypoints in $\{t_{i-1}, t_{i}\}$, and interval parameter $\alpha_{m}$ generates a interval waypoint $\bm{\theta}_\textit{intv}(\alpha_{m})$ with maximum obstacle cost: 
\begin{equation}\label{eq:alphaMax}
	\alpha_{m}=\underset{\alpha}{\arg \max~ } \mathcal{F}_{obs}\left[{\theta}_{\textit{intv}}(\alpha)\right], \alpha \in(0,1), 
\end{equation}
where
\begin{equation}\label{eq:internWaypoint}
	{\theta}_{\textit{intv}} \backsim \mathcal{GP}({\mu}_{t_i,\tau}, {\mathcal{K}}_{t_i,\tau}) \text{~with~} \tau = (1-\alpha)t_{i-1} + \alpha t_{i}.
\end{equation}
The ${\mu}_{t_i,\tau}$ and ${\mathcal{K}}_{t_i,\tau}$ above can be calculated via Eq.~\eqref{eq:prior_mean} and Eq.~\eqref{eq:prior_k}. And in the actual implementation, we select a series of $\alpha$ uniformly distributed between $0$ and $1$, whose details are described in Section~\ref{sec:setup}.
\fi

\section{Adaptive Gaussian Process based\\ Stochastic Trajectory Optimization}\label{sec:AGP-STO}

\begin{figure}[hbtp]
\begin{centering}
	\includegraphics[width=1\linewidth]{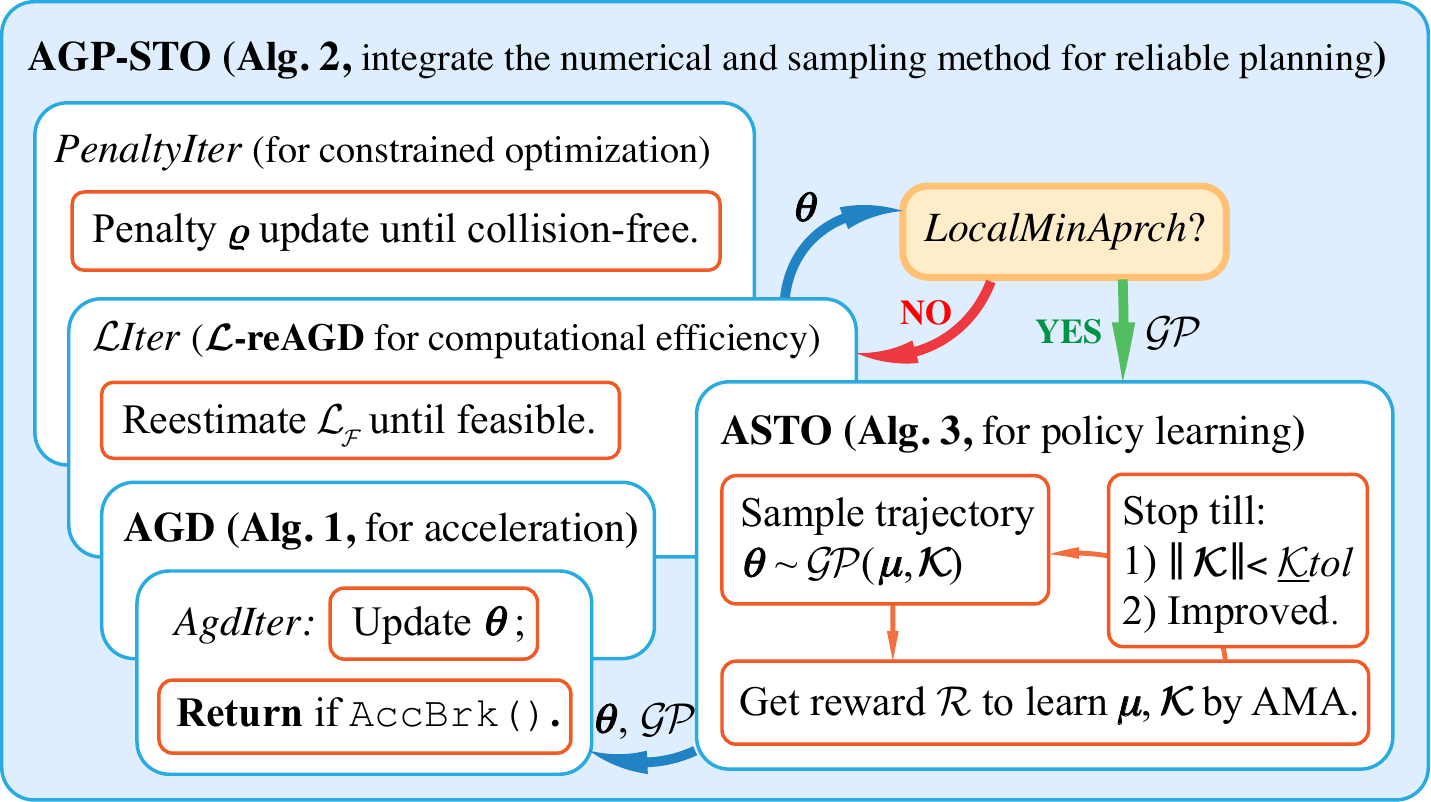}
	\caption{A block diagram illustrates how AGP-STO (Algorithm~\ref{alg:AGP-STO}) plans an optimal motion in a cluttered W-space and how the numerical ($\mathcal{L}$-reAGD, Section~\ref{sec:L-reAGD}) and sampling (ASTO, Algorithm~\ref{alg:ASTO}) methods are integrated to solve the nonconvex optimization problem. 
	\label{fig:AGP-STO_diag}} 
\end{centering}
\end{figure}

Trajectory optimization primarily finds an optimal solution via finding a Hamiltonian minimum. This paper utilizes the time discretization method~\cite{Mukadam2018GPMP} to represent the time-continuous trajectory $\bm{\theta}$ by $[{\theta}_0^{\mathrm{T}},\dots,{\theta}_N^{\mathrm{T}}]^{\mathrm{T}}$. In this way, our method, AGP-STO, in Figure~\ref{fig:AGP-STO_diag}, maximizes the posterior $p(\bm{\theta} | \bm{\theta}', \mathcal{G})$~\eqref{eq:MAP} for optimization. To solve the above MAP problem, this section first presents the intuition of our methodology, then illustrates the scheme of AGP-STO integrating the numerical and sampling method. 
%

\subsection{Intuition}\label{sec:AGP-STO_intuition}

\begin{figure}[htbp]
	\begin{centering}
		{\includegraphics[width=1\columnwidth]{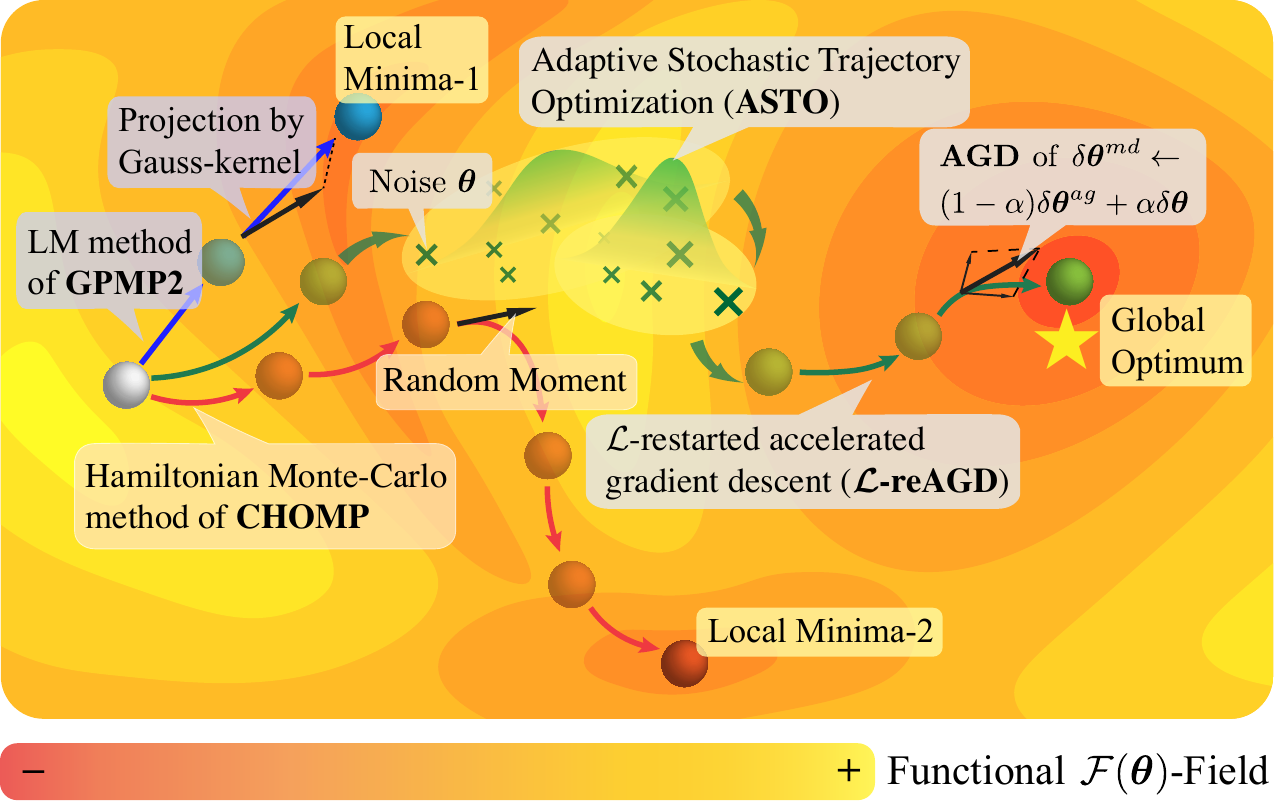}}
		\par\end{centering}
	\protect\caption{
		An example of a super-ball, rolling in a trajectory-space, illustrates how AGP-STO (Algorithm~\ref{alg:AGP-STO}), CHOMP~\cite{Zucker2013CHOMP}, and GPMP2~\cite{Mukadam2018GPMP} search for an optimal solution via momentum descent. The green trace shows how AGP-STO rolls a green ball into the global optimum by the mixed momentum from $\mathcal{L}$-reAGD (Section~\ref{sec:L-reAGD}) and  ASTO (Algorithm~\ref{alg:ASTO}). The blue trace shows how GPMP2 rolls a blue ball into a local minimum by a determinant speed with an acceleration (Gauss-kernel). The red trace shows the resampled speed of CHOMP rolling a red ball into a local minimum. 
		\label{fig:AGP-STO}}
\end{figure}

CHOMP~\cite{Zucker2013CHOMP} gives an example of a super-ball rolling on a landscape until exhausting the kinetic energy to illustrate how CHOMP utilizes the HMC method to approach the optimum. This section adopts the example and provides the intuition behind AGP-STO in Figure~\ref{fig:AGP-STO_diag}~\&~\ref{fig:AGP-STO}. In our mind, two main concerns exist in CHOMP: (i) how to resample \ul{a stochastic momentum} to drive a super-ball towards the optimum with high probability in a trajectory space; (ii) how to define \ul{an energy field} whose total value contains the potential and kinetic energy to roll the ball towards the minima efficiently. 

GPMP2~\cite{Mukadam2018GPMP} sends the blue ball (Figure~\ref{fig:AGP-STO}) off a hill downwards the nearest valley (Local Minima-1) with a few steps, calculated via the LM method utilizing the Gauss kernel to transform the gradient. In comparison, CHOMP accumulates the kinetic via the rejection sampling and moves following the differential equation derived from a Hamiltonian 
\begin{equation}\label{eq:Hamiltonian_1}
	\mathfrak{H}(\bm{\theta})  = \mathfrak{K}(\bm{\theta}) + \mathcal{F}(\bm{\theta}). 
\end{equation}
It contains a kinetic part $\mathfrak{K} = \frac{\mathcal{L}_\mathcal{F}\|\dot{\bm{\theta}}\|^2}{2}$~\footnote{$\mathcal{L}_\mathcal{F}$ is the Lipchitz constant of the objective functional $\mathcal{F}.$}\footnote{$\dot{\bm{\theta}}$ denotes the trajectory variation rate during the optimization rather than its higher derivation with respect to time $t$ of trajectory.} and a potential part $\mathcal{F}$ to overcome a wavy terrain nearby the Local Minima-1 in Figure~\ref{fig:AGP-STO}. However, CHOMP still easily rolls the red ball downwards the Local Minima-2. 

So why cannot we learn the momenta associated with its kinetic energy $\mathfrak{K}$ directly from its potential energy $\mathcal{F}$ when facing the nonconvex cases? ASTO (Algorithm~\ref{alg:ASTO}) actuates the green ball of Figure~\ref{fig:AGP-STO} by the expected value of the sampled momenta obeying the Gauss distribution in proportion to $\exp(-\zeta \mathfrak{K})$ with a flatten factor $\zeta$. Moreover, it renews the Gauss model associated with $\mathfrak{K}$ to resample the momenta performing better in ball motivation and then actuates the green ball repeatedly until the expected momentum becomes zero or the ball rolls into the convex subspace. After that, $\mathcal{L}$-reAGD which generalizes the leapfrog method stimulates the ball motion until $\{\mathfrak{K},\|\bar{\nabla}\mathcal{F}\|\} \rightarrow 0$. Unlike CHOMP reserving $\mathfrak{K}$ from a constant prior distribution, our method adaptively restarts the AGD with zero $\mathfrak{K}$ and a reestimated inertia $\mathcal{L}_{\mathcal{F}}$ to improve the movement efficiency. Considering $\mathcal{G}$~\eqref{eq:MAP}, AGP-STO adopts the penalty method when the robot confronts the collision risk to adjust the potential field $\mathcal{F}$ with $\varrho$~\eqref{eq:obs_cost}.   

In conclusion, AGP-STO (i) utilizes the stochastic momenta, reserving the kinetic energy, by ASTO (Section~\ref{sec:ASTO}) to overcome the highly wavy landscape (i.e., the high non-convexity) around the global optimum; (ii) restarts the AGD movement (Section~\ref{sec:L-reAGD}) driven by $\mathcal{L}_{\mathcal{F}}$ to improve the descent rate; (iii) integrates the penalty and Lagrangian augmentation method (Section~\ref{sec:AGP-STO_scheme}), adjusting the potential field determined by $\mathcal{F}$, to meet the inequality constraints. 

\subsection{AGP-STO for Maximum A Posterior} \label{sec:AGP-STO_scheme}

The above demonstrates an example of how a super-ball, rolls towards the lowest valley of the potential field determined by $\mathcal{F}$ to give an interpretation of how AGP-STO plans an optimal motion. Someone may ask about the alternative perspection of the MAP of $p(\bm{\theta}|\bm{\theta}',\mathcal{G})$. Section~\ref{sec:costs} gives the relation between the objective functional $\mathcal{F}$ and $\bm{\theta}$'s posterior informed by the GP prior and other constraints $\mathcal{G}$. It indicates $p(\bm{\theta}|\bm{\theta}', \mathcal{G})$ attaches the maximum when the super ball is trapped in a local minimum. Besides, the kinetic energy $\mathfrak{K}$ denotes a prior probability distribution of the optimization step. So that ASTO and $\mathcal{L}$-reAGD can infer the posterior via adaptive resampling and the restarted AGD. 

AGP-STO (Algorithm~\ref{alg:AGP-STO}) adopts a frame with the nested loops for the MAP of $p(\bm{\theta}|\bm{\theta}', \mathcal{G})$. The outer loop (\textit{PenIter}) adjusts the penalty factor $\varrho$ for the collision-free trajectory inference. Since we use the discretization form of~\eqref{eq:chomp_cost}: 
\begin{equation}\label{eq:discrt_chomp}
	\mathcal{F}(\bm{\theta}) =  \frac{1}{2}\| \bm{\theta} - \bm\mu \|_{\bm{\mathcal{K}}} + \sum_{t = t_0}^{t_g} \sum_{\mathcal{B}_i \in \bm{\mathcal{B}} } \varrho(t) c[\bm{x}_i(t)] \| \dot{\bm{x}}_i(t) \| 
\end{equation}
for the constrained optimization of discretizated trajectory $\bm{\theta} = [\theta^{\mathrm{T}}_0,\dots,\theta^{\mathrm{T}}_{N}]^\mathrm{T}$, we increase $\bm\varrho = [\varrho_0,\dots,\varrho_N]^\mathrm{T}$ corresponding to $\bm{\theta}$ until the trajectory satifies the constraints $\mathcal{G}\text{tol}$. 
In the second loop (\textit{$\mathcal{L}$Iter}, Section~\ref{sec:L-reStart}), we restart the AGD process with the reestimated $\mathcal{L}_\mathcal{F}$ ($\mathcal{L}$-reAGD) rather than the size adjustment of the trust region to solve the semi-convex problem when the improvement exceeds the threshold. 
Furthermore, AGP-STO employs ASTO (Section~\ref{sec:ASTO}, Figure~\ref{fig:ASTO}) in \textit{$\mathcal{L}$Iter} when the MAP approaches a local minimum. 

\begin{algorithm}[]
	\caption{AGP-STO}\label{alg:AGP-STO}
	\DontPrintSemicolon
	\LinesNumbered
	\SetKwInOut{Input}{Input}
	\SetKwFunction{ASTO}{ASTO}
	\SetKwFunction{MinAprch}{MinAprch}
	\SetKwFunction{AGD}{AGD}
	\SetKwFunction{AccBreak}{AccBreak}
	\SetKw{Or}{or}
	\Input {initial trajectory $\bm{\theta}_0$, prior $\mathcal{GP}(\bm{\mu}_0, \bm{\mathcal{K}}_{0})$, factor $\bm{\varrho}$, and objective functional $\mathcal{F}[\boldsymbol{\theta}(t)]$.}
	\textbf{Initialize:} 
	$\bm{\theta} \leftarrow \bm{\theta}_0$, initial Lipschitz constant $\mathcal{L}_{\mathcal{F}}$, update rules $\{\alpha, \beta, \gamma\}$ of \code{AGD}, penalty scaling factor $\kappa$, convergence thresholds $\mathcal{F}\text{tol}, \theta\text{tol}$; \;
	\For {$ \textit{PenIter}:  i = 1 \dots N_{\varrho} $}{
		\For {$\textit{$\mathcal{L}$Iter}: j = 1 \dots N_{\mathcal{L}}$}{
			Reestimate $\mathcal{L}_{\mathcal{F}}$ according to Section~\ref{sec:L-reStart}; \;
			\If {\MinAprch{}\label{alg:AGP-STO:MinAprch}}{
				$\{\bm{\theta}, \mathcal{GP}(\bm{\mu},\bm{\mathcal{K}})\} \leftarrow \ASTO{$\mathcal{F}$,$\mathcal{GP}(\bm{\theta},\bm{\mathcal{K}})$} $ via \textbf{Algorithm~\ref{alg:ASTO}}; \;
				Update $\mathcal{F}$ by~\eqref{eq:discrt_chomp} and reinitialize $\mathcal{L}_{\mathcal{F}}$; \; 
			}
			$\bm{\theta} \leftarrow \AGD{$\bm{\theta}$,$\mathcal{F}$,$\mathcal{L}_{\mathcal{F}}$,$\{\alpha, \beta, \gamma\}$,\AccBreak{}}$ via \textbf{Algorithm~\ref{alg:AGD}}; \;
			\lIf {converge to $\mathcal{F}\text{tol}$ \Or $\theta\text{tol}$}{
				\textbf{goto} \ref{alg:constraint}; 
			}
		}
		\lIf{$\mathcal{F}_{obs}(\bm{\theta}) < \mathcal{G}\text{tol}$\label{alg:constraint}}{
			\Return $\left\{\bm{\theta}, \mathcal{GP}(\bm{\mu},\bm{\mathcal{K}}), \bm\varrho \right\}$. }
		\lElse{
			$\bm\varrho \leftarrow \kappa \cdot \bm\varrho$; 
		}
	}
	\Return $\left\{\bm{\theta}, \mathcal{GP}(\bm{\mu},\bm{\mathcal{K}}), \bm\varrho \right\}$. \;
\end{algorithm}
%

\section{$\mathcal{L}$-reAGD for Trajectory Optimization}\label{sec:L-reAGD}

This Section utilizes AGD (Section~\ref{sec:AGD}) for MAP. It generalizes and upgrades HMC adopted by CHOMP to achieve high efficiency and robustness compared to GPMP2. Section~\ref{sec:AGD-TO} details how to implement AGD and gives Corollary~\ref{corol:AGD1} about AGD-step estimation. Considering the variation of the planning scenery during the collision avoidance, Section~\ref{sec:L-reStart} proposes an $\mathcal{L}$-restart method for AGD-based OMP and provides its physical interpretation. Finally, we propose a method to determine whether the local minimum approaches, combining the collision avoidance progress.  

\subsection{Accelerated Gradient Descent for Trajectory Optimization}\label{sec:AGD-TO}

The OMP problem~\eqref{eq:optimization} contains a class of composite problems~\eqref{eq:chomp_cost}, where the $\mathcal{GP}$ functional $\mathcal{F}_{{gp}}$ is strongly convex while the obstacle functional $\mathcal{F}_\textit{obs}$ is always nonconvex (Figure~\ref{fig:localmin}). So we implement Algorithm~\ref{alg:AGD} by $\mathcal{F} \rightarrow \Psi$ for MAP and assume that $\mathcal{F}$ is semi-convex and continuously differentiable in a subspace
\begin{equation}
\label{eq:semi-convexHull}
\small
	\mathcal{C} = \left\{\bm{\theta},\bm{\theta}' \left| \left| \mathcal{F} + \langle\bar\nabla\mathcal{F},\bm{\theta}'-\bm{\theta}\rangle - \mathcal{F}' \right| \leq \frac{\mathcal{L}_\mathcal{F}}{2}\|\bm{\theta}'-\bm{\theta} \|^{2}\right. \right\}
\end{equation}
 where $\mathcal{F}$, $\mathcal{F}'$ and $\bar\nabla\mathcal{F}$ simplify $\mathcal{F}(\bm{\theta})$, $\mathcal{F}(\bm{\theta}')$ and $\bar\nabla\mathcal{F}(\bm{\theta})$ respectively, and $\langle \cdot, \cdot \rangle$ denotes the inner production. The reason we call it semi-convex is that it does not have to meet the strongly convex condition: 
\begin{equation}
\label{eq:convexHull}
	\mathcal{F}\left(\alpha\bm{\theta} + (1-\alpha)\bm{\theta}' \right) 
	\leq \alpha \mathcal{F} + (1-\alpha)\mathcal{F}', \alpha \in [0,1].  
\end{equation}
%

The AGD-based MAP utilizes the historical momenta and releases the strict descent of the steepest descent to avoid some infeasible solutions~\cite{Ghadimi2016AG-NLP} in some nonconvex cases. The benchmark results of Table~\ref{table:results_A} shows the solvability and efficiency of $\mathcal{L}$-reAGD in some nonconvex cases, comparing to CHOMP and GPMP. Right now, we detail how \code{AGD($\bm{\theta}$,$\mathcal{F} $)} makes a trajectory MAP. 


\begin{corol}\label{corol:AGD1}
According to Theorem~\ref{thm:conditionAGparas} and the definition of $\mathcal{L}_{\mathcal{F}}$-convex hull~\eqref{eq:semi-convexHull}, we design $\alpha_{k}$ and $\beta_{k}$ as
\begin{equation} \nonumber 
	\alpha_{k}=\frac{2}{k+1}, \quad 
	\beta_{k}=\frac{1}{\vartheta_{1} \mathcal{L}_{\mathcal{F}}}.
\end{equation}
\if 
So the constraint of inequality $C_{k}$~\eqref{eq:conditionAGparas} need to satisfy is
\begin{equation}\label{ieq:conditionAGparas1}
	C_{k} \geq 1-\mathcal{L}_{\mathcal{F}}\left(\lambda_{k}+\frac{\mu_{1}\left(\lambda_{k}-\beta_{k}\right)^{2}}{2 \Gamma_{k} \lambda_{k}}\frac{k+1}{k}\right)
\end{equation}
under
\begin{equation}
	\Gamma_{k}=\frac{2}{k(k+1)},\quad \sum_{\tau=k}^{N} \Gamma_{\tau} \leq \frac{2}{k}. 
\end{equation}
\fi 
Following the constraint of inequality $C_{k} > 0$~\eqref{eq:conditionAGparas} and the definition of 
\begin{equation} \nonumber 
	\lambda_{k} \in\left[\beta_{k},\left(1+{\vartheta_{2}}{\alpha_{k}}\right) \beta_{k}\right], \forall k \geq 1, 
\end{equation}
we can get
\begin{equation} \nonumber 
	\left(\vartheta_{2} + \frac{\alpha_k}{2} \right)^2 \leq \vartheta_{1} - 1 + \frac{\alpha_k^2}{4}, \quad \vartheta_{2} \geq 0 
\end{equation}
by the substitution of $\lambda_{k}$ under the inequality reduction~\footnote{Appendix~\ref{appdx:paraAGD} elaborates the detailed information}. 
Then due to~\eqref{eq:converge01}, we can get the convergence of 
\begin{equation} \label{eq:corol1_converge}
	\min_{k=1\dots N_{ag}}\|\bar{\nabla}\mathcal{F}^{md}_{k}\|^2 
	\leq \frac{\vartheta_{1}^{2} \mathcal{L}_{\mathcal{F}}\left(\mathcal{F}_{0} - \mathcal{F}_{*}\right)}{N_{ag}(-\vartheta_{2}^{2}-\vartheta_{2}+\vartheta_{1}-1)}. 
\end{equation}
\end{corol}

Now, let us see how the super ball in Figure~\ref{fig:AGP-STO}, representing a trajectory, performs an accelerated descent stimulated by Algorithm~\ref{alg:AGD} and analytically compares its performance with CHOMP. According to Algorithm~\ref{alg:AGD}, 
we can get 
\begin{equation}\nonumber
\small
\begin{aligned}
	& \frac{k}{2}\left(\bm{\theta}_{k+1}^{md} - 2\bm{\theta}_{k}^{md} + \bm{\theta}_{k-1}^{md}\right) + \left(\bm{\theta}_{k+1}^{md} - \frac{\bm{\theta}_{k}^{md}+\bm{\theta}_{k-1}^{md}}{2}\right) \\
	& = -\frac{k}{2}\left( \beta_k\bar{\nabla}\mathcal{F}^{md}_{k} - \beta_{k-1}\bar{\nabla}\mathcal{F}^{md}_{k-1}\right) - \lambda_k  \bar{\nabla}\mathcal{F}^{md}_{k} - \frac{\beta_{k-1}}{2}\bar{\nabla}\mathcal{F}^{md}_{k-1}.  
\end{aligned}
\end{equation}
It can be transformed to an ordinary differential equation (ODE) due to Corollary~\ref{corol:AGD1} when $\vartheta_1 = \delta_{\uptau}^{-1}$ and $\delta_{\uptau}\rightarrow 0$: 
\begin{equation}\label{eq:ODE_1st}
	 \frac{\diff^{2}\bm{\theta}}{\diff \uptau^2} + \frac{3\diff \bm{\theta}}{\uptau \diff \uptau}  
	 = - \frac{1}{\mathcal{L}_{\mathcal{F}}}\left(1 - \frac{1}{\uptau}\right)\frac{\diff \bar{\nabla}\mathcal{F}}{ \diff \uptau} - \frac{1}{\mathcal{L}_{\mathcal{F}}\uptau} \bar{\nabla}\mathcal{F}
\end{equation}
where $\uptau = k\delta_\uptau$ denotes the optimization time of AGD process. The above ODE indicates our method approximately solves the first order system. In comparison, the leapfrog method solves the second order system 
\begin{equation}\label{eq:ODE_2nd}
	 {\diff^{2}\bm{\theta}} / {\diff \uptau^2} 
	 = -\mathcal{L}_{\mathcal{F}}^{-1}\bar{\nabla}\mathcal{F}, 
\end{equation}
which is attained via setting $\lambda_k = \alpha_k^{-1}\beta_k$. However, its update rule results in a series of negative $\{C_{k}\}$ of \eqref{eq:conditionAGparas} and the corresponding \code{AGD()} converges with the rate 
\begin{equation} \label{eq:corol2_converge}
	\mathcal{F}_{N}^{ag} - \mathcal{F}^{*} \leq \frac{2\vartheta_1 \mathcal{L}_{\mathcal{F}}\left\|\bm{\theta}_{0}-\bm{\theta}^{*}\right\|^{2}}{N_{ag}(N_{ag}+1)}
\end{equation}
rather than \eqref{eq:corol1_converge}'s rate. However, it requires a strong convexity~\eqref{eq:convexHull} or high inertia (i.e., large $\mathcal{L}_{\mathcal{F}}$) to ensure the system's stability according to \cite{Ghadimi2016AG-NLP,Feng2021iSAGO}. These requirements can resist the capacity of overcoming the potential field's fluctuation and lower the success rate. That is the reason why Corollary~\ref{corol:AGD1}, rather than \cite{Feng2021iSAGO}'s method, is utilized in \code{AGD()}. Section~\ref{sec:L-reAGD_tune} compares $\mathcal{L}$-reAGD with CHOMP to validate our choice. 

Unlike CHOMP which gains an initial momentum by Monte-Carlo sampling, the AGD method rolls the ball with constant inertia $\mathcal{L}_{\mathcal{F}}$ obeying~\eqref{eq:semi-convexHull} from the zero initial in a smooth landscape. For both of them, the constant inertia determines the ball's resistance to the velocity change by force from the potential field. It means the external force $\bar{\nabla}\mathcal{F}$ could affect the trace of the light ball more quickly than the heavy one. So a proper selection of $\mathcal{L}_{\mathcal{F}}$ highly determines an effective descent towards a minimum. 
The following subsection will detail how to estimate $\mathcal{L}_{\mathcal{F}}$ precisely and how the $\mathcal{L}$-restart method adaptively restarts the AGD-process. 

\subsection{$\mathcal{L}$-Restart of Accelerated Gradient Descent}\label{sec:L-reStart}

\begin{figure}[hbtp]
\begin{centering}
{\includegraphics[width=1\columnwidth]{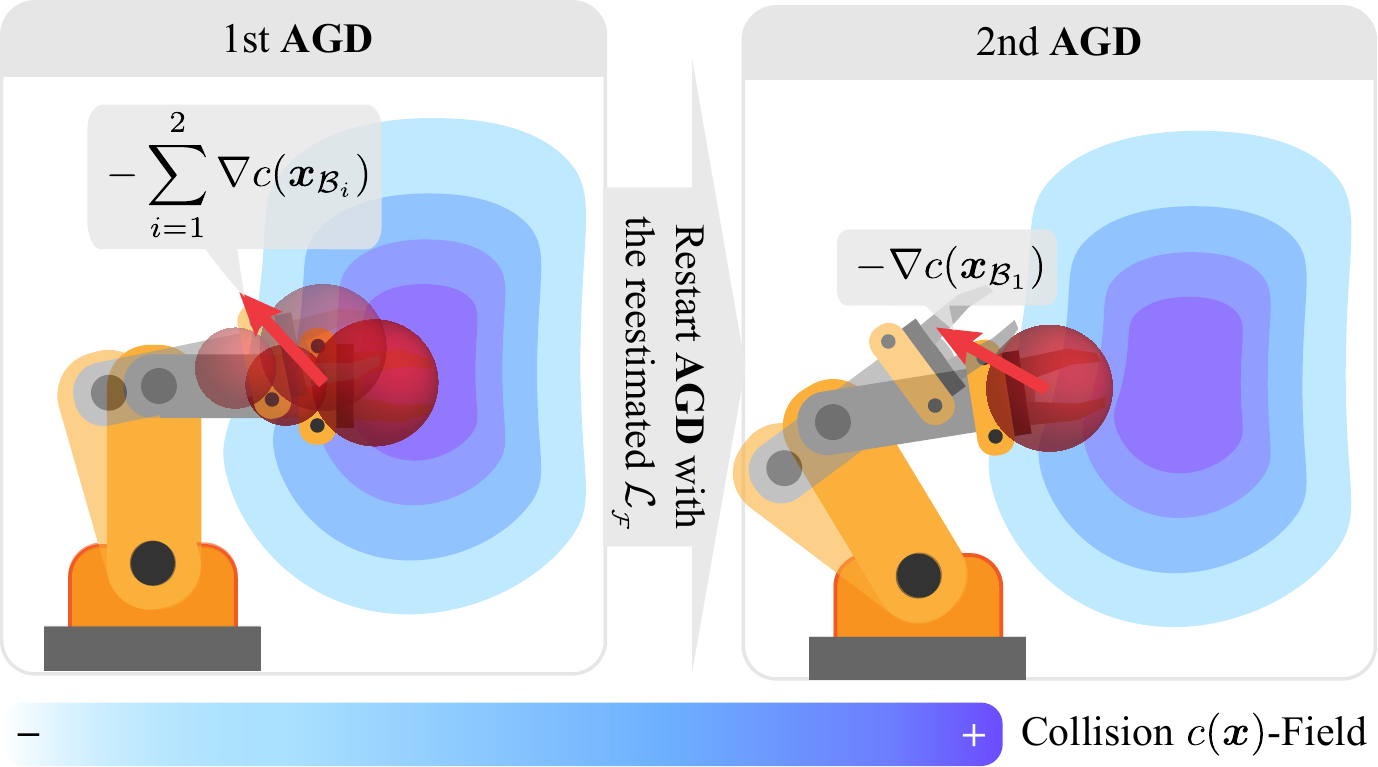}}
\par\end{centering}
\protect\caption{
An example demonstrates how the repulsive force (red arrow) from the collision field $c(x)$ (blue contour map) effects the robot arm by repelling the collision check ball (CCB, red sphere) and how $\mathcal{L}$-reAGD restarts AGD with the reestimated $\mathcal{L}_\mathcal{F}$. 
\label{fig:L-reAGD}}
\end{figure}

When the robot of Figure~\ref{fig:L-reAGD} avoids an obstacle in W-space, the designed collision field $c(\bm{x})$ of \eqref{eq:obs_cost} will provide a repulsive force to push $\mathcal{B}_i$'s body, denoted by the red ball, away from the unsafe area. Since \eqref{eq:discrt_chomp} treats the body avoidance process as the movement of a series of geometrically connected collision-check balls (CCBs) repelled by force. So how the force affects these CCBs determines how effective the collision avoidance is. Figure~\ref{fig:L-reAGD} demonstrates how the AGD-restart happens when the repelled part variates from $\{\mathcal{B}_1, \mathcal{B}_2\}$ to $\{\mathcal{B}_1\}$. When the joint force pushes $\mathcal{B}_1$ and $\mathcal{B}_2$, larger inertia ensures the super ball rolling stably. After $\mathcal{B}_2$ avoids the collision, the large inertia will dampen $\mathcal{B}_1$'s movement. In conclusion, the various combinations of the activated CCBs could lead to various $\mathcal{L}_{\mathcal{F}}$ affecting the descent efficiency. So this section will first propose the methodology of $\mathcal{L}_{\mathcal{F}}$-estimation and AGD-restart, then introduce how to decide the local minimum approaches. 

CHOMP~\cite{Zucker2013CHOMP} and iSAGO~\cite{Feng2021iSAGO} set initial inertia arbitrarily and expand it exponentially to progressively weaken the ball's response to the potential field during movements. While GPMP~\cite{Mukadam2018GPMP} estimates the inertia matrix consisted of mass and rotational inertia based on the LM method to move the ball efficiently. 
Unlike the above methods, $\mathcal{L}$-reAGD restarts the AGD process with the reestimated mass, when the descent performance exceeds expectations. Though some recent studies~\cite{Feng2021iSAGO, Ghadimi2016AG-NLP, Loshchilov2016SGDR,Donoghue2015AR-AG,Liu2010EFLA} focus on AGD-restart to promote the descent rate, 
most of them design a warm restart scheme and estimate $\mathcal{L}_{\mathcal{F}}$ with arbitrary expanding or shrink. 
Our method constructs $\bm{\mathcal{C}} = \bigcup_{j = 1}^{N_{\mathcal{C}}} \mathcal{C}_{j}$ with the reestimated $\mathcal{L}_{\mathcal{F}}^{(j)}$ based on~\eqref{eq:semi-convexHull}. 
Since a hull $\mathcal{C}$ with a larger $\mathcal{L}_\mathcal{F}$ contains a hull with a smaller $\mathcal{L}_\mathcal{F}$, some AGD methods proven optimal for 1st-order descent in some convex cases still perform poorly with an improper $\mathcal{L}_\mathcal{F}$. Section~\ref{sec:L-reAGD_tune} validates that $\mathcal{L}$-reAGD performs more efficiently than AGD with a constant $\mathcal{L}_\mathcal{F}$. 

To improve AGD, we insert \code{AccBreak($\bm{\theta}_k^{md},\mathcal{F},\mathcal{L}_{\mathcal{F}}$)} at Line~\ref{alg:AGP-STO:MinAprch} of Algorithm~\ref{alg:AGD} to evaluate the step performance and adaptively restart AGD. According to \eqref{eq:semi-convexHull} and Section~\ref{sec:AGD}, our method first defines the predicted reduction 
\begin{equation}\label{eq:pred}
\small
\begin{aligned}
	\textit{pred}_\text{low} &= \left\langle \bar{\nabla} \mathcal{F}_{k-1}^{}, \bm{\theta}_{k}^{}-\bm{\theta}_{k-1}^{}\right \rangle + \frac{c_{\textit{low}}\mathcal{L}_{\mathcal{F}}}{2}\left\|\bm{\theta}_{k}^{}-\bm{\theta}_{k-1}^{}\right\|^{2} \\
	\textit{pred}_\text{up} &= \left\langle \bar{\nabla} \mathcal{F}_{k-1}^{}, \bm{\theta}_{k}^{}-\bm{\theta}_{k-1}^{}\right \rangle - \frac{c_{\textit{up}}\mathcal{L}_{\mathcal{F}}}{2}\left\|\boldsymbol{\theta}_{k}^{}-\boldsymbol{\theta}_{k-1}^{}\right\|^{2}
\end{aligned}
\end{equation}
and then designs 
\begin{equation}
\small
	\code{AccBreak($\bm{\theta}_k^{md},\mathcal{F},\mathcal{L}_{\mathcal{F}}$)} = 
	\begin{cases}
		\textbf{True} \hfill & \text{if } \textit{ared} > \textit{pred}_\text{low} \\ 
		\textbf{True} \hfill & \text{if } \textit{ared} < \textit{pred}_\text{up} \\ 
		\textbf{False} \hfill & \text{Otherwise}, 
	\end{cases} 
\end{equation}
where $\textit{ared} = \mathcal{F}_{k-1} - \mathcal{F}_{k}$ denotes the actual reduction. Besides, $\textit{ared} > \textit{pred}_{\text{low}}$ means AGD performs with insufficient reduction, while $\textit{ared} < \textit{pred}_{\text{up}}$ means $\mathcal{L}_{\mathcal{F}}$ resists the AGD-step. So the both conditions can activate the restart process with the reestimated $\mathcal{L}_{\mathcal{F}}$. Next, we will introduce how to reestimate $\mathcal{L}_{\mathcal{F}}$ facing these two conditions. 

\subsubsection{$\textit{ared} > \textit{pred}_{\text{low}}$}

Based on Corollary~\ref{corol:AGD1} and \eqref{eq:semi-convexHull}, we could gain a prerequisite an AGD-step needs to satisfy within a proper hull $\mathcal{C}(\mathcal{L}_{\mathcal{F}})$: 
\begin{equation} \nonumber
\begin{array}{c}
	\mathcal{F}_{k-1} - \mathcal{F}\left(\bm{\theta}_{k-1} - \frac{1+\vartheta_2}{\vartheta_1\mathcal{L}_\mathcal{F}}\bar{\nabla}\mathcal{F}_{k}\right) 
	\geq \frac{\underline{\varpi}_k}{\mathcal{L}_{\mathcal{F}}},  \\[6pt]
	\underline{\varpi}_k = \frac{2\vartheta_{2}+k+1}{\vartheta_{1}(k+1)}\left( 1-\frac{2\vartheta_{2}+k+1}{\vartheta_{1}(k+1)}-\frac{\vartheta_{2}^{2}}{\vartheta_{1}} \right) \|\bar{\nabla}\mathcal{F}_{k}\|^2. 
\end{array}
\end{equation}
When the above relationship does not hold, a renewed $\mathcal{L}_{\mathcal{F}}$ will reconstruct another hull $\mathcal{C}$. To renew $\mathcal{L}_{\mathcal{F}}$, we first define 
\begin{equation}
\nonumber
\begin{aligned}
	\phi(\mathcal{L}_{\mathcal{F}}) &= \mathcal{F}\left(\bm{\theta}_{k-1} - \frac{1+\vartheta_2}{\vartheta_1\mathcal{L}_\mathcal{F}}\bar{\nabla}\mathcal{F}_{k}\right) - \mathcal{F}(\bm{\theta}_{k-1}). \\
\end{aligned}
\end{equation}

Since a low resistant (i.e. a small $\mathcal{L}_{\mathcal{F}}^{(j)}$) to exterior force leads to the unacceptable movement of the super-ball. It is reasonable to say the renewed semi-convex hull $\mathcal{C}_{j+1}$ contains the former hull $\mathcal{C}_{j}$, meaning that $\mathcal{L}_{\mathcal{F}}^{(j+1)} \in (\mathcal{L}_{\mathcal{F}}^{(j)}, \infty)$. So we formulate a quadratic approximation function 
\begin{equation} 
	\nonumber 
	\phi_{q}(\mathcal{L}_{\mathcal{F}}) = \frac{\mathcal{L}_{\mathcal{F}}^{(j)}\phi(\mathcal{L}_{\mathcal{F}}^{(j)}) - \frac{1+\vartheta_2}{\vartheta_1}\dot{\phi}(\infty)}{{\mathcal{L}_{\mathcal{F}}} ^{2}/\mathcal{L}_{\mathcal{F}}^{(j)}}
	+\frac{1+\vartheta_2}{\vartheta_1\mathcal{L}_{\mathcal{F}}}{\dot{\phi}(\infty)}
\end{equation}
like the line search methods~\cite{Al-baali1985LineSearch, Nocedal2006NumericalOpt}. Unlike these methods, however, our algorithm only approximates $\phi(\mathcal{L}_{\mathcal{F}})$ in the quadratic (i.e., second-order) form and abandons their higher-order form, considering the computational cost. Then we can update $\mathcal{L}_{\mathcal{F}}$ via solving 
\begin{equation}\nonumber
	\argmin_{\mathcal{L}_{\mathcal{F}}}~\phi_{q}(\mathcal{L}_{\mathcal{F}}) +  \frac{\underline{\varpi}_k}{\mathcal{L}_{\mathcal{F}}}
\end{equation}
and obtain
\begin{equation}\label{eq:LipschitzEst1}
	\frac{\mathcal{L}_{\mathcal{F}}^{(j+1)}}{\mathcal{L}_{\mathcal{F}}^{(j)}} = \frac{\mathcal{L}_{\mathcal{F}}^{(j)} \phi(\mathcal{L}_{\mathcal{F}}^{(j)}) - \frac{1+\vartheta_2}{\vartheta_1} \dot{\phi}(\infty^{})}{-\frac{1}{2}\left(\frac{1+\vartheta_2}{\vartheta_1} \dot{\phi}(\infty^{})+\underline{\varpi}_k\right)}.
\end{equation}
The above $\mathcal{L}_{\mathcal{F}}$-reestimation will perform sequentially till the condition $\textit{ared} \leq \textit{pred}_{\text{low}}$ is satisfied. 

The above method adjusts $\mathcal{L}_{\mathcal{F}}$ iteratively, only calculating $\dot{\phi}(\infty)$ for once. Its utilization occurs  when the AGD-step is rejected initially within $\mathcal{C}_{j}$. However, it requires the negative-definite of $\dot{\phi}(\infty)$ to ensure the positive definite of the renewed $\mathcal{L}_{\mathcal{F}}$. So how to deal with the semi-negative definition situation? An intuition that directly finds the constraint's boundary could satisfy the condition. In this way, we gain
\begin{equation} \label{eq:Lipschitz_expand_2}
	\mathcal{L}_{\mathcal{F}}^{(j+1)} = \frac{\textit{ared}+\left\langle \bar{\nabla} \mathcal{F}_{k-1}^{md}, \bm{\theta}_{k-1}^{md}-\bm{\theta}_{k}^{md}\right \rangle}{\frac{1}{2} c_{\textit{low}}\left\|\bm{\theta}_{k}^{md}-\bm{\theta}_{k-1}^{md}\right\|^{2}}. 
\end{equation}
%

\subsubsection{$\textit{ared} < \textit{pred}_{\text{up}}$}

Unlike the above condition, urging for a sufficient reduction, this condition indicates the inertia (i.e., $\mathcal{L}_{\mathcal{F}}$) resists the AGD process. To modify this poor condition, we first observe how the force $\|\bar{\nabla}\mathcal{F}\|$ variants with respect to the change of position $\|\bm{\theta}\|$ and then modify $\mathcal{L}_{\mathcal{F}}$ to meet the $\mathcal{L}_{\mathcal{F}}$-continuous requirement~\eqref{eq:semi-convexHull} of $\bar{\nabla}\mathcal{F}$. Based on that, we could get the upper bound:   
\begin{equation} \label{eq:expd_up}
	\mathcal{L}_{\mathcal{F}}^{(j+1)} \leq \frac{\| \bar{\nabla} \mathcal{F}_{k}^{md} - \bar{\nabla} \mathcal{F}_{k-1}^{md} \|}{c_{\textit{up}}\| \bm{\theta}_{k}^{md} - \bm{\theta}_{k-1}^{md} \|}.  
\end{equation}
Similar to \eqref{eq:Lipschitz_expand_2}, we could get the lower bound: 
\begin{equation} \label{eq:expd_low}
	\mathcal{L}_{\mathcal{F}}^{(j+1)} \geq \frac{\left\langle \bar{\nabla} \mathcal{F}_{k-1}^{md}, \bm{\theta}_{k}^{md}-\bm{\theta}_{k-1}^{md}\right \rangle - \textit{ared}}{\frac{1}{2} c_{\textit{up}}\left\|\bm{\theta}_{k}^{md}-\bm{\theta}_{k-1}^{md}\right\|^{2}}, 
\end{equation}
according to \eqref{eq:semi-convexHull} and \eqref{eq:pred}. In this way, our method iteratively resamples the $\mathcal{L}_{\mathcal{F}}$ between the lower~\eqref{eq:expd_low} and upper~\eqref{eq:expd_up} bound until $\textit{ared} \geq \textit{pred}_{\text{up}}$. 
\subsubsection{Local minimum approach}

\begin{figure}[hbtp]
\begin{centering}
{\includegraphics[width=1\columnwidth]{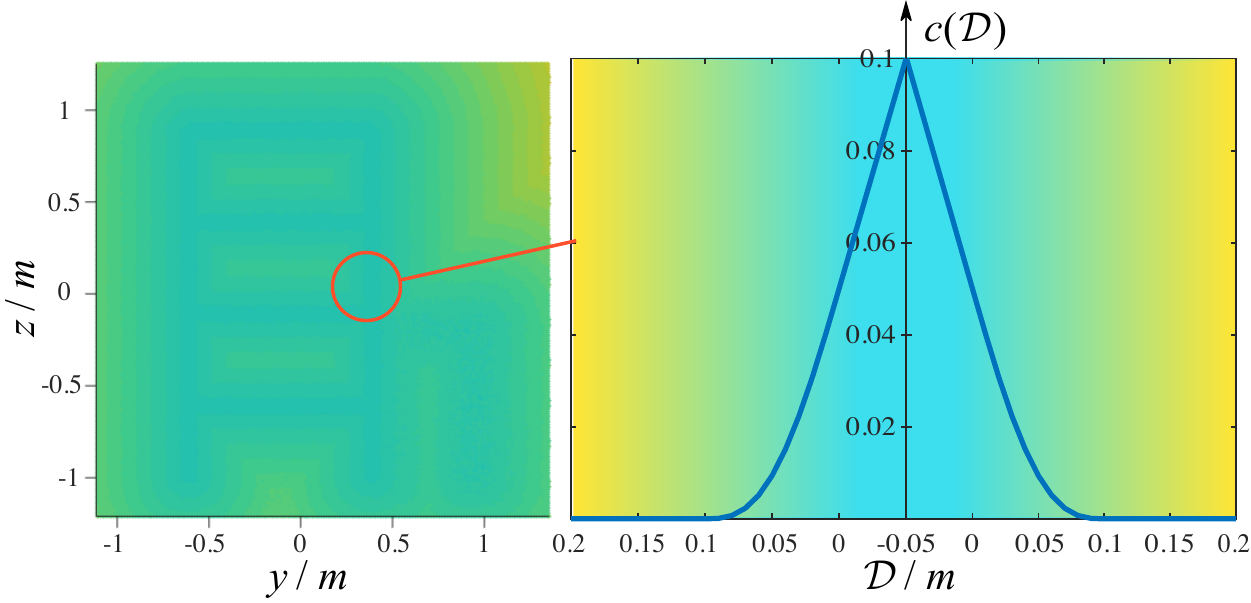}}
\par\end{centering}
\protect\caption{
An example shows the signed distance (the color lightens from blue to yellow with the signed value increasing) and the curve of collision cost generated via~\eqref{eq:collisonCost}. The triangular-shaped part illustrates the differential discontinuity of the collision cost field in the corridor case.     
\label{fig:localmin}}
\end{figure}

Though the obstacle cost function $c\left(\mathcal{D}\right)$ designed in Section~\ref{sec:collisionCost} is twice-differential continuity, the overall cost function (Figure~\ref{fig:localmin}) is usually first-order differential discontinuity when the obstacle is relatively narrow compared to the robot itself. It means that the objective functional $\mathcal{F}$ is strongly nonconvex. We call it strong because it is formed by the steep valley of $\mathcal{F}$ rather than the low fluctuation that satisfies \eqref{eq:semi-convexHull} and can be overcome by Corollary~\ref{corol:AGD1}. $\mathcal{L}$-reAGD will adaptively expand $\mathcal{L}_{\mathcal{F}}$ large enough to descend robustly by confronting the steep valley leading to a local minimum. However robustly the steps perform, they could not change the fact that the process finally approaches the local minimum. So it is meaningful to detect the local minimum approach (MinAprch) and find a way to avoid it.    

Our former study~\cite{Feng2021iSAGO} measures the included angle between the activated CCBs and assumes MinAprch when the angle is out of acceptance. It is useful, especially when implementing the robot with the serially connected CCBs. However, it highly depends on how the CCBs are connected, and it is sometimes  incompatible with the parallel connection case, such as the multi-vehicle allocation in Section~\ref{sec:AGVs}. 

This paper studies the body-obstacle stuck case shown in Figure~\ref{fig:localmin} and proposes \code{MinAprch()} to determine the local minimum approaches. When the robot is stuck in an obstacle, we will see whichever direction the robot moves in within hull $\mathcal{C}$, and the functional cost $\mathcal{F}$ will stay invariant or even grows. In such a case, the curvature of the descent step will become negative, meaning there both exists hill and valley within hull $\mathcal{C}$ shown in Figure~\ref{fig:localmin}. Another phenomenon that the over-expanded $\mathcal{L}_{\mathcal{F}}$ could not perform an acceptable descent also indicates MinAprch. In this way, we design
\begin{equation}\label{eq:MinAprch}
	\code{MinAprch()} = 
	\begin{cases}
		\textbf{True} \hfill & \text{if } \dot{\phi} \geq |\textit{ared}|\cdot\upphi\text{tol} \\ 
		\textbf{True} \hfill & \text{if } \mathcal{L}_{\mathcal{F}}^{(j+\jmath)} \geq \mathcal{L}_\mathcal{F}^{(j+1)} \mathcal{L}\text{tol} \\ 
		\textbf{False} \hfill & \text{Otherwise}, 
	\end{cases} 
\end{equation}
where $\upphi\text{tol}$ prevents the AGD-step from the negative curvature, $\mathcal{L}\text{tol}$ avoids the over-expanded $\mathcal{L}_\mathcal{F}$, and $\jmath > 1 \in \mathds{N}$ is the loop-index for the sequential estimation from $\mathcal{L}_{\mathcal{F}}^{(j)}$. Someone may ask why we choose $\jmath > 1$ than $\jmath \geq 1$? Because $\mathcal{L}$-reAGD arbitrarily initializes $\mathcal{L}_{\mathcal{F}} = \|\bar{\nabla}\mathcal{F}\|$ after ASTO or AGD-restart. So it is rational to take $\mathcal{L}_{\mathcal{F}}^{j+1}$ as the reference value and choose $\jmath>1$ to judge whether $\mathcal{L}_{\mathcal{F}}$ is over-expanded. 

When \code{MinAprch()} returns \textbf{True} in Line~\ref{alg:AGP-STO:MinAprch} of Algorithm~\ref{alg:AGP-STO}, AGP-STO will stimulate ASTO (Section~\ref{sec:ASTO}, Algorithm~\ref{alg:ASTO}) to find a semi-convex subspace nearby the optimum. Section~\ref{sec:analysis} validates that it saves the computational resource consumed for the extra iterations for an acceptable $\mathcal{C}$ and elevates the success rate via sampling-based motion planning. 

\section{Adaptive Stochastic Trajectory Optimization}\label{sec:ASTO}

This section proposes an adaptive stochastic trajectory optimization (ASTO) algorithm inspired by STOMP~\cite{Kalakrishnan2011STOMP}. ASTO involves stochastic trajectory optimization using the noisy trajectories sampled from the adaptive Gaussian distribution. Unlike STOMP whose Gaussian covariance represents the sum of squared acceleration, we utilize the GP model 
\begin{equation}\nonumber
	\begin{aligned}
		&  \underset{\boldsymbol{\theta}(t,t')}{\text{minimize}}
		&& \mathcal{F} [\boldsymbol{\theta}(t,t')] \\
		& \text{subject to}
		&&  \boldsymbol{\theta}(t,t') \sim \mathcal{GP}(\bm{\mu}(t,t'), \bm{\mathcal{K}}(t, t^{\prime})). \\
	\end{aligned}
\end{equation}
to gain the prior knowledge of the trajectory distribution like \cite{Petrovic2019HGP-STO} and interact with the potential field $\mathcal{F}(\bm{\theta})$ to learn the optimal policy adaptively via trajectory sampling. 

\begin{figure}[htbp]
\begin{centering}
{\includegraphics[width=1\columnwidth]{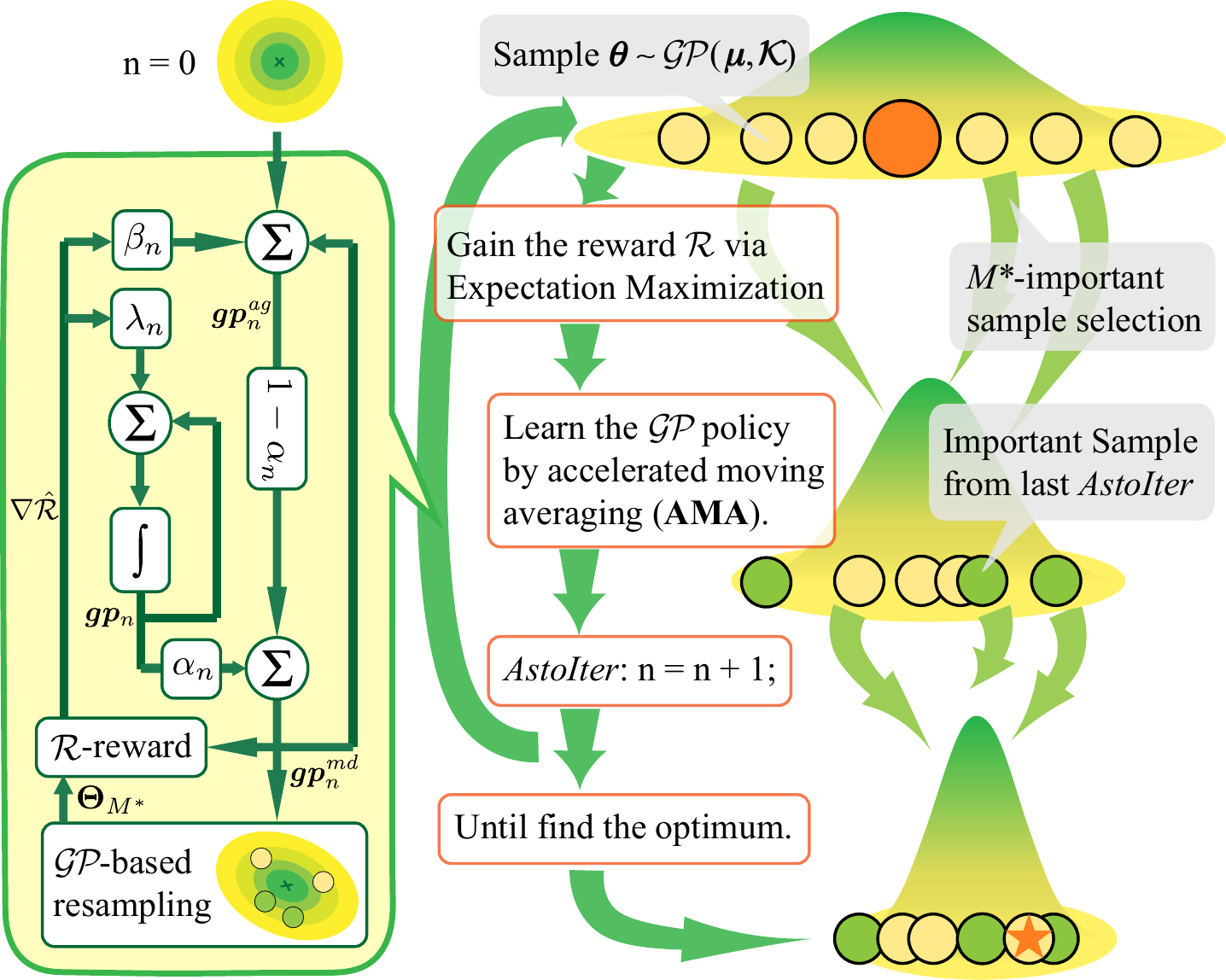}}
\par\end{centering}
\protect\caption{
A system diagram on the left side shows how ASTO applies AGD to perform an AMA process during resampling and adaptively learns a $\mathcal{GP}$ policy. The right side demonstrates how to select $M^*$ important samples with low $\mathcal{F}$-cost and utilize them to renew the $\mathcal{GP}$ model via AMA. 
\label{fig:ASTO}}
\end{figure}

Continuing the interpretation of AGP-STO in Section~\ref{sec:AGP-STO_intuition}, we treat ASTO (Figure~\ref{fig:ASTO}) as reinforcement learning (RL). The RL-based process activates the super-ball, representing the trajectory, to learn the $\mathcal{F}$-field policy (i.e., GP model). During the interaction, ASTO resamples the trajectories sequentially and evaluates their potential energy to reward. During the RL rewarding, we first adopt the arithmetic operator in~\cite{Kalakrishnan2011STOMP} and conduct an exponential measurement of the sampled trajectories. Then we utilize the expectation-maximization (EM) method in Section~\ref{sec:EM} to gain a reward. To improve the learning progress, we propose an accelerated moving averaging (AMA, Section~\ref{sec:AMA}) method based on AGD (Section~\ref{sec:AGD}), which generalizes and reformulates the exponential moving averaging (EMA) method. 

\begin{algorithm}[htbp]
\caption{ASTO}\label{alg:ASTO}
\DontPrintSemicolon
\SetKwInOut{Input}{Input}
\SetKwFunction{ASTO}{ASTO}
\SetKw{Or}{or}
\SetKwComment{Comment}{}{}

\Input{functional $\mathcal{F}$ and prior $\mathcal{GP}(\bm{\mu}_0, \bm{\mathcal{K}}_0)$; }
\textbf{Initialize}: {$\{\bm{\mu}_1^\textit{md}, \bm{\mathcal{K}}_1^\textit{md}\} = \{ \bm{\mu}_0, \bm{\mathcal{K}}_0 \}$, $K = K_0$, $\mathfrak{F}_K = \O$, $\bm{\Theta}_{K} = \O$; }\;
\For {\text{AstoIter}: $ n = 1 \dots N_{\textit{Asto}}$}{
	\If{$n > 1$}{
		$\bm{\Theta}_{K} = \bm{\Theta}_{M^{*}}$; 
		$\mathfrak{F}_{K} = \mathfrak{F}_{M^{*}}$; 
		$K = K_{0}-M^{*}$; \;
	}
	\For {\textit{SmpIter}: $k = 1 \dots K$}{
		\eIf{$\|\bm{\mathcal{K}}_n^\textit{md}\| \geq \bar{\mathcal{K}}\text{tol}$}{
			Sample trajectory $\bm{\theta}_k \sim \mathcal{U}(\bm{\theta}_\text{min}, \bm{\theta}_\text{max}); $\;
		}{
			Sample trajectory $\bm{\theta}_k \sim \mathcal{GP}(\bm{\mu}_n^\textit{md}, \bm{\mathcal{K}}_n^\textit{md})$; \;
		}
		Confine $\bm{\theta}_k$ in the motion constraints~\eqref{eq:motionConstraintCost}; \; 
		$\bm{\Theta}_{K} \leftarrow \bm{\Theta}_{K} \bigcup \bm{\theta}_k$; 
		Evaluate $\mathcal{F} (\boldsymbol{\theta}_k)$ by~\eqref{eq:chomp_cost}; \;
		\eIf {$\mathcal{F}_{\textit{obs}}(\bm{\theta}_{k}) \leq \mathcal{G}\text{tol}$}{
			\textbf{return} $\bm{\theta}_{k}$ and $\mathcal{GP}$ modified by~\eqref{eq:ASTO_GP}. 
		}{
			$\mathfrak{F}_{K} = \mathfrak{F}_{K} \bigcup \mathcal{F} (\boldsymbol{\theta}_k)$;
		}
	}
	Extract the $M^{*}$-important samples with low cost from $\bm{\Theta}_{K}$ and construct a set $\bm{\Theta}_{M^{*}}$ and its $\mathfrak{F}_{M^{*}}$; \;
	Update the reward $Q(\bm{\mu},{\bm{\mathcal{K}}} \mid \bm{\mu}_n^\textit{md},\bm{\mathcal{K}}_n^\textit{md})$ via EM; \;
	\tcc{Learn the policy $\bm{\mu},\bm{\mathcal{K}}$ via $\textit{AMA}$}\label{InAMA}
	$\bm{\mu}^{\textit{ag}}_n = \bm{\mu}^\textit{md}_n + \beta_n^\mu \nabla \hat{\mathcal{R}}(\bm{\mu}^\textit{md}_n)$; \;
	$\bm{\mathcal{K}}^{\textit{ag}}_n = \bm{\mathcal{K}}^\textit{md}_n + \beta_n^\kappa \nabla \hat{\mathcal{R}}(\bm{\mathcal{K}}^\textit{md}_n)$; \;
	$\bm{\mu}_n = \bm{\mu}_{n-1} + \lambda_n^\mu \nabla \hat{\mathcal{R}}(\bm{\mu}^\textit{md}_n)$; \;
	$\bm{\mathcal{K}}_n = \bm{\mathcal{K}}_{n-1} + \lambda_n^\kappa \nabla \hat{\mathcal{R}}(\bm{\mathcal{K}}^\textit{md}_n)$; \;
	$\bm{\mu}^\textit{md}_{n+1} = (1 - \alpha_n^\mu)\bm{\mu}_{n}^{\textit{ag}} + \alpha_n^\mu \bm{\mu}_n$; \;
	$\bm{\mathcal{K}}^\textit{md}_{n+1} = (1 - \alpha_n^\kappa)\bm{\mathcal{K}}_{n}^{\textit{ag}} + \alpha_n^\kappa \bm{\mathcal{K}}_n$; \;
	\If{\small$\|\bm{\mathcal{K}}_n^{\textit{md}}\| \leq \underline{\mathcal{K}}\text{tol}$ \Or $\exists\eig(\bm{\mathcal{K}}_n^\textit{md}) \leq 0$ \Or $n\geq N_{\textit{Asto}}$}{
		$\bm{\theta} = \argmin_{\bm{\mu}^\textit{md}_n \bigcup \Theta_K} \mathcal{F}(\bm{\mu}^\textit{md}_n) \bigcup \mathfrak{F}_K$; \;
		\eIf{${\mathcal{F}(\bm{\theta})-\mathcal{F}(\bm{\mu}_0)} \leq {\mathcal{F}(\bm{\mu}_0)}\cdot c\mathcal{F}tol$}{
			\Return $\bm{\theta}$ and $\mathcal{GP}$ modified by~\eqref{eq:ASTO_GP}. 
		}
		{
			\Return $\bm{\mu}_{0}$ and $\mathcal{GP}(\bm{\mu}_{0}, \bm{\mathcal{K}}_0)$. 
		}
	}
}
\end{algorithm}
%
\subsection{Expectation-Maximization for GP-learning} \label{sec:EM}

This Section utilizes the adaptive GP and EM~\cite{Wu1983EM} method to construct RL's reward and learning strategy. AGPR~\cite{Jiang2010AGP} constructs sliding windows to make a model-free prediction. Unlike AGPR learning from the priorly collected and labeled dataset, ASTO gains the dataset $\bm{\Theta}_{K}$ labeled by $\mathfrak{F}_{K}$ via interacting with the field $\mathcal{F}$ (i.e., online sampling and evaluation). Moreover, the $\mathcal{R}$-rewarding proceeds by the EM method. 

During the interaction, ASTO (Algorithm~\ref{alg:ASTO}) first samples a set $\bm{\Theta}_{K}$ consisting of $K$ sampled trajectories $\bm{\theta}_{k}$ obeying $\mathcal{GP}(\bm{\mu}, \bm{\mathcal{K}})$. Then it confines each sample in the motion constraints (Section~\ref{sec:motionConstraint}). Since a large $\bm{\mathcal{K}}$-norm can invoke the sample aggregation nearby the motion constraints, we sample the trajectories obeying the uniform distribution $\mathcal{U}(\bm{\theta}_\text{min},\bm{\theta}_\text{max})$ when the covariant norm is beyond the upper threshold $\bar{\mathcal{K}}\text{tol}$ for a more comprehensive gathering of environmental information. After that, it evaluates each sample's potential energy $\mathcal{F}(\bm{\theta}_k)$ and gathers them in the set $\mathfrak{F}_{K}$.  

So far, ASTO has performed the action based on the policy $\mathcal{GP}(\bm{\mu},\bm{\mathcal{K}})$ and gathered environment information $\mathfrak{F}_{K}$. Now it is time to gain the reward from the interaction. ASTO adopts the EM method to gain the rewards for policy learning like~\cite{Dayan1997EM-RL}.  It contains the E-steps, which estimate the logarithm likelihood for reward gathering, and the M-steps that maximize the reward for policy updating. 


\subsubsection{E-step} \label{sec:E-step}
Given the current policy $\mathcal{GP}(\bm{\mu}, \bm{\mathcal{K}})$, we determine the conditional distribution of $\mathcal{F}_{k}=\mathcal{F}(\bm{\theta}_{k})$~\eqref{eq:discrt_chomp} based on the Bayes theorem: 
\begin{equation}
\label{eq:prob_cond}
	p_{k} = \operatorname{P}\left(\mathcal{F}_{k} \mid \bm{\theta}_k ;\bm{\mu},\bm{\mathcal{K}}\right) = \frac{e^{-\varpi\mathcal{F}_{k}} f(\bm{\theta}_{k};\bm{\mu},\bm{\mathcal{K}})}
	{\sum_{k=1}^{K} e^{-\varpi\mathcal{F}_{k}} f(\bm{\theta}_{k};\bm{\mu},\bm{\mathcal{K}})}, 
\end{equation}
where $f(\bm{\theta}_{k};\bm{\mu},\bm{\mathcal{K}})$ is the $\mathcal{GP}$-defined probabilistic distribution function of a sample $\bm{\theta}_{k}$. $e^{-\varpi \mathcal{F}_k}$ denotes the latent probability~\footnote{The word latent means the overall probability distribution $p(\mathcal{F}[\bm{\theta}])$ of $\bm{\theta}$ in the potential field $\mathcal{F}$ is unmeasurable. } of a sample $\bm{\theta}_k$, where the arithmetic operator~\cite{Kalakrishnan2011STOMP}
\begin{equation}\nonumber
	\varpi \mathcal{F}_k
	= h_{p} \frac{\mathcal{F}_k - \min\mathfrak{F}_K}{\max\mathfrak{F}_K - \min\mathfrak{F}_K}, 
\end{equation}
regulates the probabilistic distribution given $\mathfrak{F}_{K}$ with a scalar $h_{p}$ to control the variance. 

After gaining the probability $\operatorname{P}\left(\mathcal{F}_{k} \mid \bm{\theta}_k ;\bm{\mu},\bm{\mathcal{K}}\right)$ of \eqref{eq:prob_cond}, we get the reward $\mathcal{R}$ of the policy $\mathcal{GP}(\bm{\mu}',\bm{\mathcal{K}}')$ need to learn: 
\begin{equation}\label{eq:reward}
	\begin{aligned}
		\mathcal{R}(\bm{\mu}',\bm{\mathcal{K}}' \mid {\bm{\mu}},{\bm{\mathcal{K}}}) 
		&= \mathds{E}_{\mathfrak{F}_{K}\mid\bm{\Theta}_{K};\bm{\mu},\bm{\mathcal{K}}} \left[\log L(\bm{\mu}',\bm{\mathcal{K}}'; \bm{\Theta}_{K})\right]\\
		&= \sum_{k=1}^{K} p_{k} \log L(\bm{\mu}',\bm{\mathcal{K}}';\bm{\theta}_{k}), 
	\end{aligned}
\end{equation}
where the function  
\begin{equation}\nonumber
	L(\bm{\mu}',\bm{\mathcal{K}}'; \bm{\Theta}_{K}) = \prod_{k=1}^{K} f(\bm{\theta}_{k};\bm{\mu}',\bm{\mathcal{K}}')
\end{equation}
computes the likelihood of $\mathcal{GP}({\bm{\mu}'},{\bm{\mathcal{K}}'})$ given the sample set $\bm{\Theta}_{K}$. The logarithm likelihood measures the sum of the negative Mahalanobis distance between the samples and the policy to optimize. So the reward $\mathcal{R}$ will somewhat encourage the future policy getting close to the former. At the same time, $\mathcal{R}$ will guid the future policy towards the samples with lower energy/cost through combining the conditional probability $p_k$ informed by the environment. 

\subsubsection{M-step} \label{sec:M-step}

After gaining the reward from E-step, M-step will prompt the policy learning. From  Appendix~\ref{appdx:paraASTO}, $\mathcal{R}$ could be expressed in a quadratic form. So we obtain the optimal ${\bm{\mu}'},{\bm{\mathcal{K}}'}$ by searching for the maximum of $\mathcal{R}$: 
\begin{equation} \nonumber
	\argmax_{\bm{\mu}',\bm{\mathcal{K}}'} \mathcal{R}(\bm{\mu}',\bm{\mathcal{K}}' \mid {\bm{\mu}},{\bm{\mathcal{K}}}). 
\end{equation}

However, some useless information gathered by the samples with high cost could attach some extra computational resources for the EM-based $\mathcal{GP}$-adaptation and disturb the policy learning process. So we extract $M^{*}$ important samples with low cost from $\bm{\Theta}_{K}$ like STOMP and only take $\{\bm{\Theta}_{M^{*}}, \mathfrak{F}_{M^{*}}\}$ other than $\{\bm{\Theta}_{K}, \mathfrak{F}_{K}\}$ into the calculation of the policy updating rules 
\begin{align} \label{eq:updateGP}
	\nabla \hat{\mathcal{R}}(\bm{\mu}') = \hat{\bm{\mu}} - \bm{\mu}' {,~}
	\nabla \hat{\mathcal{R}}(\bm{\mathcal{K}}') = \hat{\bm{\mathcal{K}}} - \bm{\mathcal{K}}'
\end{align}
due to Appendix~\ref{appdx:paraASTO}~\footnote{$\nabla\hat{\mathcal{R}}$ denotes a quasi-gradient than the gradient $\nabla \mathcal{R}$. }, where 
\begin{align}\nonumber
	\hat{\bm{\mu}} = \frac {\sum_{m=1}^{M^{*}} p_m \bm{\theta}_m}{ \sum_{m=1}^{M^{*}} p_m} {,~}
	\hat{\bm{\mathcal{K}}} = \frac{\sum_{m=1}^{M^{*}} p_{m} \cdot \left(\bm{\theta}_{m}- \hat{\bm{\mu}} \right)^{\otimes {2}}}{\sum_{m=1}^{M^{*}} p_{m}}, 
\end{align}
and $[\cdot]^{\otimes 2} = [\cdot] [\cdot]^{\mathrm{T}}$ denotes the outer product of the two columnized values $[\cdot]$. If we use 
\begin{equation} \nonumber
	\bm{\mu}^\text{M} \leftarrow \bm{\mu} + \nabla \hat{\mathcal{R}}({\bm{\mu}}),~ 
	\bm{\mathcal{K}}^\text{M} \leftarrow \bm{\mathcal{K}} + \nabla \hat{\mathcal{R}}(\bm{\mathcal{K}}) 
\end{equation}
to renew the policy, we can see $\nabla \hat{\mathcal{R}}(\bm{\mu}^\text{M}, \bm{\mathcal{K}}^\text{M}) = 0$, approaching the maximum reward with $\bm{\Theta}_{M^*}$. 

Now let us define $\{{\bm{\mu}}^\text{K}, {\bm{\mathcal{K}}}^\text{K}\}$ as the learned policy informed by the whole set $\bm{\Theta}_{K}$ than the selected set $\bm{\Theta}_{M^*}$ and see how they perform differently from $\{\bm{\mu}^\text{M}, \bm{\mathcal{K}}^\text{M} \}$ in the rewarding and learning process. Since \eqref{eq:reward}'s quadratic form indicates the convexity of the reward ascent problem, we directly get
\begin{equation} \nonumber
	\mathcal{R}({\bm{\mu}}^\text{K}, {\bm{\mathcal{K}}}^\text{K} | \bm{\mu}, \bm{\mathcal{K}}) \geq \mathcal{R}(\bm{\mu}^\text{M}, \bm{\mathcal{K}}^\text{M} | \bm{\mu}, \bm{\mathcal{K}}),  
\end{equation}
which reveals that the $M^*$-selection method releases the strict $\mathcal{R}$-ascend. However, a phenomenon that
\begin{equation}\nonumber
	\mathds{E}_{\mathfrak{F}_{K}\mid\bm{\Theta}_{K};\bm{\mu},\bm{\mathcal{K}}} (\mathfrak{F}_K) \geq \mathds{E}_{\mathfrak{F}_{M^*}\mid\bm{\Theta}_{K};\bm{\mu},\bm{\mathcal{K}}} (\mathfrak{F}_{M^*})
\end{equation}
gives a rough intuition that $\mathcal{F}(\bm{\mu}^\text{K}) \geq \mathcal{F}({\bm{\mu}}^\text{M})$~\footnote{The word rough means the functional $\mathcal{F}$ is nonconvex and the inequality that $\mathcal{F}(\bm{\mu}^\text{K}) \geq \mathcal{F}({\bm{\mu}}^\text{M})$ cannot hold for $\forall {\bm{\mu}^\text{K}}, {\bm{\mu}}^\text{M}$ satisfying the inequality relationship of $\mathds{E}_{\mathfrak{F}\mid\bm{\Theta};\bm{\mu},\bm{\mathcal{K}}}(\mathfrak{F})$.}. So we always use $M^* \approx K/2$ to trade off between a better rewarding and a more efficient learning. That is also why the next part proposes the AMA method to improve ASTO's learning ability.

\subsection{Accelerated Moving Averaging}\label{sec:AMA}

Moving averaging, shown in Figure~\ref{fig:ASTO}, smoothes the stochastic learning process via the damped convolution to analyze the sampled data statistically in a time series. 
Recently, AMP~\cite{Kanazawa2019AMP} and iLSTF~\cite{Kanazawa2019iLSTF} employed the exponential moving averaging (EMA) method and introduced the forgetting coefficient to make a real-time prediction of a worker's movement. One of EMA's significant contributions is its high generalization capability via historical knowledge integration in a damped or forgetting way. This section adopts the fundamental theorem of EMA and proposes AMA for adaptive learning. 

Since the end of Section~\ref{sec:M-step} indicates the release of the strict reward ascent could somewhat improve the learning procedure during the trajectory-functional interaction. We develop AMA based on AGD (Section~\ref{sec:AGD}).  It pretty much matches the rudimental theorem of AGD, which combines the historical information of momenta and releases the strict descent to approach the optimal convergence. So now let us first see how AMA performs EMA. According to \eqref{eq:updateGP} and Algorithm~\ref{alg:ASTO}, we update $\bm{\mu}, \bm{\mathcal{K}}$ with step $\{\alpha_n^{\mu},\alpha_n^{\kappa}, \beta_n^\mu, \beta_n^\kappa, \lambda_n^{\mu}, \lambda_n^{\kappa} \} = \{\alpha^{\mu}, \alpha^{\kappa}, \alpha^{\mu}, \alpha^{\kappa}, \alpha^{\mu}, \alpha^{\kappa}\}$, then gain
\begin{equation}
\label{eq:EMA}
\begin{aligned} 
	\bm{\mu}^{\textit{md}}_{n+1} &= (1-\alpha^\mu)\bm{\mu}_n^\textit{md} + \alpha^\mu \hat{\bm{\mu}}_n, \\
	\bm{\mathcal{K}}^{\textit{md}}_{n+1} &= (1-\alpha^\kappa)\bm{\mathcal{K}}_n^\textit{md} + \alpha^\kappa \hat{\bm{\mathcal{K}}}_n, 
\end{aligned}
\end{equation}
which is the EMA method. It could release the reward ascent further, because the steps are selected as $ \alpha^{\mu},  \alpha^{\kappa} \in (0,1)$.  

Adam [38], a popular stochastic optimization algorithm for deep learning, upgrades the EMA method with bias correction and improves logistic regression or neural network learning performance. So we reset $\beta^\mu_n = (1-\alpha^{\mu})^n, \beta^\kappa_n = (1-\alpha^{\kappa})^n$ and $\lambda^\mu_n = 1+\beta^\mu_n, \lambda^\kappa_n = 1+\beta^\kappa_n$ 
to correct the bias brought by EMA and gain
\begin{equation}
\label{eq:bias}
\begin{aligned} 
	\bm{\mu}^{\textit{md}}_{n+1} &= {\hat{\bm{\mu}}_n} + \alpha^{\mu}\sum_{i=1}^{n-1} \frac{(\hat{\bm{\mu}}_i - \bm{\mu}^{\textit{md}}_i) - (\hat{\bm{\mu}}_n - \bm{\mu}^{\textit{md}}_n)}{(1-\alpha^{\mu})^{i-n}}, \\
	\bm{\mathcal{K}}^{\textit{md}}_{n+1} &= {\hat{\bm{\mathcal{K}}}_n} + \alpha^{\kappa}\sum_{i=1}^{n-1} \frac{(\hat{\bm{\mathcal{K}}}_i - \bm{\mathcal{K}}^{\textit{md}}_i) - (\hat{\bm{\mathcal{K}}}_n - \bm{\mathcal{K}}^{\textit{md}}_n)}{(1-\alpha^{\kappa})^{i-n}}, 
\end{aligned}
\end{equation}
according to Appendix~\ref{appdx:AMA}. The above quasi-Adam (qAdam)~\footnote{We call it qAdam because Adam~\cite{Kingma2014Adam} uses $\frac{1}{1-(\alpha^{\mu})^{n}}$ and $\frac{1}{1-(\alpha^{\kappa})^{n}}$ to correct \eqref{eq:EMA}'s bias. } method performs differently from EMA~\eqref{eq:EMA} since qAdam approximately transforms the step-size from $\beta$ to ${(1-\alpha)^n}$ with $\alpha \in (0, \frac{3-\sqrt{5}}{2})$.  It reveals that the maximum values $\{\hat{\bm{\mu}}_n, \hat{\bm{\mathcal{K}}}_n\}$ make damped efforts from the initial to the final. To look inside the above methods deeper, we  derive the convergence process of EMA~\eqref{eq:EMA_converge} and qAdam~\eqref{eq:bias_converge} in each reward-step of \textit{AstoIter} according to AGD in Section~\ref{sec:AGD} and its ODE-interpretation in Section~\ref{sec:AGD-TO}: 
\begin{gather}
	\label{eq:EMA_converge}
	\min _{n=1, \ldots, N}\left\| \nabla \mathcal{R}_{n}^{\textit{md}}\right\|^{2} \leq \frac{\mathcal{R}^{*}-\mathcal{R}_{0}}{\alpha(1-\alpha\mathcal{L}_\mathcal{R}) N}, \\
	\label{eq:bias_converge}
	\mathcal{R}^{*} - \mathcal{R}^{\textit{ag}}_{N} \leq \frac{1}{2}(1-\alpha)^{N-1}\|\bm{gp}_0 - \bm{gp}^{*}\|^2, 
\end{gather}
where $\bm{gp} = \{\bm{\mu},\bm{\mathcal{K}}\}$, $\alpha = \min\{\alpha^{\mu},\alpha^{\kappa}\}$, and $\mathcal{L}_\mathcal{R}$ is $\mathcal{R}$'s Lipschitz constant. Since ASTO sequentially gathers the environmental information and the whole learning process is strongly nonconvex and latent, we analyze the rewarding process informed by one single sampling or interaction to see how the above methods perform during the whole ASTO process. As a result, \eqref{eq:bias_converge} shows the squared convergence rate of qAdam, and Section~\ref{sec:ASTO_tune} validates that the damped process can correct the arbitrarily initialized $\mathcal{GP}$ and stabilize the final steps. However, these advantages easily collapse due to the strongly nonconvex and unobservable RL process of ASTO. In this case, \eqref{eq:bias}'s residue brings more disturbances from the historical information forgotten exponentially than \eqref{eq:EMA}. 

So another question about how to perform a damped process free of the prerequisites of convexity and observability comes up. Surprisingly, Corollary~\ref{corol:AGD1} provides a possible solution: we reset $\alpha_n = \frac{2}{n+1}, \beta = \mathcal{L}_Q^{-1}, \lambda_n \backsim \beta\cdot\mathcal{U}(1, 1+\frac{\alpha_n}{4})$. Then we gain the AMA process according to Appendix~\ref{appdx:AMA}:
\begin{equation}
\label{eq:AMA}
\begin{aligned} 
	\bm{\mu}^{\textit{md}}_{n+1} &= \bm{\mu}_n^\textit{md} + \alpha_n(\lambda_n-\beta) (\hat{\bm{\mu}}_n - \bm{\mu}_n^\textit{md}) \\[-4pt]
	&\quad + \alpha_{n-1}\Gamma_{n-1} \sum_{i=1}^{n-1} \Gamma_{i-1}^{-1} (\lambda_i-\beta)(\hat{\bm{\mu}}_i - \bm{\mu}_i^\textit{md}), \\
	\bm{\mathcal{K}}^{\textit{md}}_{n+1} &= \bm{\mathcal{K}}_n^\textit{md} + \alpha_n(\lambda_n-\beta) (\hat{\bm{\mathcal{K}}}_n - \bm{\mathcal{K}}_n^\textit{md}) \\[-4pt]
	&\quad + \alpha_{n-1}\Gamma_{n-1} \sum_{i=1}^{n-1} \Gamma_{i-1}^{-1} (\lambda_i-\beta)(\hat{\bm{\mathcal{K}}}_i - \bm{\mathcal{K}}_i^\textit{md}). 
\end{aligned}
\end{equation}
According to Corollary~\ref{corol:AGD1}, we derive AMA's convergence: 
\begin{equation} \label{eq:AMA_converge}
	\min_{k=1\dots N}\|{\nabla}\mathcal{R}^{\textit{md}}_{n}\|^2 
	\leq \frac{25\mathcal{L}_\mathcal{R} \left(\mathcal{R}^{*} - \mathcal{R}_{0} \right)}{4(\sqrt{2}-1)N}. 
\end{equation}
It shows a similar convergence process as EMA~\eqref{eq:EMA_converge}, and both of them could deal with some nonconvex cases. Unlike EMA using the constant forgetting factor $\beta$,  AMA~\eqref{eq:AMA} utilizes the factor $\alpha = \frac{2}{n+1}$ to damp the factor during RL like qAdam~\eqref{eq:bias}. Section~\ref{sec:ASTO_tune} validates its high robustness and efficiency.  

In practice, we set the maximum number $N_{\textit{Asto}} \backsim \mathcal{U}(5, 15)$ of \textit{AstoIter}, because ASTO depends on the initial steps deciding the searching area more than the upcoming steps for solution refinement mainly assigned to $\mathcal{L}$-reAGD of Section~\ref{sec:L-reAGD}. Moreover, there are four terminate conditions for ASTO: (i) one of the sampled trajectories in $\bm{\Theta}_K$ is collision-free; (ii) the renewed GP-model is feasible; (iii) the maximum iteration time $N_\textit{Asto}$ is attached; (iv) $\|\bm{\mathcal{K}}_n^\textit{md}\|$ converges to $\underline{\mathcal{K}}$tol; (v) $\bm{\mathcal{K}}_n^\textit{md}$ is negative definite. When the condition (iii), (iv), or (v) is satisfied and the policy improvement misses the forecast, we will reject the updated GP model and return the original. Otherwise, we will accept and return the learned GP based on Section~\ref{sec:gpModel}: 
\begin{equation}
\label{eq:ASTO_GP}
\begin{array}{c}
	\bm{\mu} = \bm{\mu}_{0} + (\mathbf{C}\bm{\mathcal{K}}_0)^\mathrm{T} ( \mathbf{C}\bm{\mathcal{K}}_0\mathbf{C}^\mathrm{T} + \bm{\mathcal{K}}_\epsilon )^{-1} (\bm{\theta}^{*} - \mathbf{C}\bm{\mu}_0), \\[6pt]
	\bm{\mathcal{K}} = \bm{\mathcal{K}}_0 - (\mathbf{C}\bm{\mathcal{K}}_0)^{\mathrm{T}} (\bm{\mathcal{K}}_0 + \bm{\mathcal{K}}_\epsilon )^{-1} \mathbf{C}\bm{\mathcal{K}}_0,  
\end{array}
\end{equation}
where $\bm{\theta}^{*}$ is a learned policy when ASTO is terminated, $\mathbf{C}$ is a system matrix defined by $\mathfrak{H}$~\eqref{eq:Hamiltonian_1}, and $\bm{\mathcal{K}}_\epsilon$ decides the modification rate of the returning GP. After that, AGP-STO (Algorithm~\ref{alg:AGP-STO}) will adopt the learned policy/trajectory and the modified GP to continue the MAP of trajectory. 

\section{Incrementally Optimal Motion planning via Bayes tree inference}\label{sec:iOMP}



So far, we have introduced AGP-STO for the whole trajectory optimization. However, it always requires extra computation resources because of the high-dimension and step-size inconsistency in the first-order optimization case. Our former work~\cite{Feng2021iSAGO} adopts  the methodologies of iSAM2~\cite{Kaess2012iSAM2} and iGPMP2~\cite{Dong2016GPMP2} and proposes reTraj-BTI for iSAGO to elevate the planning efficiency. However, iSAGO sometimes confronts the failure case where a significant gap exists between the feasible solution and the initial trajectory in the $\mathcal{F}$-field (Figure~\ref{fig:AGP-STO}). TrajOpt~\cite{Schulman2014SCO} manually selects a few guidance waypoints and interpolates the support waypoints to gain the initial trajectory to moderate the above concerns.  

Based on the manual selection in TrajOpt, this section proposes an incremental optimal motion planning (iOMP, Algorithm~\ref{alg:iOMP}) method for time-continuous safety. iOMP first generates an initial trajectory consisting of a few guidance waypoints via linear interpolation between the start and goal point. Then it uses AGP-STO (Algorithm~\ref{alg:AGP-STO}) iteratively in \code{iSAGO.reTrajIter()}. 
\begin{figure}[hbtp]
\begin{centering}
{\includegraphics[width=1\columnwidth]{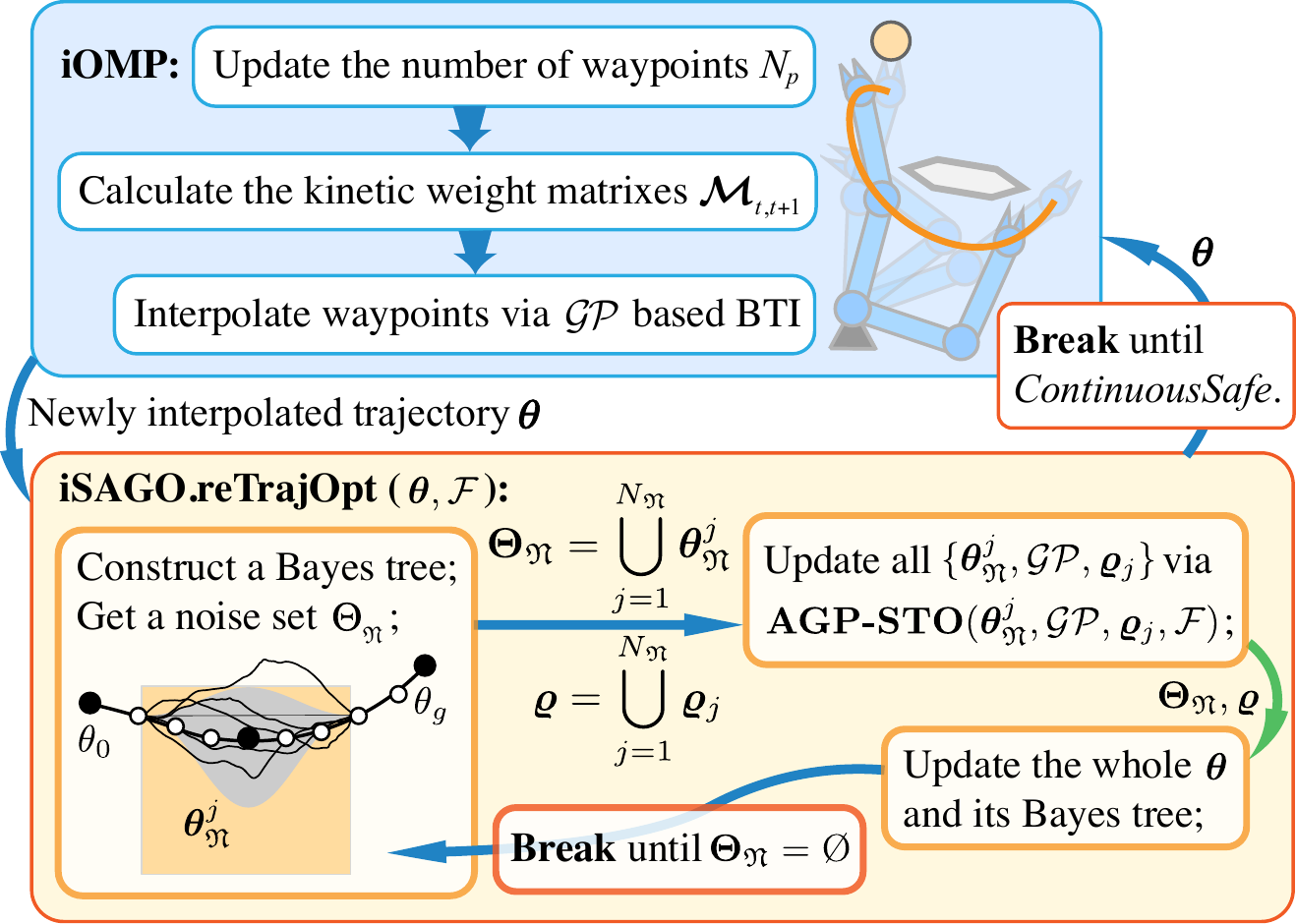}}
\par\end{centering}
\protect\caption{
A diagram shows how the incremental optimal motion planner (iOMP, Algorithm~\ref{alg:iOMP}) interpolates the waypoints among the last support waypoints via the $\mathcal{GP}$-based Bayes tree inference (BTI). Moreover, the lower part of the diagram visualizes how \code{iSAGO.reTrajOpt($\bm{\theta},\mathcal{F}$)} constructs a Bayes tree and gains a noise set $\bm{\Theta}_\mathfrak{N}$ to incrementally plan an optimal trajectory via \code{AGPSTO($\bm{\theta}_\mathfrak{N}^j, \mathcal{GP}, \bm{\varrho}_j, \mathcal{F}$)}. 
\label{fig:ITI}}
\end{figure}
\begin{algorithm}[]
\caption{iOMP}\label{alg:iOMP}
\DontPrintSemicolon
\LinesNumbered
\SetKwInOut{Input}{Input}
\SetKwFunction{reTraj}{iSAGO.reTrajIter}
\SetKwFunction{numel}{numel}
\SetKwFunction{AGPSTO}{AGPSTO}
\SetKwProg{Fn}{}{\string:}{}
\SetKw{Break}{break}
\Input {Start point ${\theta}_0$ and goal point ${\theta}_g$; }
\textbf{Initialize: }{number of waypoints $N_{p}$, time interval $\Delta t$; }\;
\For {\textit{itiIter}: $i = 1 \dots N_{\textit{iti}}$}{
	Get $\mathcal{GP}(\bm{\mu},\bm{\mathcal{K}})$ via Bayes inference from $\bm{\theta}$; \;
	Renew $\bm{\theta}$ via \eqref{eq:points_allocate}-interpolation; \;
	\Fn{\reTraj{$\bm{\theta}$,$\mathcal{F}$}}{
			$k$ += 1; Get noisy set $\bm{\Theta}_{\mathfrak{N}}=\bigcup_{j = 1}^{N_{\mathfrak{N}}}\bm{\theta}_{\mathfrak{N}}^{j}$; \;
			\lIf{$\bm{\Theta}_{\mathfrak{N}} = \O$ \textbf{or} $k \geq N_\textit{uf}$}{
			\Break; 
			}
			\For {$j = 1, \dots, N_\mathfrak{N}$}{
				$\{\bm{\theta}_\mathfrak{N}^j,\mathcal{GP},\bm{\varrho}_j\} \leftarrow \AGPSTO{$\bm{\theta}_\mathfrak{N}^j$,$\mathcal{GP}$,$\bm{\varrho}_j$,$\mathcal{F}$}$ via \textbf{Algorithm~\ref{alg:AGP-STO}}; \label{alg:ITI:AGPSTO} \;
				Update the whole $\bm{\theta}$ and Bayesian tree; \;
			}
			Eliminate the waypoints $\theta \in \bm{\Theta}_\mathfrak{N}$ temporarily; \;
	}
	\lIf {\textit{ContinuousSafe}}{
		\Return  $\bm{\theta}$. 
	}
	\lElseIf{$i \geq N_\textit{iti}$}{
		\Return Failed. 
	}
	Update $N_{p} \leftarrow {\tau_\textit{ip}} \cdot N_{p}$, $\Delta t \leftarrow {\tau_\textit{ip}^{-1}} \cdot \Delta t $; \;
}
\end{algorithm}

In the outer loop \textit{itiIter}, iOMP interpolates the support waypoints among the previously optimized trajectory via BTI. It means a proper allocation of the waypoints to interpolate can not only ensure the time-continuous safety but improve the efficiency of trajectory smoothing. So this section utilizes the kinetic energy $\mathcal{F}_{\kappa}(t,t+1)$ between any adjacent waypoints $\theta_{t}, \theta_{t+1}$ in the previous trajectory to determine the allocation. 
However, a direct calculation of the kinetic energy needs extensive computational resources for $\int_{t}^{t+1} \frac{1}{2}\mathcal{I}(t)\dot{\theta}(t)^{2}\diff t$~\footnote{$\dot{\theta}(t)$ denotes the higher derivation of the state $\theta$ with respect to time $t$. } because the inertia $\mathcal{I}(t)$ of the whole robot is time-variant. So we first infer the mid-point $\theta_{t+\frac{1}{2}}$ by GP~\eqref{eq:prior_mean}, then approximate the kinetic energy via 
\begin{equation}
\nonumber
	\mathcal{F}_{\kappa}(t,t+1) = \frac{1}{2}\|\theta_{t} - \theta_{t+1}\|_{\bm{\mathcal{M}}_{t,t+1}^{-1}}^2,
\end{equation}
with the kinetic weight matrix
\begin{equation}
	\bm{\mathcal{M}}_{t,t+1} = \text{diag}\left(\mathcal{M}_{i}\right),  i\in \mathcal{Q} = \{1,\dots,N_\mathcal{Q}\} 
\end{equation}
where $\mathcal{Q}$ stores the indexes of robot's joints. With the gathered physical information, such as the inertia $\bm{I}_\textit{zz}$, center of mass $\bm{p}_\text{CoM}$, and mass $m$ of each rigid body, we approximate the inertia resisting joint-$i$ by 
\begin{equation} 
\nonumber
	\mathcal{M}_{i} = \bm{I}_\textit{zz}^{i} + \sum_{j\in\mathfrak{B}_{i}} m_{j} \cdot \|_{i}^{j}\mathbf{T}(\theta_{t+\frac{1}{2}})\cdot\bm{p}_{\text{CoM}}^{j}\|^{2},
\end{equation}
where set $\mathfrak{B}_{i}$ stores the indexes of the bodies actuated by joint-$i$, and matrix $_{i}^{j}\mathbf{T}$ uses the midway state $\theta_{t+\frac{1}{2}}$ to transform body-$j$ from joint-$i$ in $\mathds{SE}(3)$. After that, we allocate the number of interpolation waypoints between $\theta_{t}$ and $\theta_{t+1}$ by
\begin{equation}\label{eq:points_allocate}
	N_\textit{p}^{t,t+1} = \operatorname{round}\left(N_\textit{p}\cdot \frac{\mathcal{F}_{\kappa}(t,t+1)}{\sum_{t\in[t_{0},t_{g})} \mathcal{F}_{\kappa}(t,t+1)}\right).  
\end{equation}
It means that there are more waypoints in the interval with relatively higher kinetic energy. 



In the inner loop \code{iSAGO.reTrajIter($\bm{\theta},\mathcal{F}$)}~\cite{Feng2021iSAGO}, iOMP decomposes the whole trajectory optimization into a series of sub-replanning problems to improve the computation efficiency and success rate. This iteration first gains the noisy set $\bm{\Theta}_\mathfrak{N}$ consisted of $N_\mathfrak{N}$ sub-trajectories $\bm{\theta}_\mathfrak{N}$ with significant Bayes tree factor. Then it uses AGP-STO to optimize $\bm{\theta}_\mathfrak{N}$ with its corresponding penalty factor $\bm{\varrho}$ incrementally and updates the Bayes tree informed by the GP and constraints $\mathcal{G}$. When all the significant factors are flattened (i.e., $\bm{\Theta}_\mathfrak{N} = \O$), it will terminate. To prevent the informational leakage of $\mathcal{G}$, iOMP exponentially expands the number $N_\textit{p}$ of the waypoints to interpolate until the trajectory is \textit{ContinuousSafe}.

\section{Implementation details}\label{sec:details}

%
\begin{figure*}[hbtp]
\centering
\hspace*{\fill}
	\begin{subfigure}[b]{0.25\textwidth}		
		\centering
		\includegraphics[width=1\linewidth]{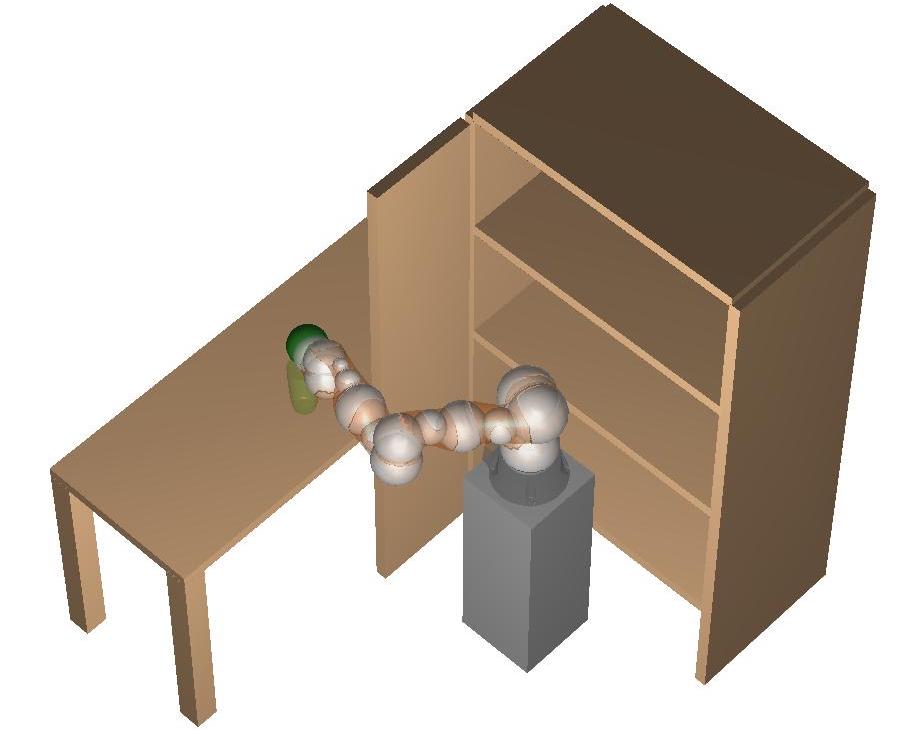}
		\caption{LBR-iiwa grabs a cup on a desktop. }
		\label{fig:CCB_iiwa}
	\end{subfigure}	
	\hfill
	\begin{subfigure}[b]{0.25\textwidth}		
		\centering
		\includegraphics[width=1\linewidth]{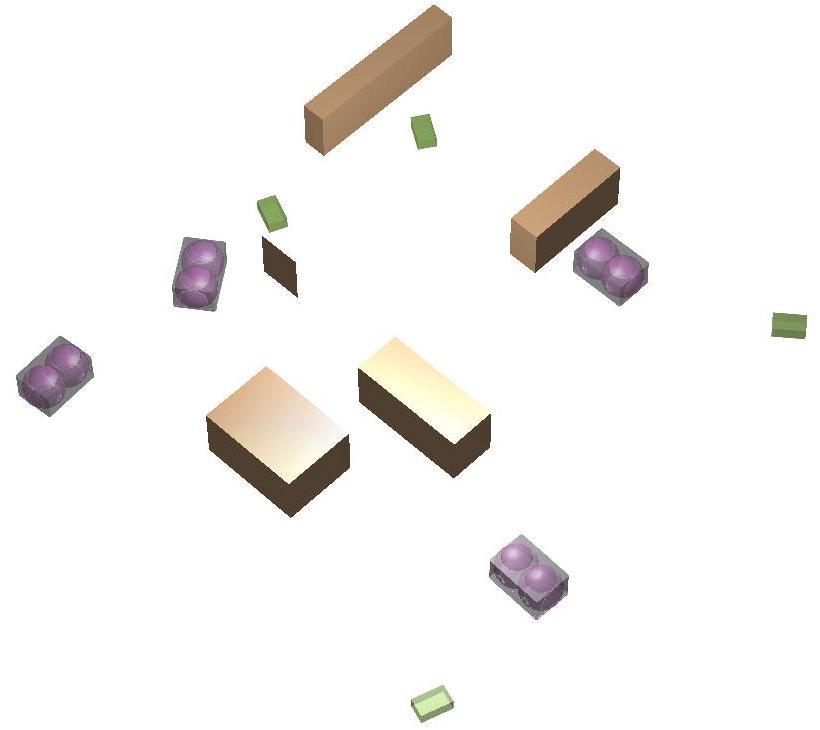}
		\caption{4 AGVs grab some boxes in a warehouse. }
		\label{fig:CCB_AGVs}
	\end{subfigure}
	\hfill
	\begin{subfigure}[b]{0.25\textwidth}
		\centering
		\includegraphics[width=1\linewidth]{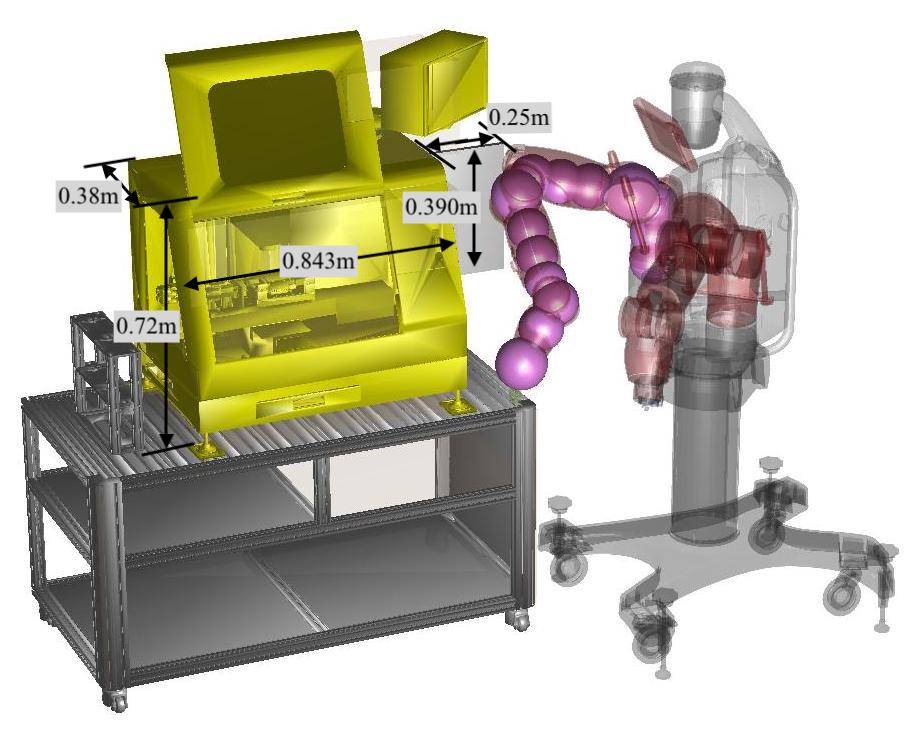}
		\caption{Baxter grabs a workpiece in a machine. }
		\label{fig:CCB_Baxter}
	\end{subfigure}
\hspace*{\fill}
\protect\caption{
The geometrically connected collision check balls (CCBs) assemble the robot. 
\label{fig:CCB}}
\end{figure*}
%


\subsection{GP prior}\label{sec:const_gp_prior}
This paper utilizes the LTV-SDE with constant velocity (i.e., zero acceleration) for the GP prior construction through~\eqref{eq:LTV-SDE} according to~\cite{Mukadam2018GPMP}: 
\begin{equation}\nonumber 
	\mathbf{A}(t) = \begin{bmatrix} \mathbf{0} & \mathbf{I} \\ 
	\mathbf{0} & \mathbf{0} \end{bmatrix}, \mathbf{u}(t) = \mathbf{0}, 
	\mathbf{F}(t) = \begin{bmatrix} \mathbf{0} \\ \mathbf{I} \end{bmatrix},
\end{equation}
and given $\Delta t_i = t_{i+1} - t_{i}$,
\begin{equation}\label{eq:Phi-Qc01}
\small
\mathbf{\Phi}(t,s) = 
\begin{bmatrix} 
	\mathbf{I} & (t-s)\mathbf{I} \\ 
	\mathbf{0} & \mathbf{I} 
\end{bmatrix},
\mathbf{Q}_{t,t+1} = 
\begin{bmatrix} 
	\frac{1}{3} \Delta t^3 & \frac{1}{2} \Delta t^2  \\ 
	\frac{1}{2} \Delta t^2 & \Delta t  
\end{bmatrix} \otimes \mathbf{Q}_c.
\end{equation}

In practice, the ASTO method in Section~\ref{sec:ASTO} only samples the trajectories consisted of position states under the zero velocity assumption and utilizes the discretization method $\frac{x(t_{i+1}) - x(t_{i})}{\Delta t}$ rather than the analytic method requiring Jacobian to calculate $\dot{x}(t_i)$ more efficiently for $\mathcal{F}$-estimation.

\subsection{Collision-cost function}\label{sec:collisionCost}


Some previous works~\cite{Zucker2013CHOMP, Schulman2013SCO, Schulman2014SCO, Liu2016SafetyMeasure} are almost first-order continuously differentiable. It always requires extra $\mathcal{L}_{\mathcal{F}}$-estimation steps in Section~\ref{sec:L-reStart} because of the second-order discontinuity. Our work designs a collision cost function with the forth-order polynomial piecewise form:  
%
\begin{equation} \label{eq:collisonCost}
	c(\mathcal{D})=
	\begin{cases} 
	\hspace*{\fill} {\frac{\epsilon}{2}-\mathcal{D}} \hspace*{\fill} & {\text{ if } \mathcal{D} < 0} \\
	\hspace*{\fill} {\frac{1}{\epsilon^2}(\epsilon-\mathcal{D})^3-\frac{1}{2\epsilon^3}(\epsilon-\mathcal{D})^4} \hspace*{\fill} & {\text { if } 0 \leq \mathcal{D} \leq \epsilon} \\ 
	\hspace*{\fill} {0} \hspace*{\fill} & {\text { otherwise, } } \\
	\end{cases} 
\end{equation}
where $\mathcal{D}\left(\bm{x}\right): \Re^{3} \rightarrow \Re$ maps the CCB-position in $SE(3)$ into the singed distance, and $\bm{x}(\theta, S_{i}): \Re^{N_{\bm{\theta}}} \rightarrow \Re^{3}$ maps the joint states into the position of bounding box $S_{i} \in {\mathcal{B}}$. The collision cost function above is a quartic polynomial and second continuously differentiable. 
Figure~\ref{fig:localmin} visualizes an example of the collision field defined by~\eqref{eq:collisonCost}, while Figure~\ref{fig:CCB} visualizes three types of robot represented by a series of CCBs in three planning scenarios.

\subsection{Motion constraints}\label{sec:motionConstraint}
As for the motion constraints, we uses the same cost function as~\cite{Mukadam2018GPMP}:
\begin{equation}\label{eq:motionConstraintCost}
	{c}(\theta_i^{d}) =
	\begin{cases} 
	      \hspace*{\fill} -\theta_i^{d} +\theta_{\text{min}}^{d} - \epsilon \hfill & \text{if} \ \theta_i^{d} < \theta_{\text{min}}^{d} + \epsilon_d  \\	      
	      \hspace*{\fill} \theta_i^{d} -\theta_{\text{max}}^{d} + \epsilon  & \text{if} \ \theta_i^{d} > \theta_{\text{max}}^{d} - \epsilon_d  \\
	      \hspace*{\fill} \bm{0} \hspace*{\fill} & \text{otherwise,} \  \\ 
	\end{cases}
\end{equation}
where $\theta_{i}^{d}$ is the $d^{\text{th}}$ joint state at time $i$, $\theta_{\text{min}}^{d}$ and $\theta_{\text{max}}^{d}$ are the lower and upper joint motion limits respectively, and $\epsilon_d$  aims to bound the joint states within a safe area.

\section{Evaluation}\label{sec:evaluation}

\begin{figure*}[hbt]
	\begin{centering}
		\begin{subfigure}[b]{0.16\textwidth}
			\centering
			\includegraphics[width=1\linewidth]{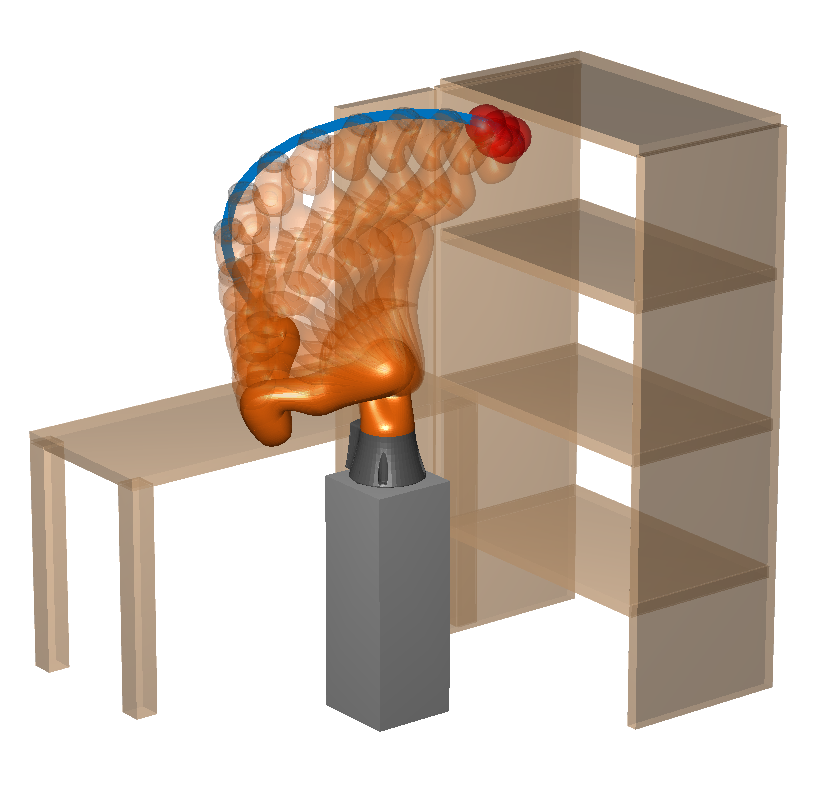}
			\caption{Task A-1, $0.08 | 0.00$}
		\end{subfigure}
		\begin{subfigure}[b]{0.16\textwidth}
			\centering
			\includegraphics[width=1\linewidth]{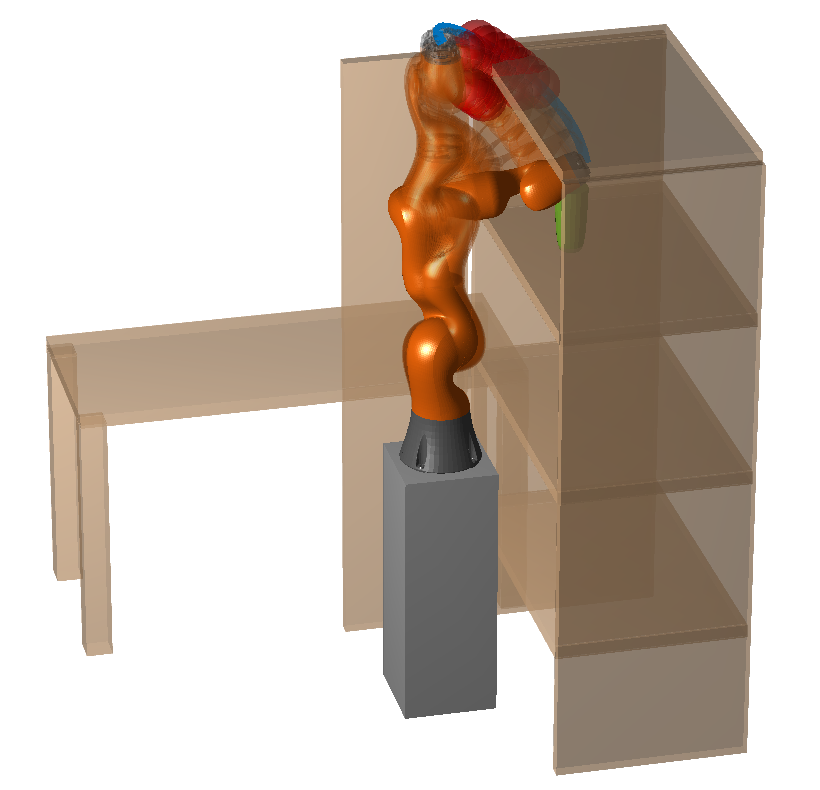}
			\caption{Task A-2, $0.15 | 0.59$}
		\end{subfigure}
		\begin{subfigure}[b]{0.16\textwidth}
			\centering
			\includegraphics[width=1\linewidth]{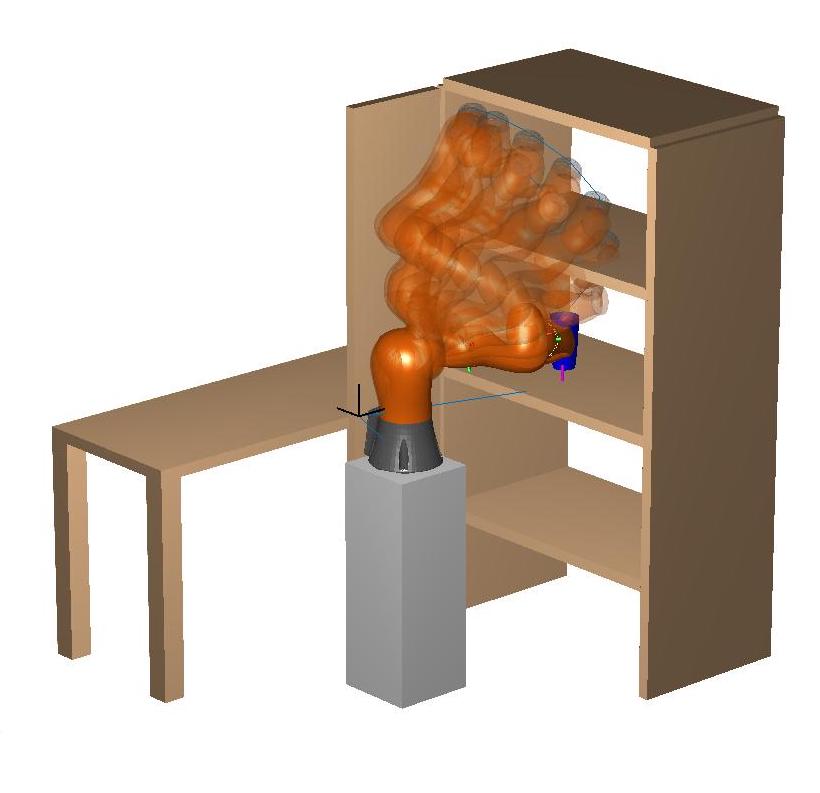}
			\caption{Task B-1, $0.26 | 0.47$}
		\end{subfigure}
		\begin{subfigure}[b]{0.16\textwidth}
			\centering
			\includegraphics[width=1\linewidth]{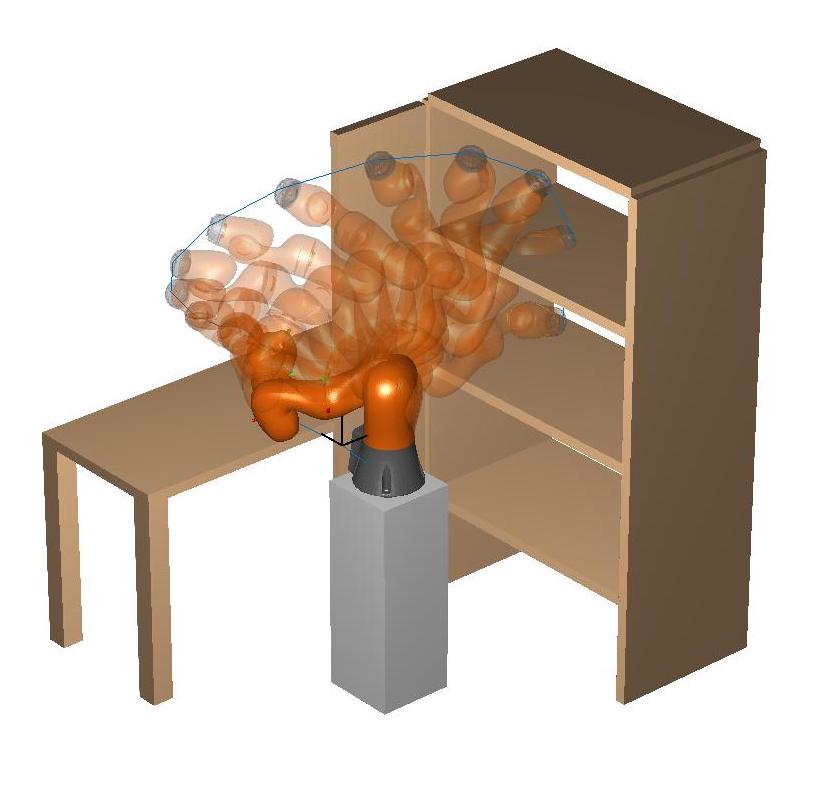}
			\caption{Task B-2, $0.44 | 0.65$}
		\end{subfigure}
		\begin{subfigure}[b]{0.16\textwidth}
			\centering
			\includegraphics[width=1\linewidth]{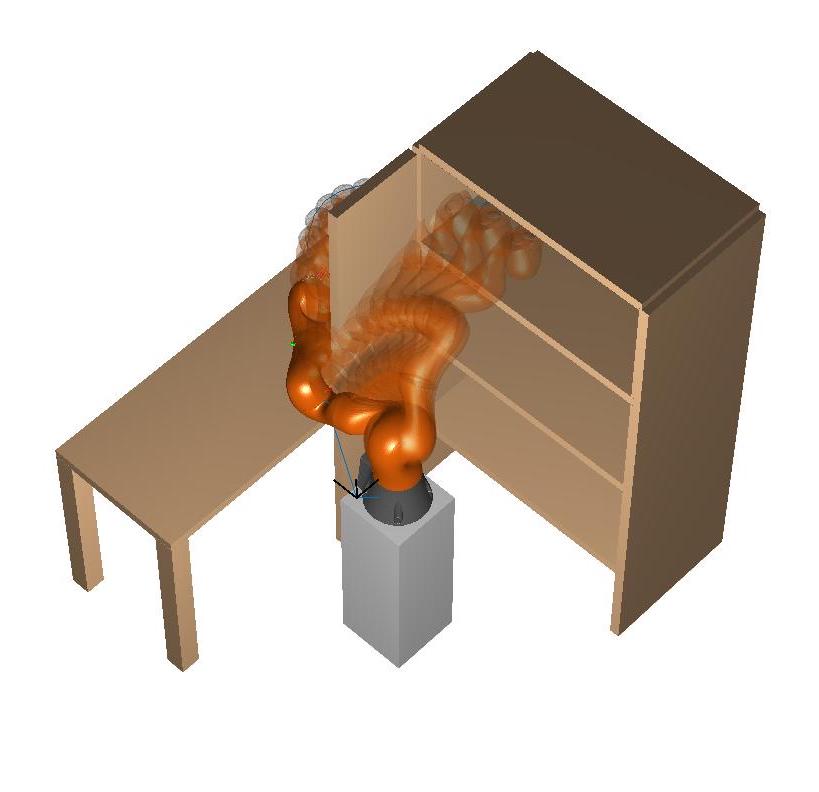}
			\caption{Task B-3, $1.33 | 0.88$}
		\end{subfigure}
		\begin{subfigure}[b]{0.16\textwidth}
			\centering
			\includegraphics[width=1\linewidth]{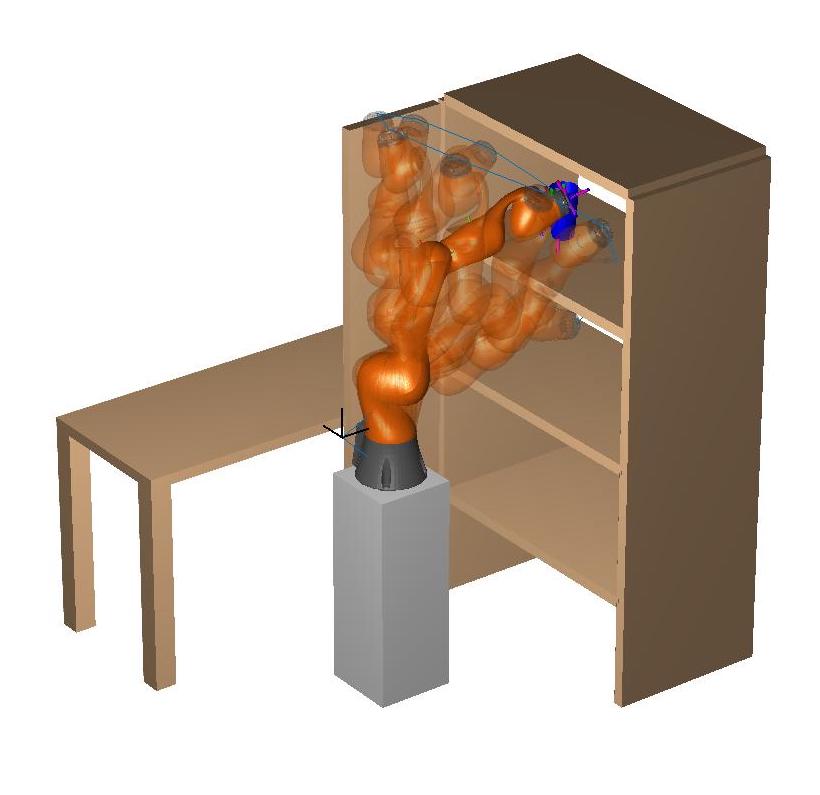}
			\caption{Task B-4, $0.46 | 0.94$}
		\end{subfigure}
		\begin{subfigure}[b]{0.16\textwidth}
			\centering
			\includegraphics[width=1\linewidth]{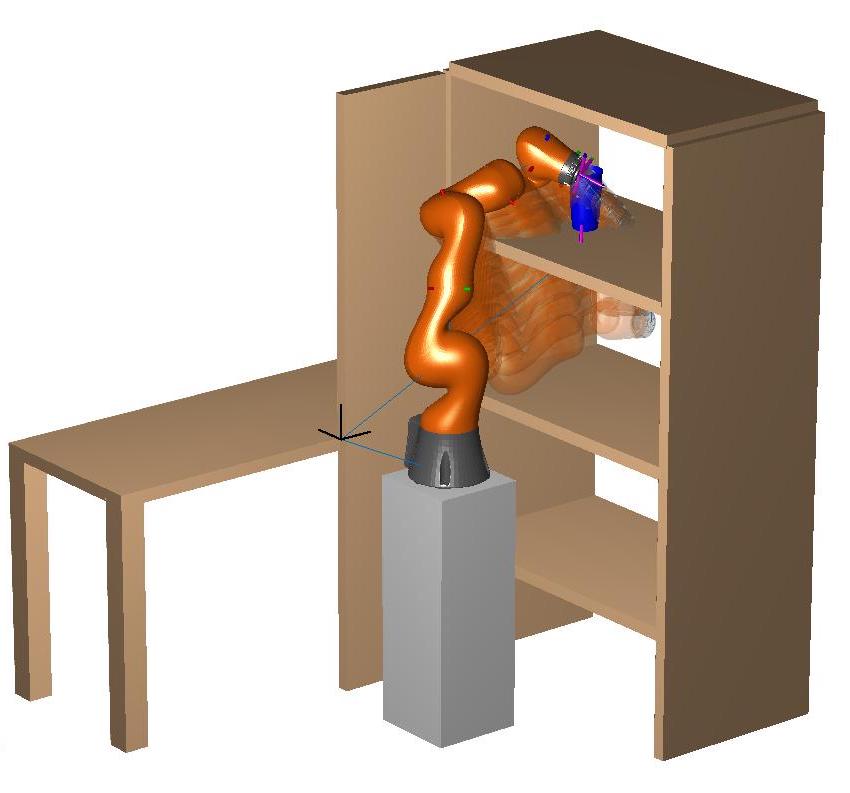}
			\caption{Task C-1, $0.64 | 1.00$}
		\end{subfigure}
		\begin{subfigure}[b]{0.16\textwidth}
			\centering
			\includegraphics[width=1\linewidth]{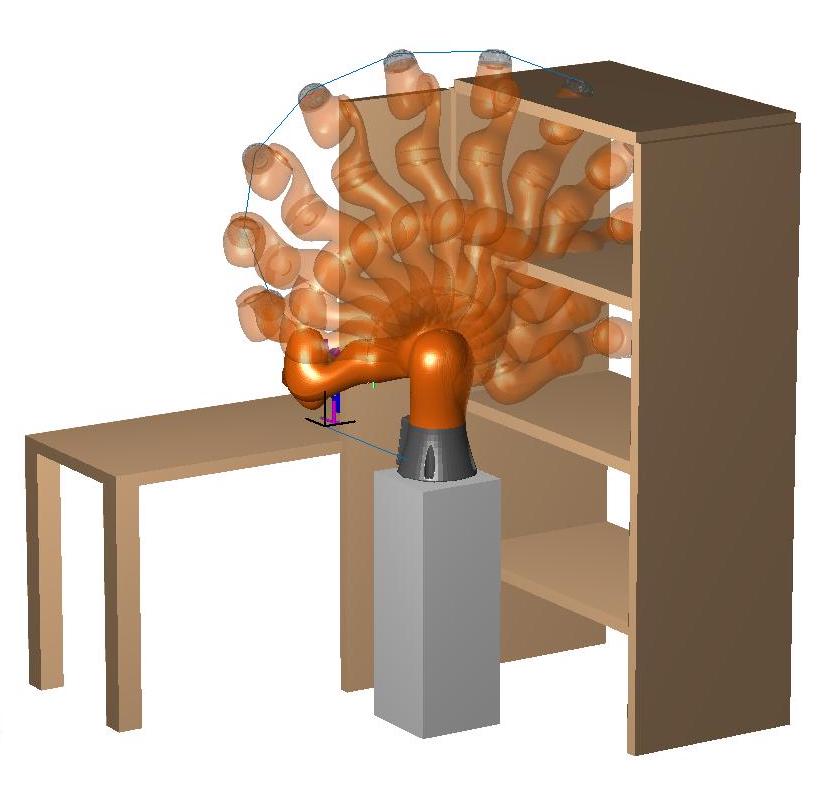}
			\caption{Task C-2, $1.02 | 0.94$}
		\end{subfigure}
		\begin{subfigure}[b]{0.16\textwidth}
			\centering
			\includegraphics[width=1\linewidth]{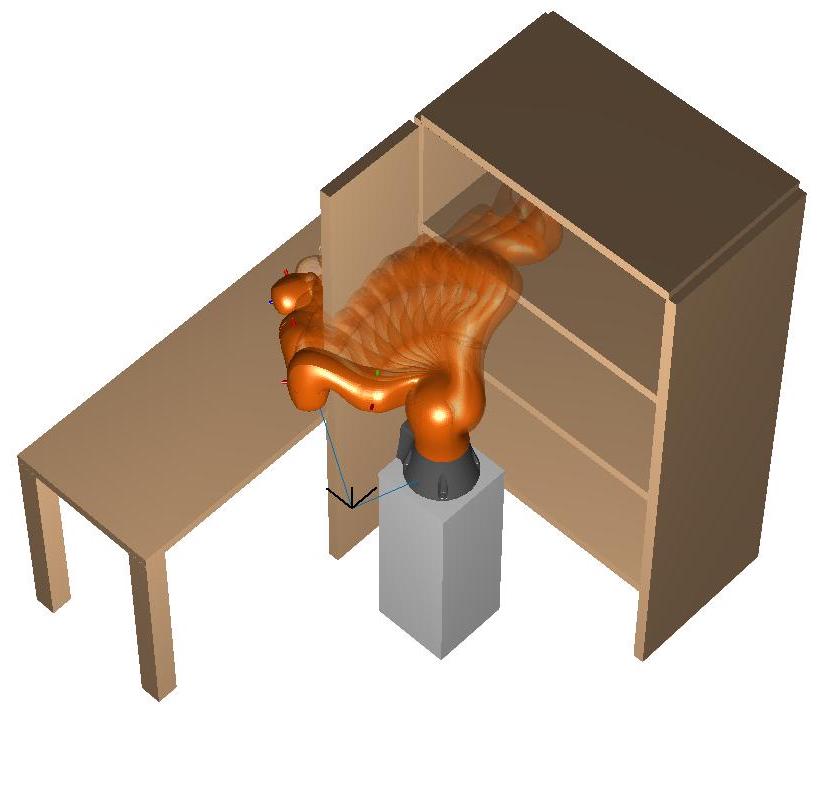}
			\caption{Task C-3, $1.01 | 1.00$}
		\end{subfigure}
		\begin{subfigure}[b]{0.16\textwidth}
			\centering
			\includegraphics[width=1\linewidth]{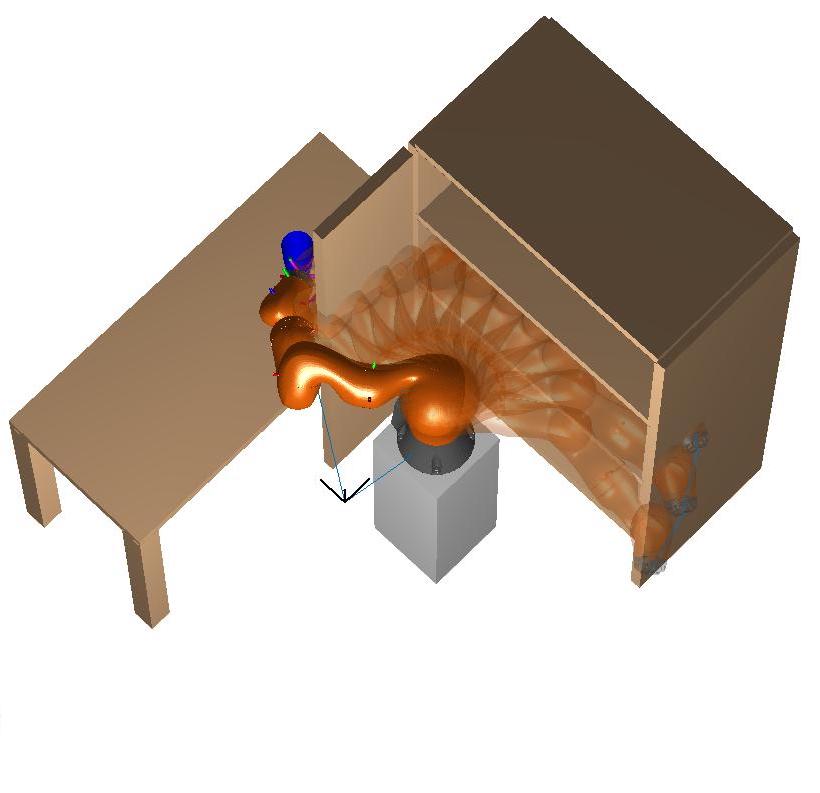}
			\caption{Task C-4, $1.44 | 0.94$}
		\end{subfigure}
		\begin{subfigure}[b]{0.16\textwidth}
			\centering
			\includegraphics[width=1\linewidth]{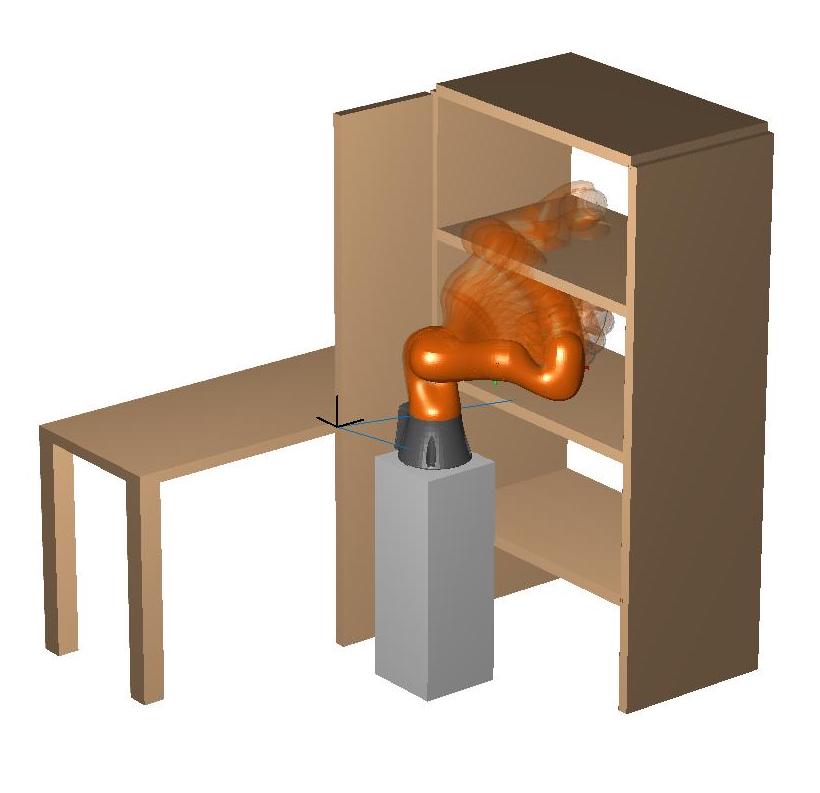}
			\caption{Task C-5, $0.81 | 1.00$}
		\end{subfigure}
		\begin{subfigure}[b]{0.16\textwidth}
			\centering
			\includegraphics[width=1\linewidth]{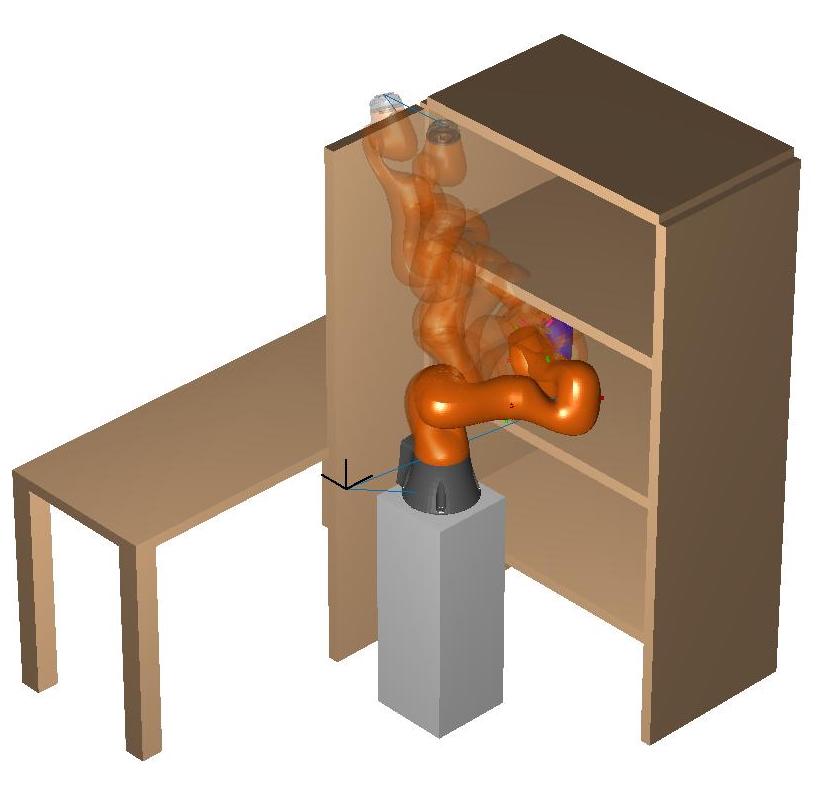}
			\caption{Task C-6, $1.95 | 1.00$}
		\end{subfigure}
	\end{centering}
	\caption{
	The initial trajectory with the red boxes in collision demonstrates the task categorized in Table~\ref{tab:probClassify}. Note that \textit{Task A-1, 0.04\textbar0.15} means task 1 of class A, whose $\bar{\mathcal{F}} | \mathfrak{S} = 0.04|0.15$ due to \eqref{eq:probClassify}. }
	\label{fig:problems_ABC}
\end{figure*}

We evaluate iAGP-STO on multi-AGV collaboration, LBR-iiwa manipulation, and rethink-Baxter assistance with 10, 50, and 5 planning problems, respectively, in three scenarios (Figure~\ref{fig:CCB}). Our experiment first tunes their key parameters on LBR-iiwa in Section~\ref{sec:paras_tune} to show the contribution of each part of iAGP-STO. It then benchmarks iAGP-STO against $\mathcal{L}$-reAGD, AGP-STO, TrajOpt, GPMP2, CHOMP, RRT-Connect, and STOMP for cross-validation in Section~\ref{sec:analysis}. After that, Section~\ref{sec:AGVs} and Section~\ref{sec:baxter} implement iAGP-STO on multi-AGV and rethink-Baxter. 




\subsection{Setup for planning benchmark}\label{sec:setup}

Our experiment compares iAGP-STO with the numerical planners: CHOMP~\cite{Zucker2013CHOMP}, TrajOpt~\cite{Schulman2013SCO,Schulman2014SCO}, and GPMP2~\cite{Mukadam2018GPMP}, as well as the sampling planners: RRT-Connect \cite{Kuffner2000RRT-connect} and STOMP~\cite{Kalakrishnan2011STOMP}. Since all of our experiments are evaluated in MATLAB, we encode GPMP2 with option \code{GPMP2\_BUILD\_MATLAB\_TOOLBOX = ON} to utilize \code{Batch\-Traj\-Optimize\-3D\-Arm} and use \code{manipulator\-RRT}~\footnote{See more at \url{https://www.mathworks.com/help/robotics/ref/manipulatorrrt.html}. } with $\code{Max\-Connection\-Distance} = \frac{\|\theta_g-\theta_0\|}{N}$ for RRT-Connect. Since TrajOpt internally uses Gurobi for the trust-region method, our benchmark uses \code{fmincon}~\footnote{See more at \url{https://www.mathworks.com/help/optim/ug/minimization-with-equality-constraints.html?searchHighlight=trust-region-reflective&s_tid=srchtitle_trust-region-reflective_5}. } with \code{optim\-options(\-'fmin\-con','Algorithm','trust\--region\--reflective', 'Specify\-Objective\-Gradient',true)}, specifying \code{Objective\-Gradient} analytically and \code{Aeq} as a matrix consisted of the active constraint gradients calculated via numerical differentiation like TrajOpt. As for CHOMP, our benchmark defines an \code{hmcSampler} object with logarithm \textit{pdf} (logpdf) defined by \eqref{eq:chomp_cost} and uses \code{hmcSampler.drawSamples} for HMC~\footnote{See more at \url{https://www.mathworks.com/help/stats/hamiltoniansampler.drawsamples.html}. }. The above MATLAB functions are encoded into C-MEX files, because iAGP-STO is developed on {GPMP2} and {GTSAM}~\cite{Dellaert2012GTSAM} and encoded via CMake. As for STOMP, we only use {ASTO} with $\{\alpha_n^{\mu},\alpha_n^{\kappa}, \beta_n^\mu, \beta_n^\kappa, \lambda_n^{\mu}, \lambda_n^{\kappa} \} = \{\bm{\mathcal{K}}_n^\textit{md}, \mathbf{I}, \mathbf{I}, \mathbf{0}, \mathbf{0}, \mathbf{0}\}$ for the sequential resampling and updating of noise-trajectory. Furthermore, all benchmarks are run on a single thread of 2.3 GHz Intel Core i9 with 16 GB 2667 MHz DDR4.


Since a large number of waypoints could prevent the leakage of collision information $\mathcal{G}$ while requires a large amount of computational resource, our benchmark first selects the initial number of waypoints 
\begin{equation}\label{eq:N_p0}
	N_{p} =
	\begin{cases} 
	      \hfill 1 \text{ or } 2 \hfill & \text{if } 0 < \frac{\|\theta_g-\theta_0\|}{\|\theta_\text{max}-\theta_\text{min}\|} \leq \frac{1}{2} \\
	      \hfill 3 \text{ or } 4 \hfill & \text{if } \frac{1}{3} < \frac{\|\theta_g-\theta_0\|}{\|\theta_\text{max}-\theta_\text{min}\|} \leq 1, \\
	\end{cases}
\end{equation}
depending on the length $\|\theta_g-\theta_0\|$ between the initial waypoint $\theta_0$ and goal waypoint $\theta_g$. The overlapping among different $N_{p}$ exists because some long feasible trajectories have a small value of $\|\theta_g-\theta_0\|$. Then we generate the initial trajectory consisting of $N_{p}$ support waypoints with 8 interval states through Bayes inference~\cite{Mukadam2018GPMP} according to Section~\ref{sec:const_gp_prior} rather than linear interpolation~\cite{Kalakrishnan2011STOMP, Zucker2013CHOMP, Schulman2013SCO, Schulman2014SCO}. After that, ITI plans an optimal motion incrementally until the trajectory is \textit{ContinuousSafe}. According to a large amount of experiments, the final feasible trajectory contains 12 to 20 support waypoints with 8 interval waypoints. In this way, our benchmark sets 16 support states with 8 interval states (152 states in total) for GPMP2, 152 support states for CHOMP and STOMP, 76 support states for TrajOpt, and $\code{Max\-Connection\-Distance} = \frac{\|\theta_g-\theta_0\|}{76}$ for RRT-connect. Except for the above settings, our benchmark uses the default settings of other planners.  


To show the competence of iAGP-STO in solving OMP problems, we conduct $50$ experiments with different start and goal states inside the desktop and shelf (Figure~\ref{fig:CCB_iiwa}). The experiments are classified into 3 categories (A, B, C). Planning difficulty increases with the stuck cases~\cite{Feng2021iSAGO} of initial trajectory because the feasible T-space shrinks when the surrounding obstacles tighten and restrain the robot's motion. For an appropriate classification, we first estimate 
\begin{equation}\label{eq:probClassify}
	\bar{\mathcal{F}} = {\mathcal{F}(\bm{\theta})}/{N_{p}}, \makebox[1em]{}
	\mathfrak{S} =  \max_{t = 1\dots N_{p}} {N_\mathfrak{S}^t}/{17}, 
\end{equation}
where $N_\mathfrak{S}^t$ denotes the number of stuck cases on $t$-th support waypoint~\footnote{Since our benchmark sets 8 interval waypoints for continuous safety, it is reasonable to apply $2\times8+1 = 17$ for stuck-case estimation. }. Then we classify the problems according to Table~\ref{tab:probClassify},
\begin{table}[hbt]
\caption{Planning Problem Classification}
\label{tab:probClassify}
\centering
\scalebox{1}{
\begin{tabular}{ccccccccccc}
\toprule 
& Class A & Class B & Class C  \\ 
\midrule
$\bar{\mathcal{F}}$  & (0, 0.18]  & (0.18, 1.40)  & (0.60, $+\infty$)  \\
$\mathfrak{S}$ & [0, 0.53) & (0.41, 0.88] & (0.88, 1.00] \\
\bottomrule
\end{tabular}}
\end{table} 
and generate 3 tasks for class A, 5 tasks for class B and 8 tasks for class C. Figure~\ref{fig:problems_ABC} shows that each problem type is a specific grabbing task with a previously specified goal requirement and contains 3 or 4 sub-problems with a random collision-free initial state and preset goal state. Moreover, we can directly observe that the number of collision cases denoted by the red boxes in Figure~\ref{fig:problems_ABC} increases from class A to class C. 
Since iAGP-STO, RRT-connect, STOMP, and CHOMP utilize the sampling method to search for feasible trajectory, we conduct 5 repeated trials for each problem to gain reliable experiment results.

\subsection{Key parameter tuning}\label{sec:paras_tune}

Section~\ref{sec:AGP-STO} shows how AGP-STO (Algorithm~\ref{alg:AGP-STO}) infers a collision-free trajectory from $\mathcal{GP}$ prior and environmental information $\mathcal{G}$ via integrating $\mathcal{L}$-reAGD and ASTO. So this section shows how their performance varies in corresponding to parameter variation in Section~\ref{sec:L-reAGD_tune} and Section~\ref{sec:ASTO_tune}.

\subsubsection{$\mathcal{L}$-reAGD parameter} \label{sec:L-reAGD_tune}

Since $\mathcal{L}$-reAGD adaptively restarts the AGD process with the reestimated $\mathcal{L}_\mathcal{F}$, this section tunes $\{\vartheta_1, \vartheta_2\}$ around $\{2, 0.25\}$ according to Corollary~\ref{corol:AGD1} and \cite{Ghadimi2016AG-NLP}, and compares it with the AGD with a constant $\mathcal{L}_\mathcal{F}$=1e2 (AGD-1e2). Since they are designed for the semi-convex problem, we select 10 problems from class A \& B for the tuning experiment and analyze the results statistically. 

\begin{figure*}[hbt]
	\begin{centering}
		\hfill
		\begin{subfigure}[h]{0.19\textwidth}
			\centering
			\includegraphics[width=1\linewidth]{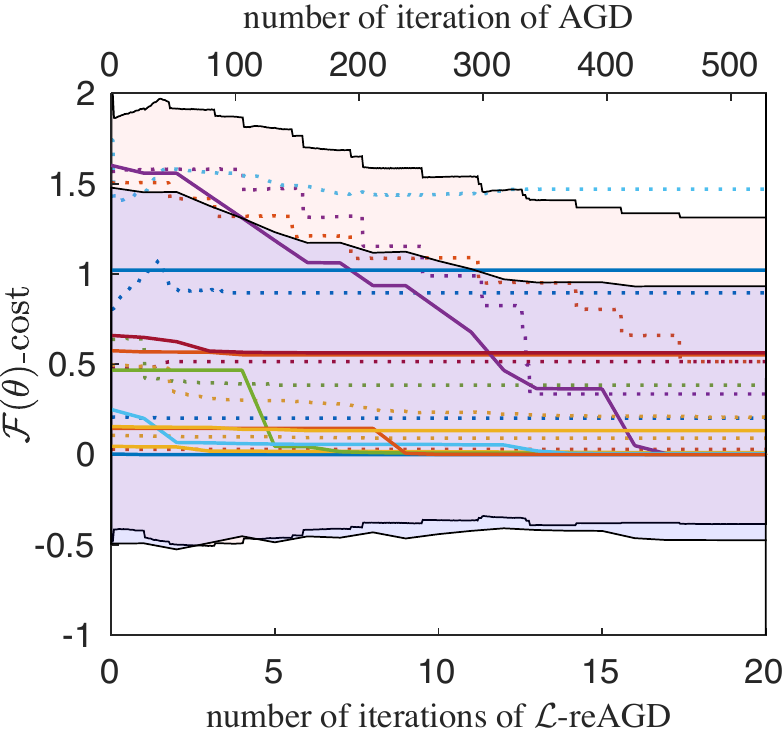}
			\caption{$\vartheta_1 = 1.000$, $\vartheta_2 = 0.000$}
		\end{subfigure}
		\hfill
		\begin{subfigure}[h]{0.19\textwidth}
			\centering
			\includegraphics[width=1\linewidth]{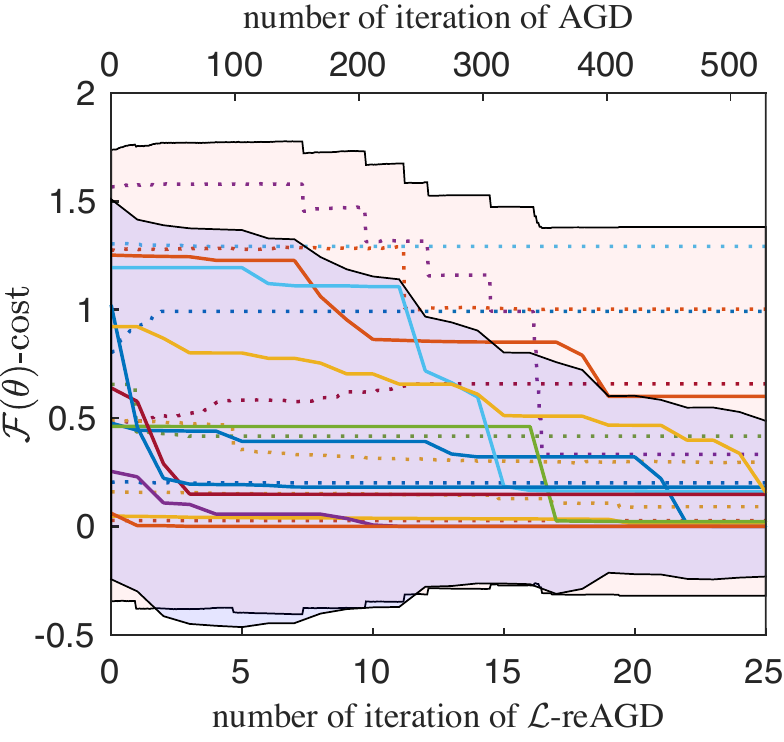}
			\caption{$\vartheta_1 = 1.414$, $\vartheta_2 = 0.125$}
		\end{subfigure}
		\hfill
		\begin{subfigure}[h]{0.19\textwidth}
			\centering
			\includegraphics[width=1\linewidth]{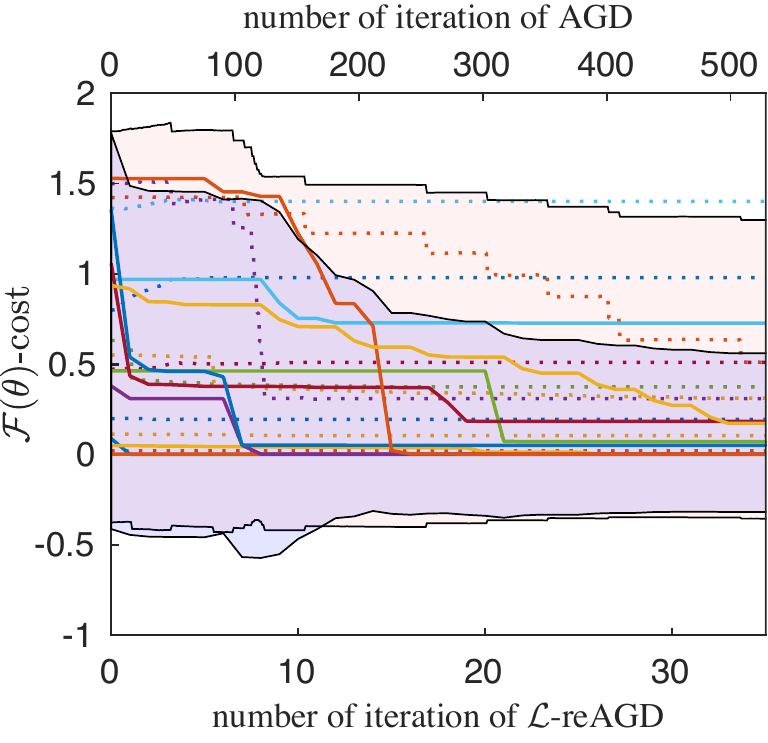}
			\caption{$\vartheta_1 = 2.000$, $\vartheta_2 = 0.125$}
		\end{subfigure}
		\hfill
		\begin{subfigure}[h]{0.19\textwidth}
			\centering
			\includegraphics[width=1\linewidth]{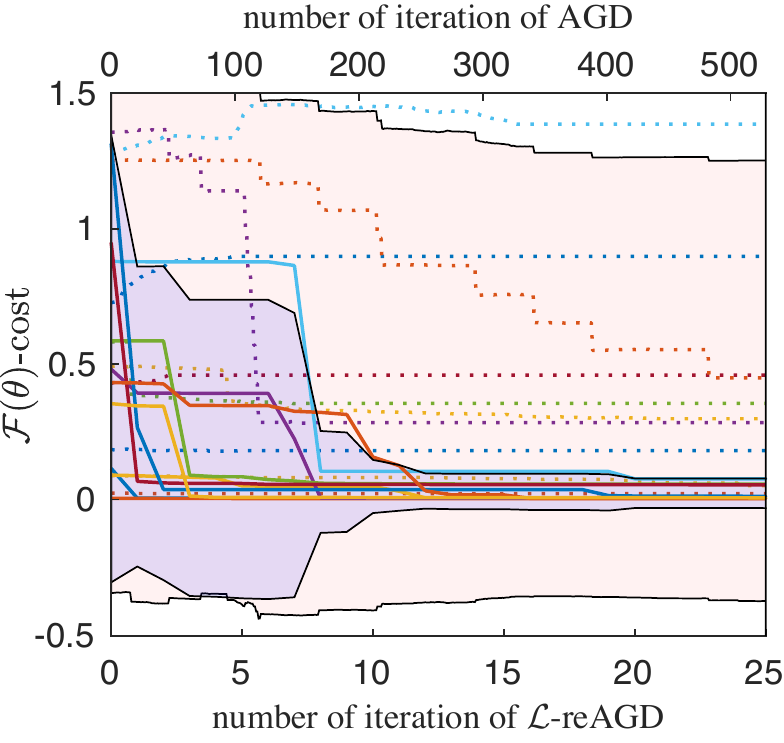}
			\caption{$\vartheta_1 = 2.000$, $\vartheta_2 = 0.250$}
		\end{subfigure}
		\hfill
		\begin{subfigure}[h]{0.19\textwidth}
			\centering
			\includegraphics[width=1\linewidth]{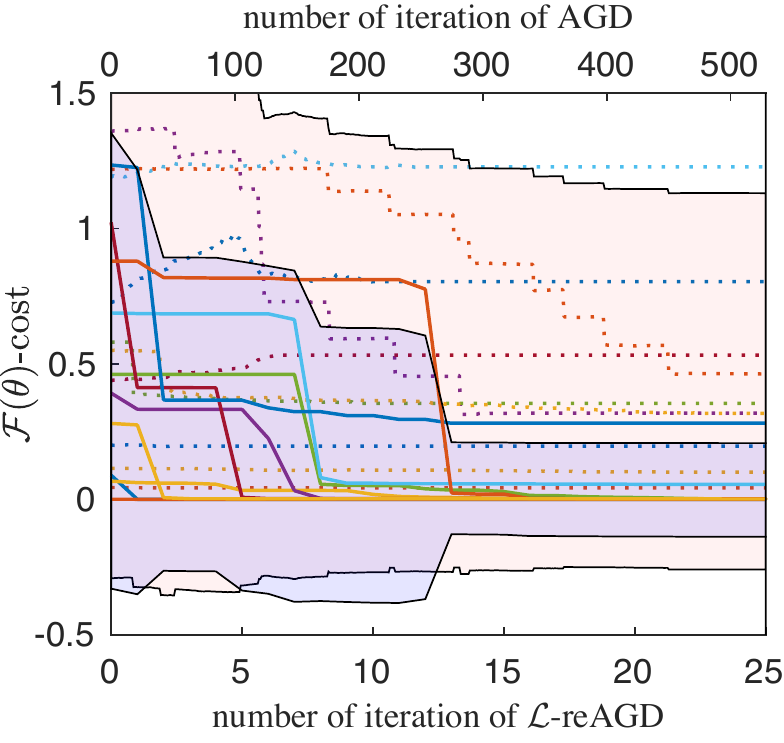}
			\caption{$\vartheta_1 = 2.000$, $\vartheta_2 = 0.500$}
		\end{subfigure}
		\hfill
		\\[4pt]
		\hfill
		\begin{subfigure}[h]{0.19\textwidth}
			\centering
			\includegraphics[width=1\linewidth]{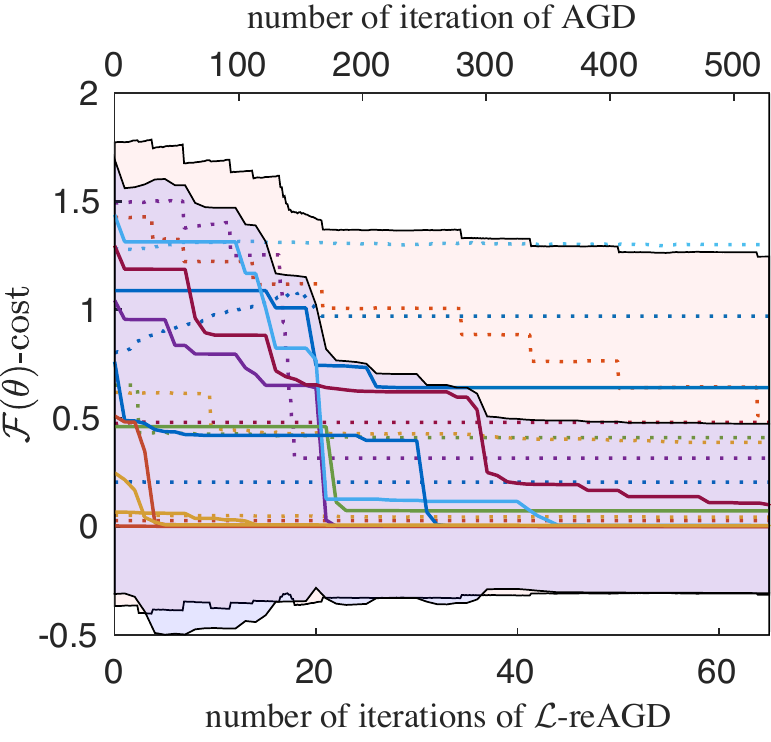}
			\caption{$\vartheta_1 = 2.828$, $\vartheta_2 = 0.125$}
		\end{subfigure}
		\hfill
		\begin{subfigure}[h]{0.19\textwidth}
			\centering
			\includegraphics[width=1\linewidth]{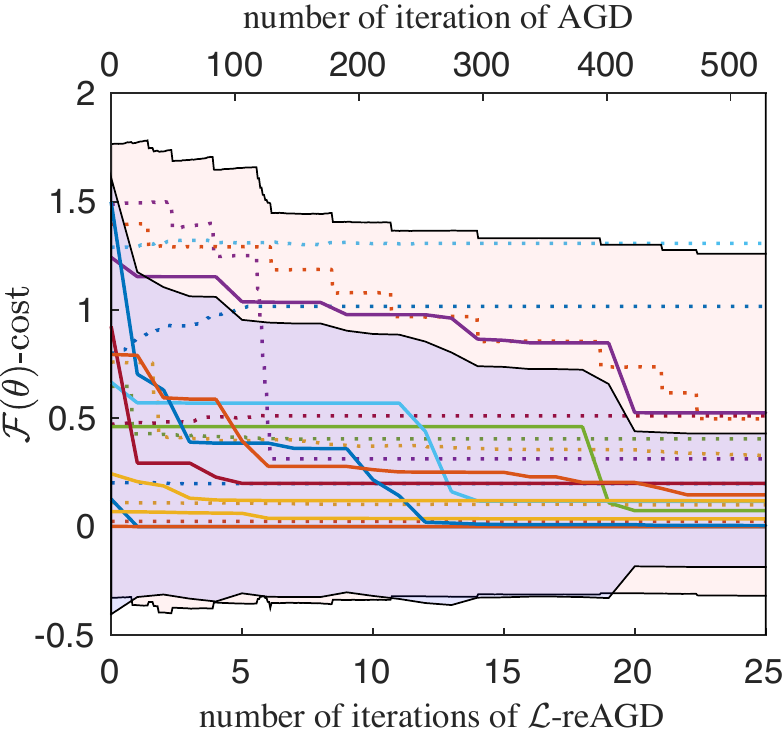}
			\caption{$\vartheta_1 = 2.828$, $\vartheta_2 = 0.375$}
		\end{subfigure}
		\hfill
		\begin{subfigure}[h]{0.19\textwidth}
			\centering
			\includegraphics[width=1\linewidth]{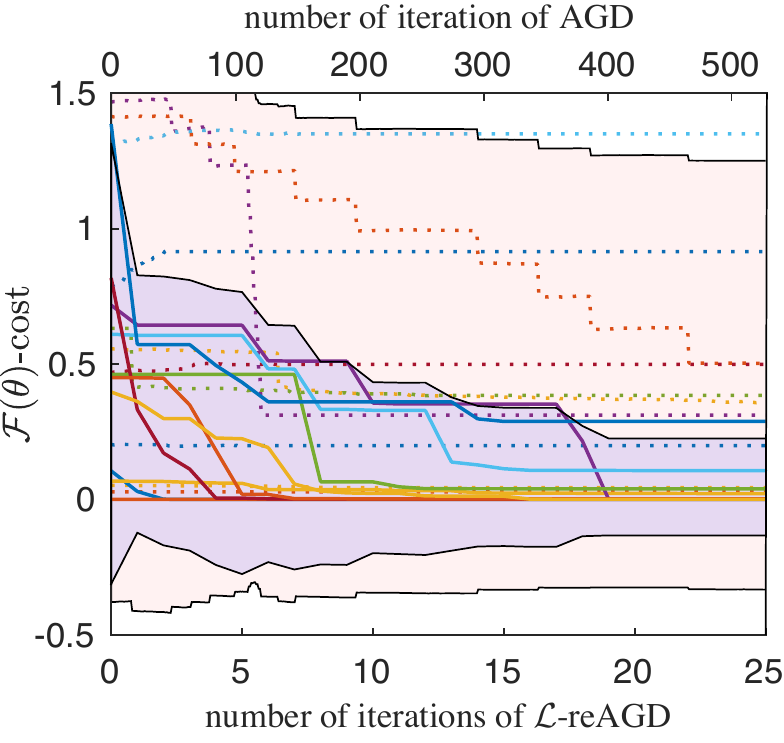}
			\caption{$\vartheta_1 = 2.828$, $\vartheta_2 = 0.875$}
		\end{subfigure}
		\hfill
		\begin{subfigure}[h]{0.19\textwidth}
			\centering
			\includegraphics[width=1\linewidth]{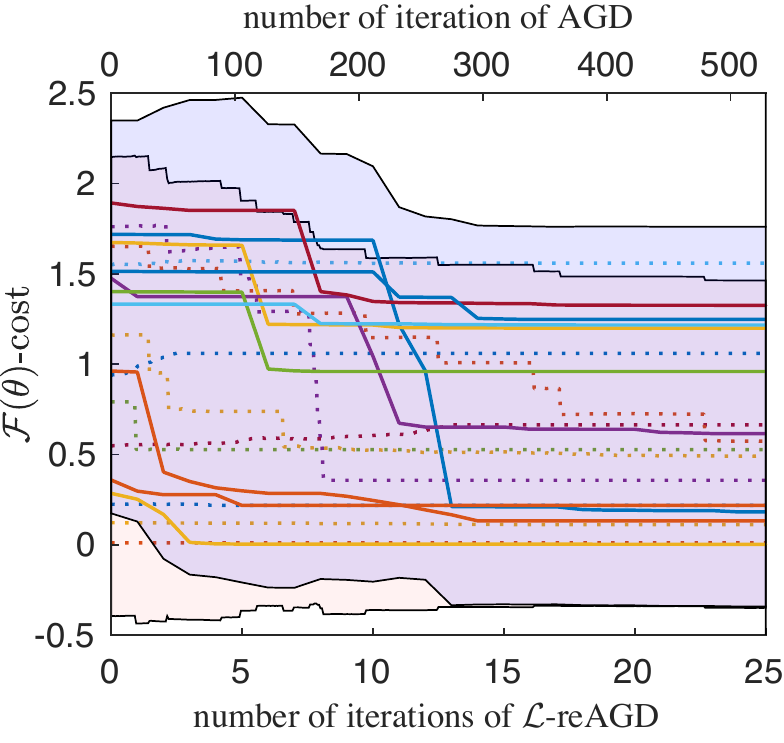}
			\caption{$\vartheta_1 = 4.000$, $\vartheta_2 = 0.500$}
		\end{subfigure}
		\hfill
		\begin{subfigure}[h]{0.19\textwidth}
			\centering
			\includegraphics[width=1\linewidth]{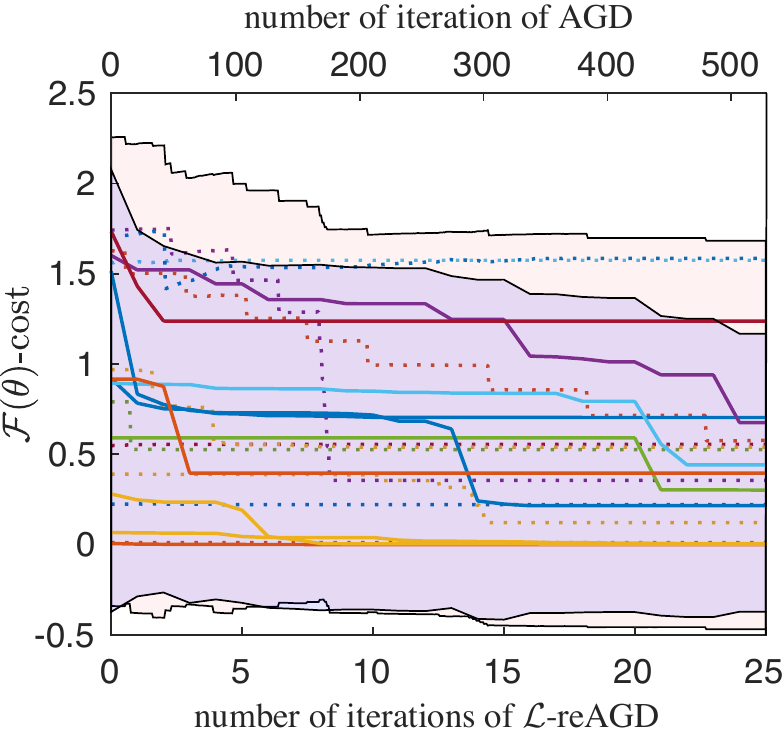}
			\caption{$\vartheta_1 = 5.657$, $\vartheta_2 = 0.625$}
		\end{subfigure}	
		\hfill	
	\end{centering}
	\caption{
	A figure shows how $\mathcal{L}$-reAGD and AGD perform under variant key parameters during the tuning experiments. The solid and dash lines denote the processes of $\mathcal{L}$-reAGD and AGD, while the purple and pink areas denote the 95\% confidence interval of their expectation, respectively. }
	\label{fig:tune_L-reAGD}
\end{figure*}

The results of tuning experiments (Figure~\ref{fig:tune_L-reAGD}) show that $\vartheta_1$ mainly determines the overall performance of $\mathcal{L}$-reAGD. The process tends to converge to $0$ when $\vartheta_1$ is nearby $2.000$ while drops into the local minimum otherwise. So Figure~\ref{fig:tune_L-reAGD} details how $\mathcal{L}$-reAGD performs differently when $\vartheta_1 \in [1.000, 5.567]$ and $\vartheta_2 \in [0.000, 0.875]$ especially $\vartheta_1 \in \{2.000, 2.282\}$ and shows the best performance when $\{\vartheta_1,\vartheta_2\} = \{2.000, 0.250\}$. Table~\ref{tab:AGD_tune} indicates that a larger $\vartheta_2$ determining the accelerated step of AGD can improve the convergence rate by 10\%-70\% and the success rate by 10\%-150\%. Moreover, each sub-figure of Figure~\ref{fig:tune_L-reAGD} illustrates 10 different $\mathcal{L}$-reAGDs denoted by the solid lines and 10 different AGDs denoted by the dash lines and uses the purple and pink area to show their 95\% confidence interval of expectation under a specific $\{\vartheta_1, \vartheta_2\}$. Moreover, Table~\ref{tab:AGD_tune} validates that the $\mathcal{L}$-restart significantly improves computational efficiency by 95\% and robustness by 30\%-80\% compared to AGD-1e2.   

\begin{table}[hbt]
\caption{Tuning $\vartheta_1$ and $\vartheta_2$ of $\mathcal{L}$-reAGD}
\label{tab:AGD_tune}
\centering
\scalebox{1}{
\begin{tabular}{c|p{20pt}p{22pt}p{22pt}p{22pt}p{22pt}p{22pt}}
\toprule
\scalebox{0.75}{\diagbox{$\vartheta_2$}{ (\%)\textbar(s)}{$\vartheta_1$}} & 1.000 & 1.414 & 2.000 & 2.828 & 4.000 & 5.657 \\ 
\midrule
0.000 & 50\textbar1.54 & 50\textbar2.03 & 40\textbar2.51 & 40\textbar2.65 & 30\textbar2.98 & 20\textbar3.13 \\
0.125 & — & 60\textbar2.16 & 70\textbar3.03 & \textbf{90}\textbar4.87 & 30\textbar2.44 & 20\textbar2.98\\
0.250 & — & 70\textbar1.89 & \textbf{100\textbar0.98} & 80\textbar3.27 & 40\textbar2.25 & 30\textbar2.78\\
0.375 & — & — & \textbf{90}\textbar1.49 & 80\textbar4.26 & 40\textbar1.92 & 20\textbar2.45 \\
0.500 & — & — & 80\textbar1.76 & 80\textbar2.03 & 30\textbar1.77 & 20\textbar2.89\\
0.625 & — & — & — & 80\textbar1.88 & 30\textbar1.89 & 20\textbar2.56\\
0.875 & — & — & — & 80\textbar1.67 & 40\textbar1.73 & 40\textbar2.43\\
\midrule
AGD-1e2 & 20\textbar46.2 & 20\textbar49.3 & 20\textbar50.3 & 20\textbar55.0 & 20\textbar55.0 & 20\textbar55.0 \\
\bottomrule
\end{tabular}}
\end{table} 

\subsubsection{ASTO parameter} \label{sec:ASTO_tune}

According to Section~\ref{sec:ASTO} and Algorithm~\ref{alg:ASTO}, the prior GP knowledge, the number of samples, and the Lipschitz constant $\mathcal{L}_\mathcal{R}$ of the rewards $\mathcal{R}$ highly affect the learning process of ASTO. Since the covariance matrix $\bm{\mathcal{K}}$, calculated by $\Delta t$~\eqref{eq:Phi-Qc01}, mainly determines the searching space, our experiment first tunes $\Delta t$ around $1$ and keeps $K | M^* = 12|6$ constant. Then our experiment tunes K\textbar M* around $12 | 6$ and keeps $\Delta t = 4$ constant to show how the number of samples affects the learning process. Moreover, the above experiments both tune $\mathcal{L}_\mathcal{Q}$ around $10$ and compare AMA, EMA, and qAdam, because AMA determined by $\mathcal{L}_\mathcal{R}$ refines the learning process informed by the reward $\mathcal{R}$ and the selected samples $\bm{\Theta}_{M^*}$. Moreover, we set $\{\alpha^\mu, \alpha^\kappa\}$ as $\{0.1, 0.1\} \text{ or } \{0.9, 0.9\}$ to compare AMA with EMA-0.1/EMA-0.9 and qAdam-0.1/qAdam-0.9 according to \eqref{eq:EMA} and \eqref{eq:bias}, respectively. All of the tuning tests run the whole iAGP-STO to solve 25 problems of class C, where $\mathcal{L}$-reAGD's parameters are set following Table~\ref{tab:paras_set}. 

Table~\ref{tab:ASTO_tune1} shows that a small $\bm{\mathcal{K}}$ with a small $\Delta t$ tightens the searching space around the prior mean value $\bm{\mu}$. It reduces the success rate '(\%)' by 55\% and increases the average computation time '(s)' by 200\% for the extra sampling and numerical process. On the contrary, a large $\bm{\mathcal{K}}$ with the over-released searching space also lowers the computational efficiency by 25\% and the algorithmic robustness by 10\%. Table~\ref{tab:ASTO_tune2} indicates that a more significant number of samplings improves the success rate of ASTO by 10\% while increasing the computational cost by  300\% caused by the functional estimation of the extra samples. They both validate our intuition (the end of Section~\ref{sec:M-step}) that a proper release of strict rewarding ascent can improve the learning ability. That is because a larger $\mathcal{L}_\mathcal{R}$ meaning more leakage of rewarding strengthens the learning ability of ASTO and improves the success rate by 50\% and computation time by 35\%. Moreover, they both show AMA's higher robustness (20\%) and efficiency (40\%) compared to EM and qAdam. 

\begin{table*}[htbp]

\begin{subtable}{\textwidth}
\caption{Tuning $\Delta t$ and $\mathcal{L}_\mathcal{R}$ of ASTO}
\label{tab:ASTO_tune1}
\centering
\scalebox{1}{
\begin{tabular}{c|ccccc|cccccccc}
\toprule
\scalebox{0.8}{\diagbox{$\Delta t$}{ (\%)\textbar(s)}{$\mathcal{L}_\mathcal{R}$}} & 100 & 10.0 & 4.00 & 2.00 & 1.25 & EMA-0.1 & EMA-0.9 & qAdam-0.1 & qAdam-0.9 \\ 
\midrule
0.25 & 36\textbar 8.389 & 40\textbar 7.557 & 44\textbar 6.313 & 44\textbar 5.988 & 36\textbar 6.274 
& 28\textbar 10.93 & 32\textbar 9.598 & 40\textbar 5.235 & 36\textbar 6.231 \\
1.00 & 52\textbar 4.124 & 72\textbar 3.382 & 68\textbar 3.434 & 68\textbar 3.513 & 48\textbar 4.012 
& 56\textbar 4.817 & 60\textbar 3.981 & 64\textbar 4.845 & 60\textbar 4.234\\
2.00 & 64\textbar 3.912 & 80\textbar 2.899 & 84\textbar 2.982 & 84\textbar 3.123 & 64\textbar 3.233 
& 68\textbar 3.235 & 80\textbar 2.678 & 76\textbar 4.456 & 72\textbar 4.617 \\
4.00 & 76\textbar 3.329 & \textbf{92\textbar 2.442} & 88\textbar 2.763 & 76\textbar 3.892 & 60\textbar 3.723 
& 76\textbar 3.238 & 76\textbar 3.557 & 84\textbar 4.744 & 76\textbar 3.554 \\
8.00 & 68\textbar 3.981 & 84\textbar 2.983 & 80\textbar 3.201 & 76\textbar 3.456 & 76\textbar 3.211 
& 64\textbar 4.099 & 68\textbar 3.778 & 80\textbar 3.283 & 56\textbar 3.202 \\
\bottomrule
\end{tabular}}
\end{subtable}

\par\bigskip

\begin{subtable}{\textwidth}
\caption{Tuning K\textbar M* and $\mathcal{L}_\mathcal{R}$ of ASTO}
\label{tab:ASTO_tune2}
\centering
\scalebox{1}{
\begin{tabular}{c|ccccc|cccccccc}
\toprule
\scalebox{0.8}{\diagbox{K\textbar M*}{ (\%)\textbar(s)}{$\mathcal{L}_\mathcal{R}$}} & 100 & 10.0 & 4.00 & 2.00 & 1.25 & EMA-0.1 & EMA-0.9 & qAdam-0.1 & qAdam-0.9 \\ 
\midrule
8\textbar4 & 76\textbar 2.991 & 88\textbar 1.722 & 84\textbar \textbf{1.709} & 64\textbar 2.923 & 54\textbar 2.679 
& 72\textbar 2.255 & 52\textbar 2.371 & 88\textbar 2.760 & 48\textbar 2.364 \\
12\textbar6 & 76\textbar 3.329 & 92\textbar 2.442 & 88\textbar 2.763 & 76\textbar 3.892 & 60\textbar 3.723 
& 76\textbar 3.072 & 76\textbar 3.557 & 84\textbar 4.774 & 76\textbar 3.554 \\
18\textbar9 & 76\textbar 6.317 & 92\textbar 3.553 & 84\textbar 4.083 & 68\textbar 7.328 & 52\textbar 5.384 
& 76\textbar 4.678 & 68\textbar 5.078 & 92\textbar 3.678 & 60\textbar 4.289 \\
36\textbar18 & 84\textbar13.92 & \textbf{96}\textbar 7.125 & 92\textbar 8.032 & 80\textbar 11.40 & 60\textbar 10.25
& 80\textbar 10.40 & 84\textbar 9.323 & 92\textbar 5.981 & 72\textbar 6.889 \\
\bottomrule
\end{tabular}}
\end{subtable}

\end{table*} 
%

\subsection{Parameter setting}\label{sec:paras_set}

Besides the above settings, we set $\epsilon = 0.05, \epsilon_d = \text{1e-2}$ and other parameters according to Table~\ref{tab:paras_set}. 

%
\begin{table}[htbp]
\caption{Parameter Setting}
\label{tab:paras_set}
\centering
\scalebox{1}{
\begin{tabular}{ccccccccccc}
\toprule
algo. & para. & meaning & value  \\ 
\midrule
\multirow{3}*{iOMP} & $\Delta t$ & initial time interval & 4.0 \\
& $\mathbf{Q}_c$ & power-spectral density matrix & $\mathbf{I}$ \\
& $\tau_\textit{ip}$ & interpolation scaling factor & 2 \\ 
\midrule
\multirow{5}*{AGP-STO} & $\bm{\varrho}$ & initial Lagrangian factor & $\textbf{0.01}$ \\
& $\mathcal{F}$tol & convergence of functional cost & 1e-5 \\
& $\theta$tol & convergence of gradient norm & 1e-4 \\
& $\mathcal{G}$tol & collision cost tolerance & 1e-4 \\
& $\kappa$ & penalty scaling factor & 0.4 \\
\midrule
\multirow{6}*{$\mathcal{L}$-reAGD} & $c_\textit{low}$ & factor of lower predicted reduction & 1.25 \\
& $c_\textit{up}$ & factor of upper predicted reduction & 0.15 \\
& $\vartheta_1$ & scaling factor of $\beta_k$ & 2 \\
& $\vartheta_2$ & scaling factor of $\lambda_k$ & 0.25 \\
& $\upphi\text{tol}$ & slope tolerance of $\phi$ & -0.1 \\
& $\mathcal{L}\text{tol}$ & expanding tolerance of $\mathcal{L}_\mathcal{F}$ & 1e2 \\
\midrule
\multirow{7}*{ASTO} & K\textbar M* & number of overall\textbar selected samples & 12\textbar 6 \\
& $\mathcal{L}_\mathcal{R}$ & Lipschitz constant of reward $\mathcal{R}$ & 10 \\
& $c\mathcal{F}$tol & improvement rate tolerance & 0\% \\
& $\bar{\mathcal{K}}$tol & upper tolerance of $\bm{\mathcal{K}}$-norm & \scalebox{0.7}{$\frac{\|\bm{\theta}_\text{max} - \bm{\theta}_\text{min}\|^2}{1.25}$} \\
& $\underline{\mathcal{K}}$tol & lower tolerance of $\bm{\mathcal{K}}$-norm & 1e-2 \\
& $N_\textit{Asto}$ & maximum number of \textit{AstoIter} & $\mathcal{U}(5,15)$ \\
\bottomrule
\end{tabular}}
\end{table} 
%

\subsection{Results analysis}\label{sec:analysis}

\begin{figure*}[hbtp]
	\begin{centering}
		\begin{subfigure}[b]{0.16\textwidth}
			\centering
			\includegraphics[width=1\linewidth]{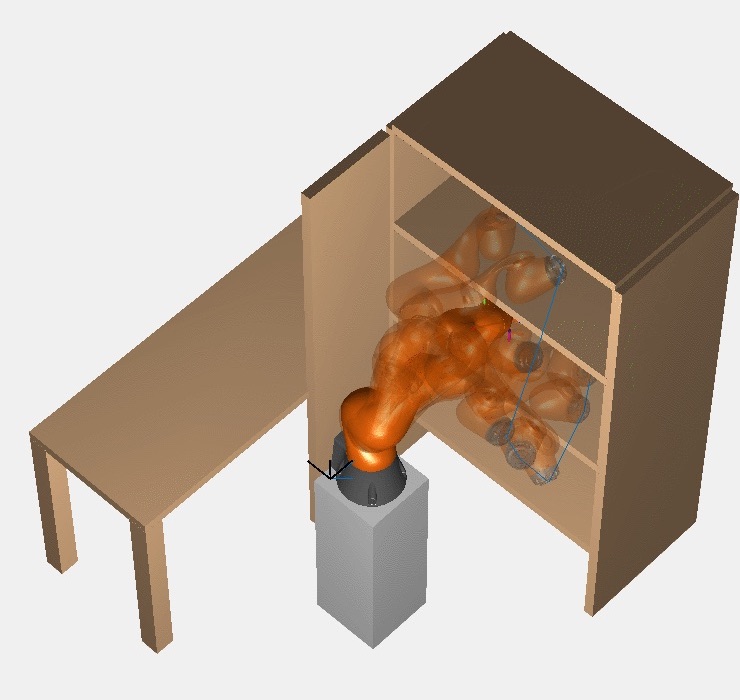}
			\caption{Trajectory B-1.2}
		\end{subfigure}	
		\begin{subfigure}[b]{0.16\textwidth}
			\centering
			\includegraphics[width=1\linewidth]{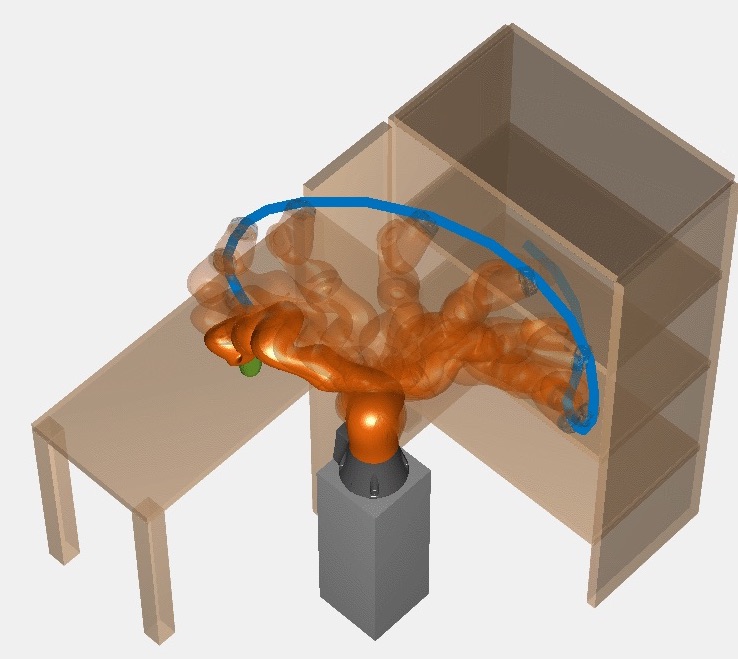}
			\caption{Trajectory B-2.1}
		\end{subfigure}	
		\begin{subfigure}[b]{0.16\textwidth}
			\centering
			\includegraphics[width=1\linewidth]{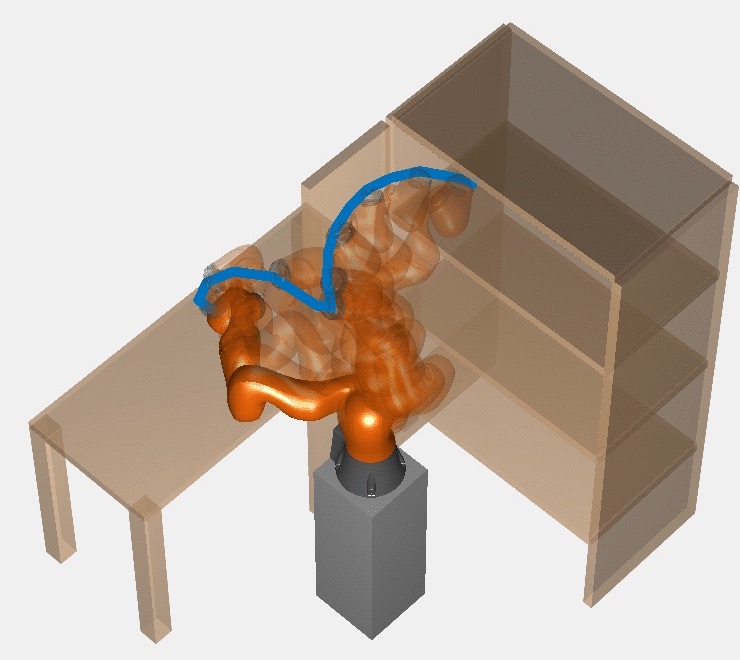}
			\caption{Trajectory B-3.2}
		\end{subfigure}	
		\begin{subfigure}[b]{0.16\textwidth}
			\centering
			\includegraphics[width=1\linewidth]{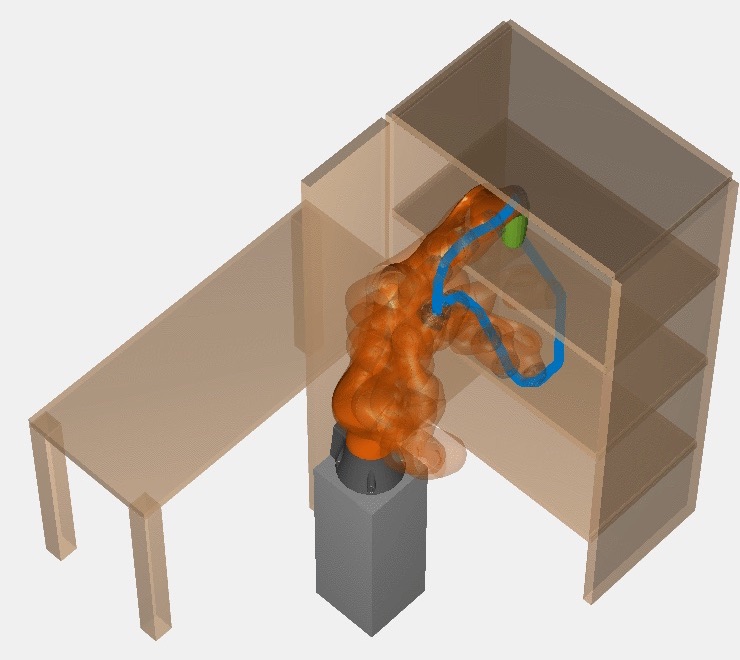}
			\caption{Trajectory B-4.2}
		\end{subfigure}	
		\begin{subfigure}[b]{0.16\textwidth}
			\centering
			\includegraphics[width=1\linewidth]{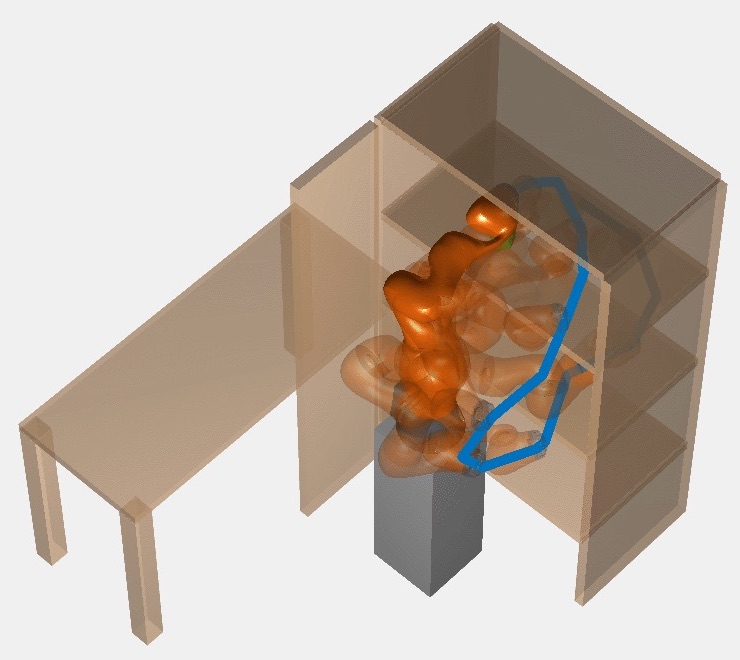}
			\caption{Trajectory C-1.4}
		\end{subfigure}	
		\begin{subfigure}[b]{0.16\textwidth}
			\centering
			\includegraphics[width=1\linewidth]{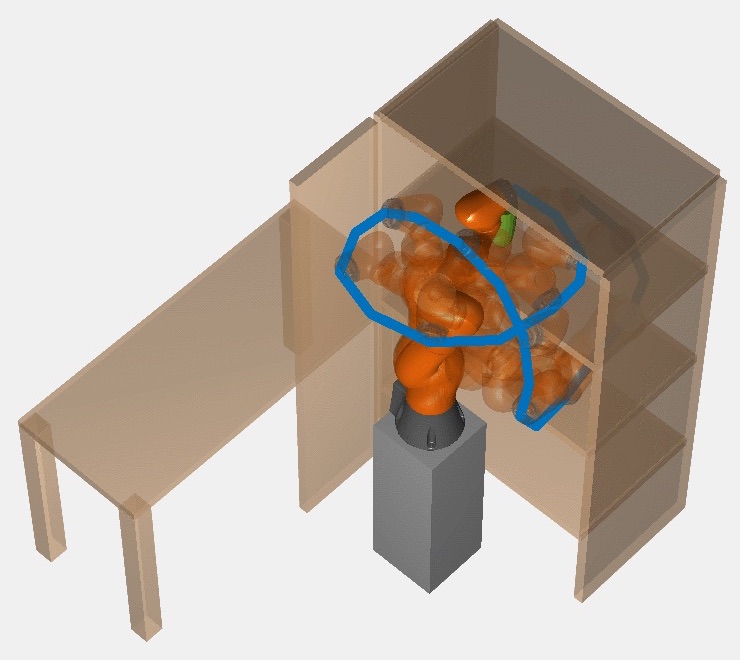}
			\caption{Trajectory C-1.5}
		\end{subfigure}
		\begin{subfigure}[b]{0.16\textwidth}
			\centering
			\includegraphics[width=1\linewidth]{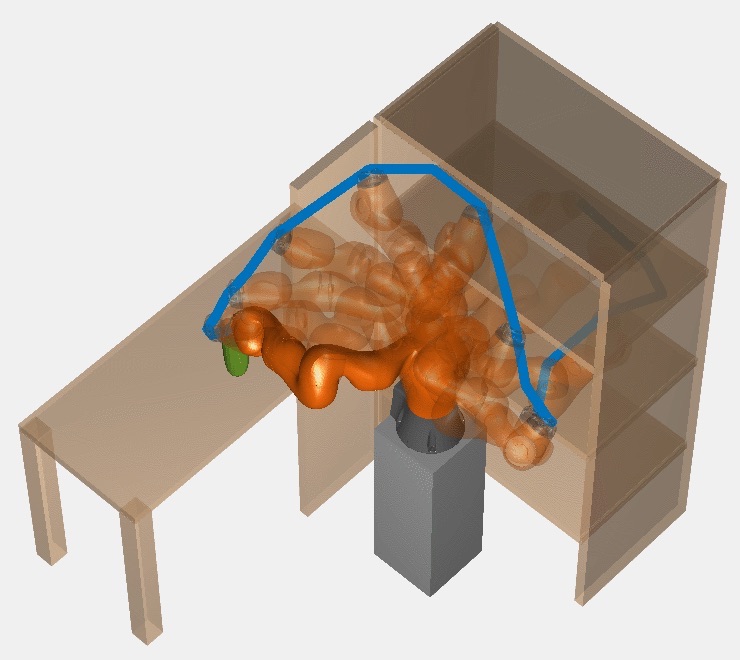}
			\caption{Trajectory C-2.4}
		\end{subfigure}	
		\begin{subfigure}[b]{0.16\textwidth}
			\centering
			\includegraphics[width=1\linewidth]{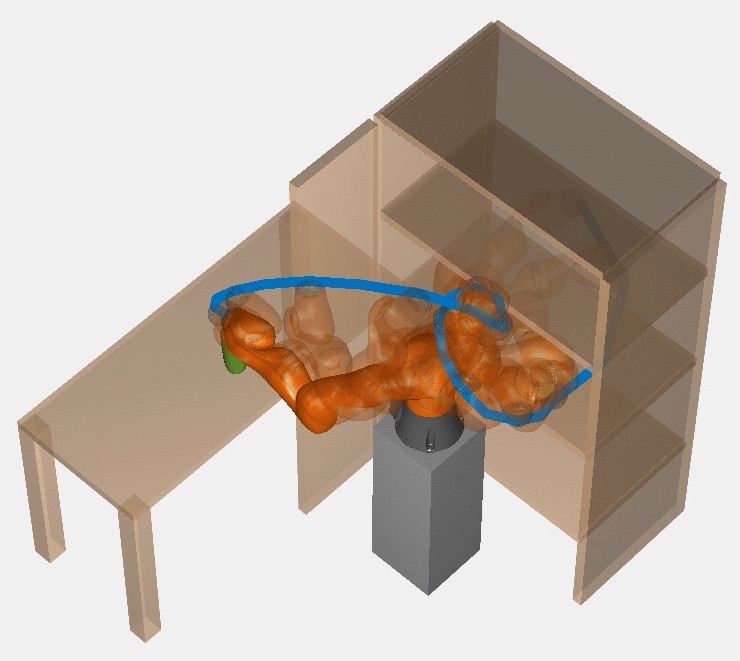}
			\caption{Trajectory C-2.8}
		\end{subfigure}
		\begin{subfigure}[b]{0.16\textwidth}
			\centering
			\includegraphics[width=1\linewidth]{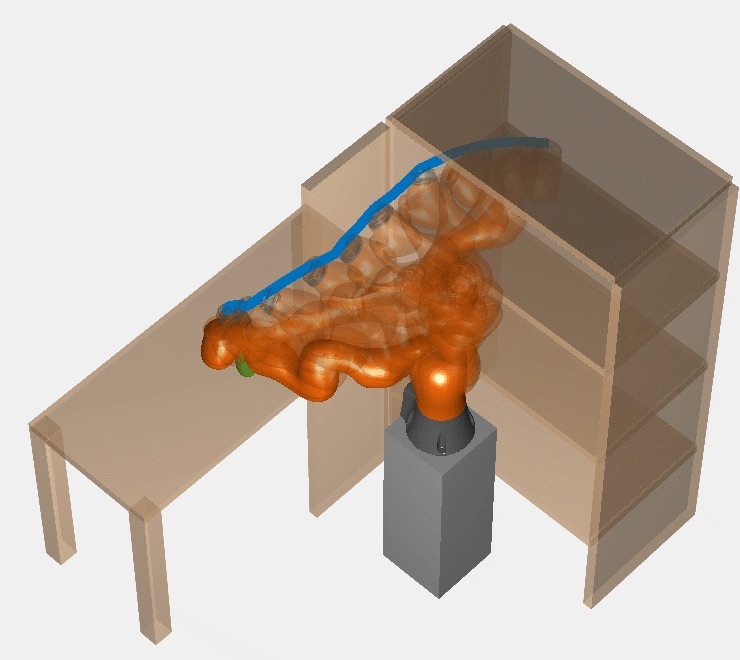}
			\caption{Trajectory C-3.1}
		\end{subfigure}
		\begin{subfigure}[b]{0.16\textwidth}
			\centering
			\includegraphics[width=1\linewidth]{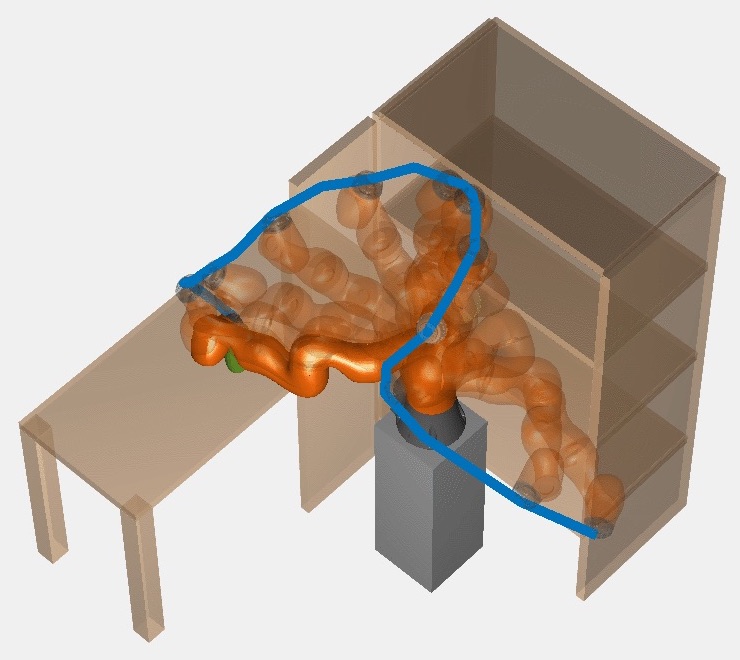}
			\caption{Trajectory C-4.2}
		\end{subfigure}	
		\begin{subfigure}[b]{0.16\textwidth}
			\centering
			\includegraphics[width=1\linewidth]{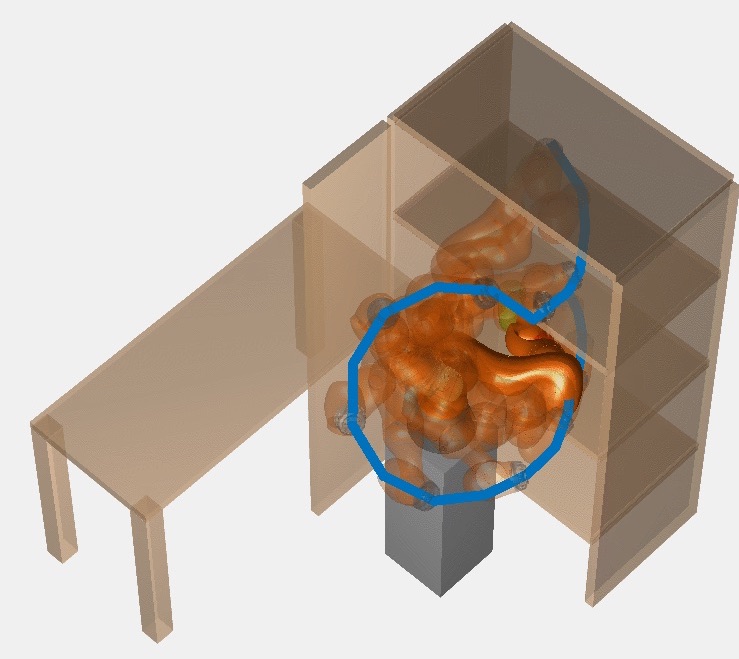}
			\caption{Trajectory C-5.2}
		\end{subfigure}
		\begin{subfigure}[b]{0.16\textwidth}
			\centering
			\includegraphics[width=1\linewidth]{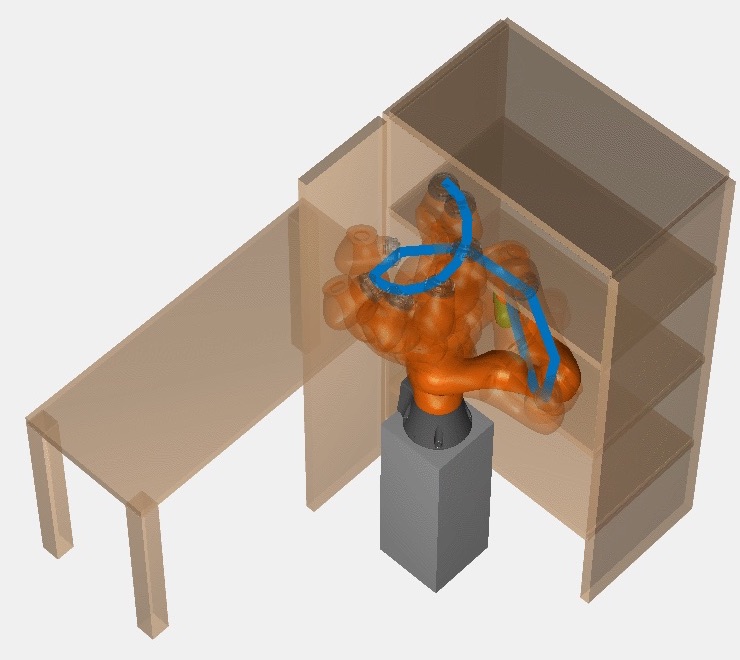}
			\caption{Trajectory C-6.1}
		\end{subfigure}
		
	\end{centering}
	\caption{A figure shows the randomly selected results of iAGP-STO in class B and C, where the randomly selected goal states satisfy the same goal requirement in one task. Note that \textit{X3} in \textit{Trajectory X1-X2.X3} means a single problem \textit{X3} in task \textit{X2} categorized into class \textit{X1}. }
	\label{fig:results_BC}
\end{figure*}

This section shows the benchmark results of all OMP problems (class A, B, C) specified by Section~\ref{sec:setup} and Table~\ref{tab:probClassify} in Table~\ref{table:results_A}-\ref{table:results_C}. Besides, Figure~\ref{fig:results_BC} randomly draws the planning results of the problems in class B and C. 


Table~\ref{table:results_A} shows that iAGP-STO could gain the optimal trajectory in the second- lowest computational time and highest success rate. Because the iAGP-STO only utilizes the gradient information without the Hessian calculation adopted by GPMP and makes 5\% improvement. Besides, the fair judging of \textit{MinAprch} successfully integrates ASTO to place the trajectory in the feasible space with a few resampling processes. It makes 60\%, 65\%, and 80\% improvement of efficiency compared to CHOMP-HMC, STOMP, and RRT-Connect. That is because STOMP needs plenty of time for the iterative processes containing resampling and simulation, though there is no need for gradient calculation. The vast searching space of RRT-connect easily neglects the subspace near the initial trajectory. TrajOpt shows 70\% better convergence rate while 10\% lower success rate for the convex constrained problems in the trust-region method. 
As for CHOMP-HMC, though HMC costs 10\%-60\% less time than STOMP and RRT-connect, the iterative alternation between the Hamiltonian simulation and the Monte Carlo sampling take up extra computation resource. Besides, CHOMP (without HMC) saves 50\% more computation resources while reducing the success rate by 10\% than CHOMP-HMC and validates $\mathcal{L}$-reAGD's priority over the leapfrog method in success rate by 5\%.  


Table~\ref{table:results_C} shows that iAGP-STO gains the highest success rate compared to the other planners' thanks to our numerical and sampling methods integration. Because ASTO performs the resampling process iteratively rather than the iterative alternation between numerical optimization and single resampling process, iAGP-STO can find a feasible subspace more reliably (about 200\%) and save 65\% computation time in comparison to CHOMP. Moreover, the AMA method of ASTO and appropriate selection of $\bm{\mathcal{K}}$ elevate the success rate by 250\% and save 65\% computation resources compared to STOMP. 
Although RRT-connect adopts the bidirectional searching method and is provably probabilistically complete, it occupies 50\% more computation resources to store the searching tree and check collision facing the narrow feasible space. Though HMC samples the momentum and finds the Hamiltonian minimum numerically, CHOMP performs badly when nonconvex infeasible sets surround the initial trajectory. As for GPMP and TrajOpt, the LM and trust-region method help them approach the local minima rapidly. However, whether the final optimized trajectory is feasible depends on whether the initial point is in the convex neighbor of the global minimum. 

Table~\ref{table:results_A}~\&~\ref{table:results_C} both show that iOMP utilized by iAGP-STO improves the computation efficiency by 25\%-50\% compared to AGP-STO, which benefited from the incremental method. Moreover, the integration of ASTO and $\mathcal{L}$-reAGD elevates the success rate by 600\% and only takes  50\% more computation resources for trajectory resampling and numerical refinement than $\mathcal{L}$-reAGD.

\begin{table*}[hbtp]

\begin{subtable}{\textwidth}
\caption{Results of 10 class \textbf{A} and 15 class \textbf{B} planning problems (with 5 repeated trials for each) on LBR-iiwa with  7-DoF.}
\label{table:results_A}
\centering
\scalebox{1}{
\begin{tabular}{lcccccccccc}
\toprule
& \textbf{$\bm{\mathcal{L}}$-reAGD} & \textbf{AGP-STO} & \textbf{iAGP-STO} & {TrajOpt-78} & {GPMP-16} & CHOMP &  {CHOMP-HMC} & {STOMP}  & {RRT-Connect} \\ 
\midrule
Success (\%)  & 84.8 & \textbf{100} & \textbf{100} & 88 & 95.2 & 81.6 & 93.6 & {96.8} & \textbf{100} \\
Avg. Time (s) & 1.318 & 3.047 & 1.718 & \textbf{0.513} & 1.771 & 2.219 & 4.503 & 5.101 & 11.00 \\
Std-Dev. Time (s) & 0.735 & 1.048 &  0.139 & 0.050 & 0.325 & 0.412 & 0.899 & 3.233 & 4.728 \\
\bottomrule
\end{tabular}}
\end{subtable}

\par\bigskip

\begin{subtable}{\textwidth}
\caption{Results of 25 class \textbf{C} planning problems (with 5 repeated trials for each) on LBR-iiwa with 7-DoF. }
\label{table:results_C}
\centering
\scalebox{1}{
\begin{tabular}{lccccccccc}
\toprule
& \bf{$\bm{\mathcal{L}}$-reAGD} & \bf{AGP-STO} & \bf{iAGP-STO} & {TrajOpt-78} & {GPMP2-16} & CHOMP & {CHOMP-HMC} & {STOMP} & {RRT-Connect} \\ 
\midrule
Suc. Rate (\%)  & 12.8 & 87.2 & \bf{91.2} &  8.8 & 11.2 & 8.0 & 30.4 & {24.8} & 72.8 \\
Avg. Time (s) & 1.622 & 4.284 & 2.442 & \textbf{1.163}  & 2.587 & 3.584 & 7.238 & 8.176 & 17.08 \\
Std-Dev. Time (s) & 0.735 & 1.048 &  0.419 & 0.104 & 1.345 & 0.577 & 1.322 & 5.364 & 2.726 \\
\bottomrule
\end{tabular}}
\end{subtable}

\end{table*}
%


\subsection{Multiple AGV collaboration} \label{sec:AGVs}

\begin{figure*}[hbtp]
	\begin{subfigure}[b]{0.12\textwidth}
		\centering
		\includegraphics[width=1\linewidth]{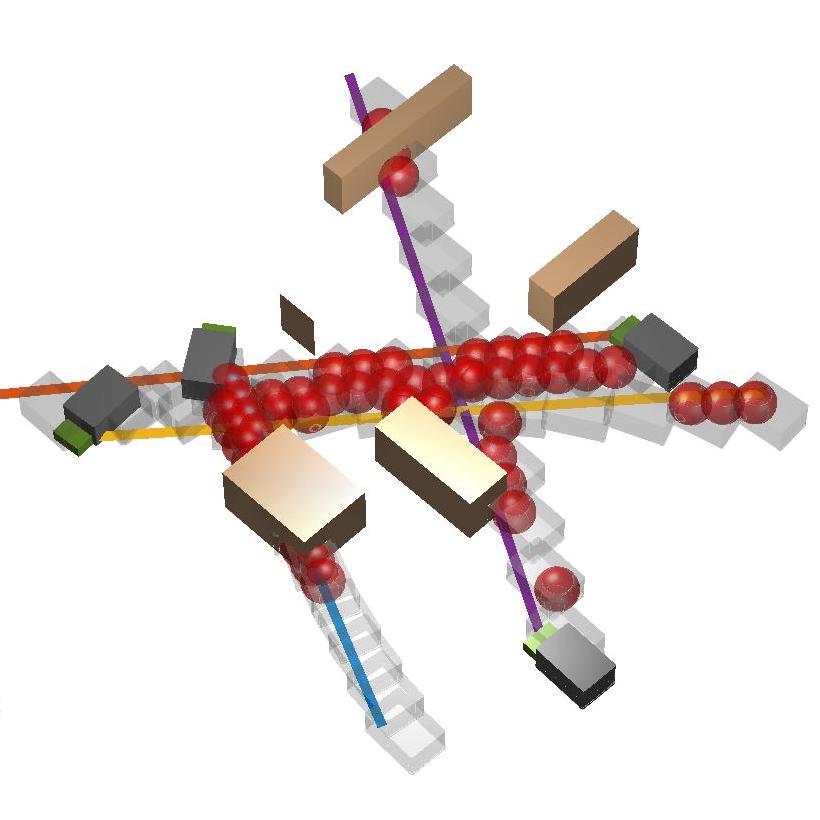}
		\caption{AGV-1, 11.4\textbar 0.47}
		\label{fig:AGV-1}
	\end{subfigure}	
	\hfill
	\begin{subfigure}[b]{0.86\textwidth}
		\centering
		\includegraphics[width=1\linewidth]{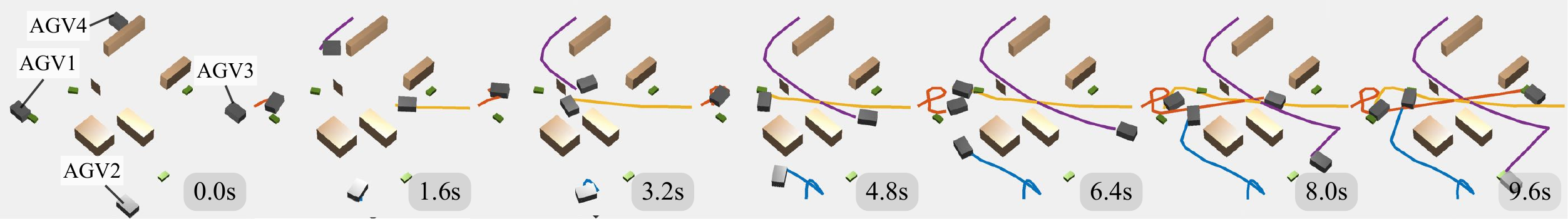}
		\caption{Trajectory AGV-1.1}
		\label{fig:AGVs-traj-1.1}
	\end{subfigure}	
	\\ 
	\begin{subfigure}[b]{0.12\textwidth}
		\centering
		\includegraphics[width=1\linewidth]{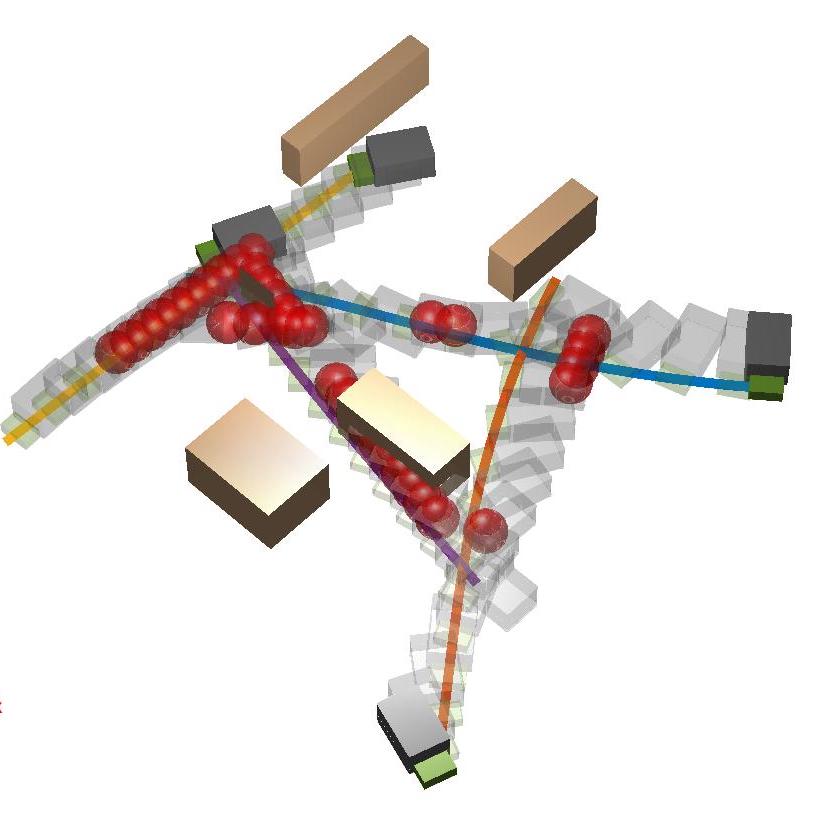}
		\caption{AGV-2, 3.71\textbar 0.59}
		\label{fig:AGV-2}
	\end{subfigure}	
	\hfill
	\begin{subfigure}[b]{0.86\textwidth}
		\centering
		\includegraphics[width=1\linewidth]{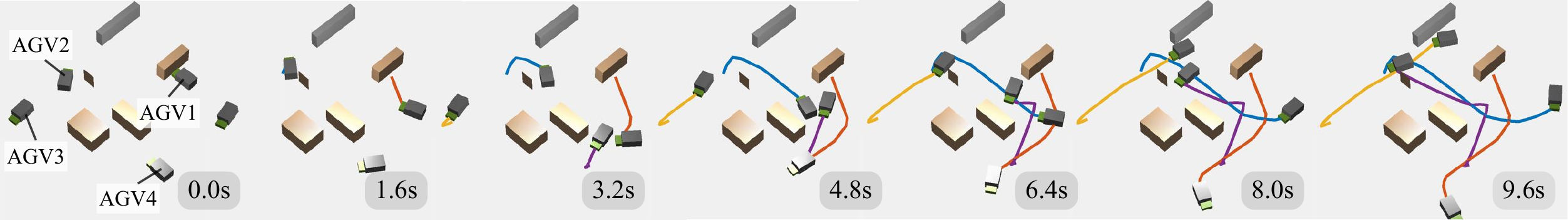}
		\caption{Trajectory AGV-2.1}
		\label{fig:AGVs-traj-2.1}
	\end{subfigure}	
	\caption{A figure shows the results of multi-AGV allocation planned by iAGP-STO. }
\end{figure*}

This part will illustrate how a multi-AGV robotic system with 12 DoFs reallocates 4 AGVs to safely grab and load 4 boxes in a warehouse containing obstacles. The red area of initial trajectories in Figures~\ref{fig:AGV-1}-\ref{fig:AGV-2} shows the interference between multiple AGVs and obstacles when AGVs transfer the boxes to different locations. In this way, our implementation sets $\epsilon = 0.25$ for collision-check considering the safety of hybrid motion, initial penalty factor $\varrho = 0.03125$, and the others according to Table~\ref{tab:paras_set}. To further test iAGP-STO's solvability, we conduct 5 repeated trials for each task given an initial state and goal constraint. 

Figure~\ref{fig:AGVs-traj-1.1} shows that 4 AGVs grab the boxes located at the safe places of the warehouse from the default state, bypassing the crossroad surrounded by some obstacles. While Figure~\ref{fig:AGVs-traj-2.1} shows they load the boxes to other randomly generated locations, avoiding self-collision. The 100\% success rate with 4.537s average time shows the high reliability and efficiency of iAGP-STO. 

%

\subsection{rethink-Baxter assistance} \label{sec:baxter}

\begin{figure}[hbtp]
	\begin{centering}
		\begin{subfigure}[b]{0.23\textwidth}
			\centering
			\includegraphics[width=1\linewidth]{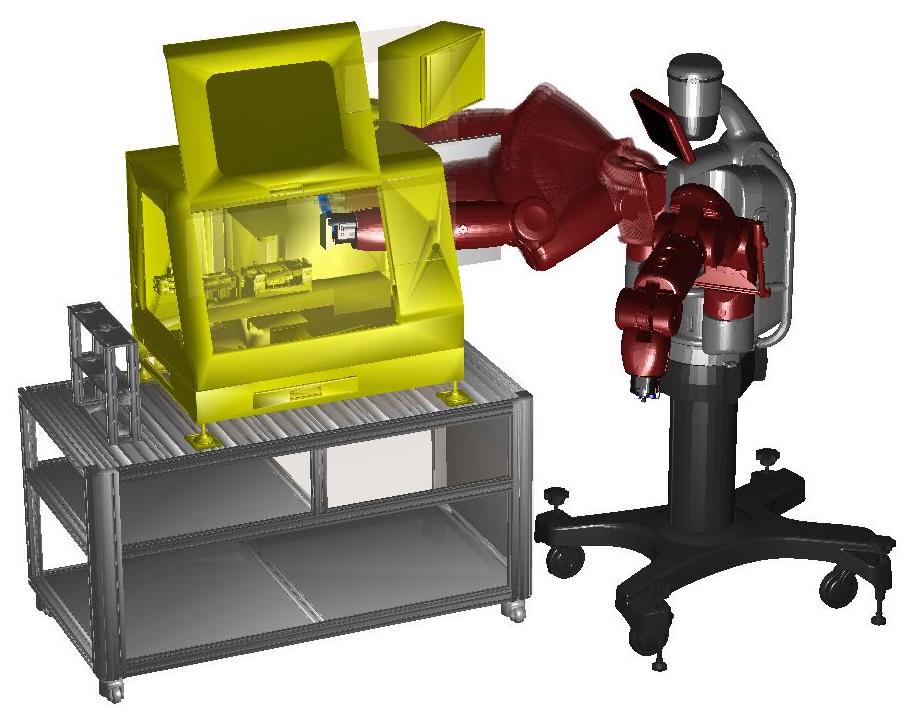}
			\caption{Task Baxter-1, $0.36 | 0.88$}
		\end{subfigure}	
		\hfill
		\begin{subfigure}[b]{0.23\textwidth}
			\centering
			\includegraphics[width=1\linewidth]{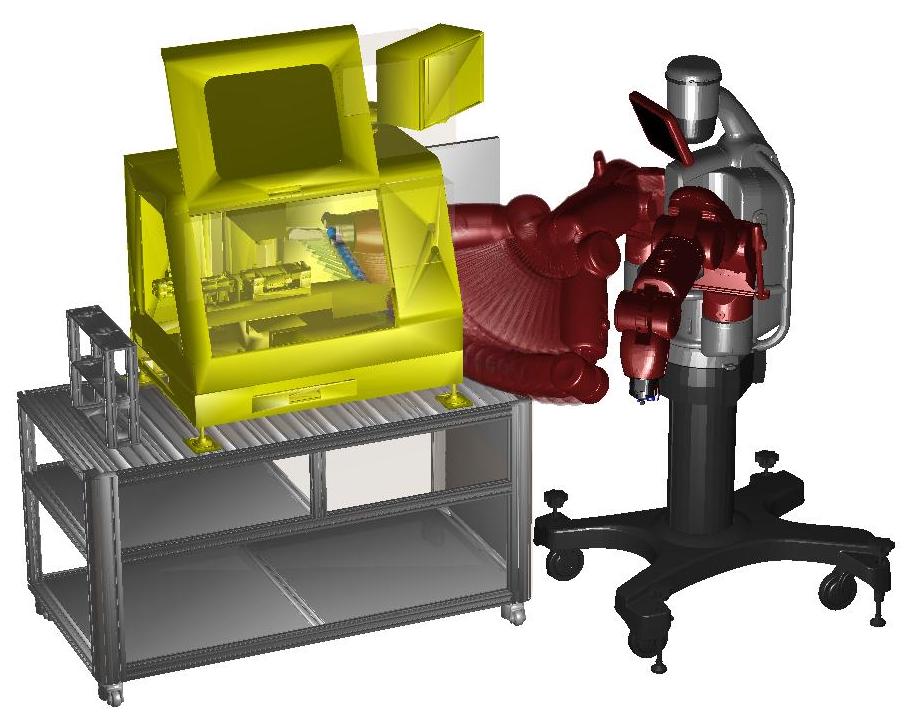}
			\caption{Task Baxter-2, $0.53 | 0.94$}
		\end{subfigure}	
	\end{centering}
	\caption{Planning problems of rethink-Baxter assistance. }
	\label{fig:baxter}
\end{figure}

This part will illustrate how a rethink-Baxter robot with 7 DoFs executes a collision-free motion to grab and load an aluminum workpiece in a computer numerical control (CNC) machine with a highly compact structure. The red area on initial trajectories of the tasks in Figure~\ref{fig:baxter} shows the interference between the arm and the shelf when AUBO-i5 is transferring the workpiece between different cells. In this way, our implementation sets $\epsilon = 0.025$ for collision-check considering the narrow space of the CNC machine, $\Delta t  = 2$ because of the narrower feasible motion region, and the others according to Table~\ref{tab:paras_set}. To further test iAGP-STO's solvability, we conduct 5 repeated trials for each task. 

\begin{figure*}[hbtp]
		\hspace*{\fill}
		\begin{subfigure}[b]{0.49\textwidth}
			\centering
			\includegraphics[width=1\linewidth]{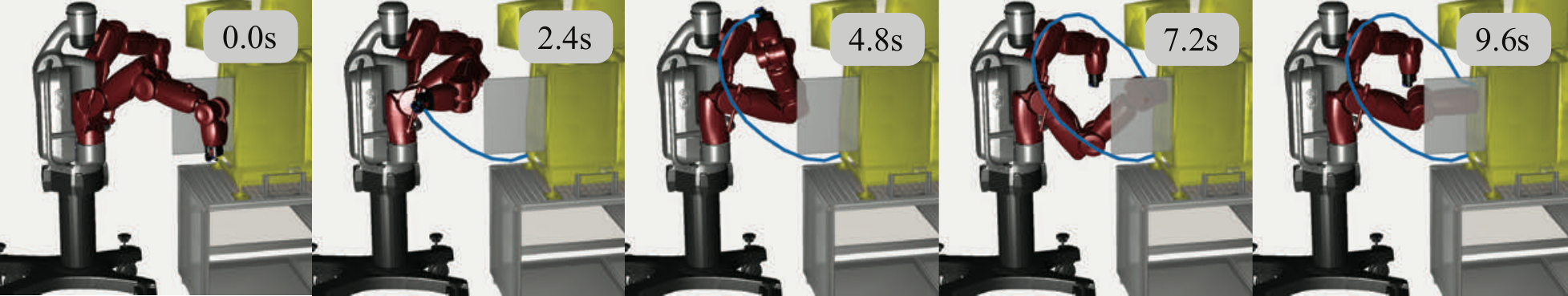}
			\caption{Simulation trajectory of Baxter-1}
			\label{fig:baxter-traj-1.1-s}
		\end{subfigure}	
		\hspace*{\fill}
		\begin{subfigure}[b]{0.49\textwidth}
			\centering
			\includegraphics[width=1\linewidth]{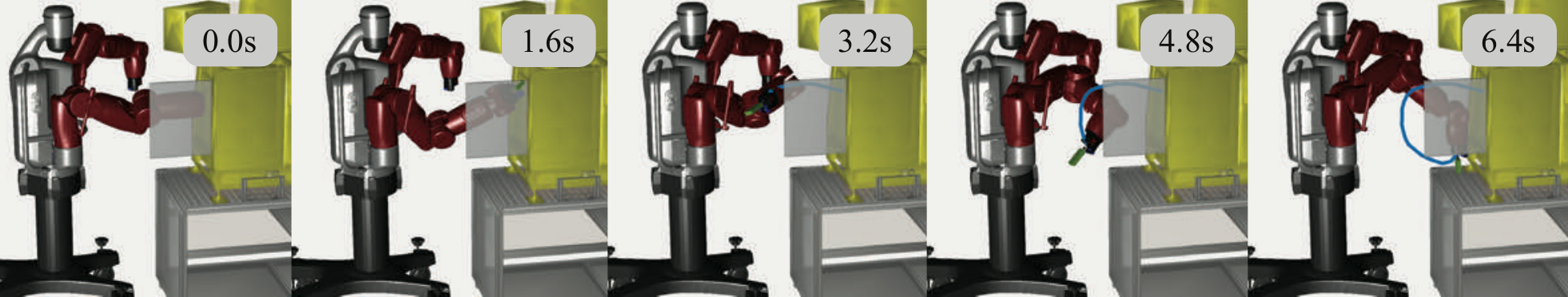}
			\caption{Simulation trajectory of Baxter-2}
			\label{fig:baxter-traj-2.1-s}
		\end{subfigure}	
		\hspace*{\fill} 
		\begin{subfigure}[b]{0.49\textwidth}
			\centering
			\includegraphics[width=1\linewidth]{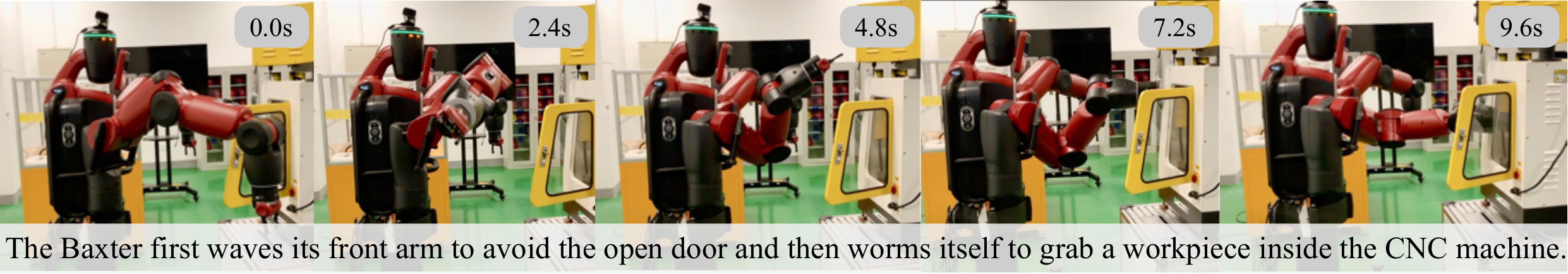}
			\caption{Real trajectory of Baxter-1}
			\label{fig:baxter-traj-1.1-r}
		\end{subfigure}	
		\hspace*{\fill}
		\begin{subfigure}[b]{0.49\textwidth}
			\centering
			\includegraphics[width=1\linewidth]{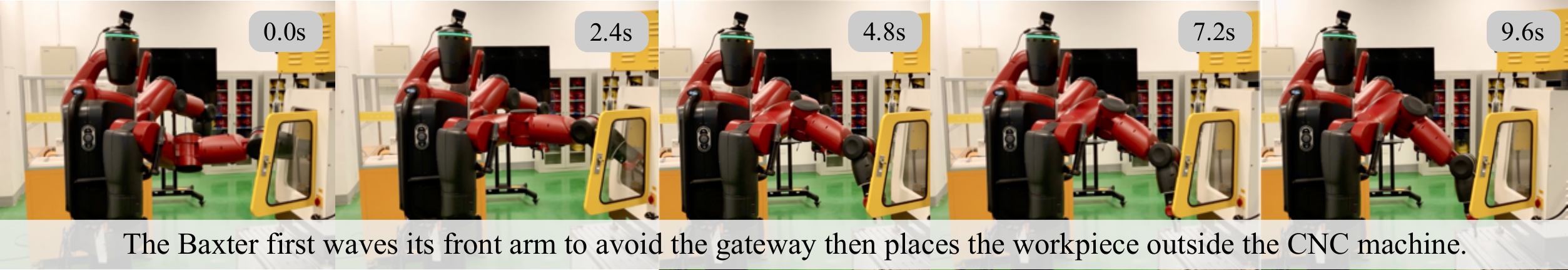}
			\caption{Real trajectory of Baxter-2}
			\label{fig:baxter-traj-2.1-r}
		\end{subfigure}
	\caption{A figure shows the results of rethink-Baxter assistance planned by iAGP-STO. }
\end{figure*}

Figure~\ref{fig:baxter-traj-1.1-s} shows that Baxter first waves its front arm to avoid the open door and then worms itself to grab a workpiece inside the CNC machine from the default state. While Figure~\ref{fig:baxter-traj-2.1-s} shows that it loads the workpiece from one station to another outside the CNC machine, bypassing the opening door. The 90\% success rate and 2.072s average time show the high reliability and efficiency of iAGP-STO. Moreover, we experiment in a real industrial environment to validate the above simulation in Figure~\ref{fig:baxter}, and their results are shown in Figure~\ref{fig:baxter-traj-1.1-r}-\ref{fig:baxter-traj-2.1-r}. 

%

\section{Discussion}\label{sec:discussion}

AGP-STO adopts GPMP to formulate the OMP problem and integrates the numerical and sampling method for highly reliable and efficient planning in the hazardous W-space. 

GPMP mainly concentrates on generalizing the collision-free OMP problem based on CHOMP and TrajOpt. However, AGP-STO directly adopts the GPMP's formulation and focuses on approaching the global optima more reliably and efficiently. To infer a trajectory by MAP given the information of GP prior and obstacle, we adopt the penalty method like TrajOpt and propose the $\mathcal{L}$-restart method based on AGD. Its numerical MAP only requires the first order momenta to reduce memory resources consumed for the LM-Hessian of GPMP. Moreover, it adopts AGD to generalize the leapfrog method adopted by CHOMP and reestimates the $\mathcal{L}$ipschitz constant to elevate the convergence rate. 

STOMP proposes a stochastic trajectory optimization to save the computation resource for gradient (i.e., momentum) estimation and overcome the local minima of some nonconvex cases. However, its small covariance ensuring the safe descent restrains the search space. Our method proposes ASTO on STOMP to gather the environmental information of the objective functional. To enlarge the searching space, ASTO samples trajectory obeying the Gaussian distribution (with a significant covariance) even the uniform distribution. Moreover, it adopts EM for reward and proposes the AMA method to reward the policy learning of the GP model. 

To find the Hamiltonian minimum, CHOMP adopts HMC, which alternates between the leapfrog and Monte-Carlo methods. However, AGP-STO first proposes \code{MinAprch} to judge the local minima approaching when reestimating the $\mathcal{L}$ipschitz constant. It then learns the GP-policy via ASTO when the minimum approaches and performs a numerical MAP of the learned policy. Moreover, we upgrade AGP-STO to iAGP- STO, which selects the guidance waypoints and optimizes the sub-trajectory incrementally to improve the success rate and computation efficiency, unlike iGPMP upgrade, directly applying iSAM2 to GPMP.

\section{Conclusion}\label{sec:conclusion}

iAGP-STO integrates a numerical and sampling method for optimal motion planning to overcome the local minima and converge to global minimum rapidly. 
\begin{enumerate}[leftmargin = 0pt, itemindent = 2\parindent, label=(\roman*)]
	\item The $\mathcal{L}$-reAGD method reestimates the $\mathcal{L}$ipschitz constant to converge 95\% faster in the semi-convex subspace.
	\item The introduction of \textit{MinAprch} judgment and ASTO helps search for the convex neighborhood of global minimum. It improves the success rate by 200\% compared to the numerical methods and reduces the computation time by 65\%-85\% compared to the sampling method. 
	\item  Furthermore, we use the incremental method based on Bayesian inference to optimize the whole trajectory via solving the noisy subproblems, which could expedite the whole trajectory optimization process by 25\%- 50\% and improve the success rate by 5\% with searching space reduction. 
\end{enumerate}


%

\appendices
\section{ Parameter derivation of AGD}\label{appdx:paraAGD}

According to Corollary~\ref{corol:AGD1} and Lemma~\ref{lemma:Gamma}, we can get
\begin{equation}\label{eq:GammaSum1}
\small
	\Gamma_{k} = \frac{2}{k(k+1)}, \quad \
	\sum_{\tau = k}^{N_{ag}}\Gamma_{\tau} = 2\left(\frac{1}{k}-\frac{1}{N_{ag}}\right) . 
\end{equation}
In this way, \eqref{eq:conditionAGparas} can be scaled as
\begin{equation}\label{ieq:Ck1}
\small
\begin{aligned}
	C_{k} &= 1 - \mathcal{L}_{\mathcal{F}}\lambda_{k} - \frac{\mathcal{L}_{\mathcal{F}}\left(\lambda_{k}-\beta_{k}\right)^{2}}{2\alpha_{k}\Gamma_{k}\lambda_{k}} \sum_{k=1}^{N_{ag}}\Gamma_{\tau} \\
	&\geq 1 - \mathcal{L}_{\mathcal{F}}{\left(1+\vartheta_{2}\alpha_{k}\right)}\beta_{k} - \frac{\mathcal{L}_{\mathcal{F}}\left(\vartheta_{2}\alpha_{k}\beta_{k}\right)^2}{2\alpha_{k}\Gamma_{k}\beta_{k}}\cdot\frac{2}{k} \\
	&= 1 - \frac{1+\vartheta_{2}\alpha_{k}}{\vartheta_{1}} - \frac{\alpha_{k} \vartheta_{2}^{2}}{k\Gamma_{k}\vartheta_{1}} 
	\geq \frac{-\vartheta_{2}^{2}-\vartheta_{2}+\vartheta_{1}-1}{\vartheta_{1}}
\end{aligned}
\end{equation}
when assigning $\{\mathcal{F}, \mathcal{L}_\mathcal{F}\}$ to $\{\Psi, \mathcal{L}_\Psi\}$. So the condition
\begin{equation}
\small
	-\vartheta_{2}^{2}-\vartheta_{2}+\vartheta_{1}-1 > 0, 
\end{equation}
satisfies the prerequisite that $C_k > 0, \vartheta_1 \geq 1, \vartheta_2 > 0$. By the substitution from \eqref{ieq:Ck1}  into \eqref{eq:converge01}, we gain 
\begin{equation}
\small
	\min_{k=1\dots N_{ag}}\|\bar{\nabla}\mathcal{F}^{md}_{k}\|^2 \
	\leq \frac{\vartheta_{1}^{2} \mathcal{L}_{\mathcal{F}}\left(\mathcal{F}_{0} - \mathcal{F}^*\right)}{N_{ag}\left(-\vartheta_{2}^{2}-\vartheta_{2}+\vartheta_{1}-1\right)}.  
\end{equation}
%

\section{Derivation of Adaptive Stochastic\\ Trajectory Optimization}\label{appdx:paraASTO}
According to the definition of the reward $\mathcal{R}$ in \eqref{eq:reward}, we gain
\begin{equation}\label{eq:R_cost}
\small
\begin{aligned}
	& \mathcal{R}(\bm{\mu}',\bm{\mathcal{K}}' | {\bm{\mu}},{\bm{\mathcal{K}}})
	= \mathds{E}_{{\mathfrak{F}}_{\text{M}^{*}}\mid\bm{\Theta}_{\text{M}^*};\bm{\mu},\bm{\mathcal{K}}}
	\log L (\bm{\mu}',\bm{\mathcal{K}}'; \bm{\theta}_{m},\mathcal{F}_{m}) \\
	& = \sum_{m=1}^{M^{*}} {p_m} \left[\log p_{m} - \frac{d\log2\pi + \log |\bm{\mathcal{K}}'| + \left\|\bm{\theta}_{m} - \bm{\mu}' \right\|_ {\bm{\mathcal{K}}'}^2}{2} \right]
\end{aligned}
\end{equation}
Then we could find the optimal policy via
\begin{equation}
\small
	\hat{\bm{\mu}}, \hat{\bm{\mathcal{K}}} = \argmin_{\bm{\mu}', \bm{\mathcal{K}}'} \mathcal{R} (\bm{\mu}',\bm{\mathcal{K}}' | {\bm{\mu}},{\bm{\mathcal{K}}})
\end{equation}
informed by $\bm{\Theta}_{M^*}$, which could be simplified as
\begin{equation}
\small
	\begin{aligned}
		\hat{\bm{\mu}} 
		= \argmin_{\bm{\mu}'} \sum_{m=1}^{M^{*}} \frac{p_{m}}{2} \left\| \bm{\theta}_{m} - \bm{\mu}' \right\|_{\bm{\mathcal{K}}'}^2 \
		\Leftrightarrow \sum_{m=1}^{M^{*}} p_{m}\left(\bm{\theta}_{m} - \bm{\mu}' \right) = 0,
	\end{aligned}
\end{equation}
\begin{equation}
\small
\begin{aligned}
	\hat{\bm{\mathcal{K}}} 
	&= \argmin_{\bm{\mathcal{K}}'} \sum_{k=1}^{M^{*}} \frac{p_m}{2} \left(\log |\bm{\mathcal{K}}'| + \left\| \bm{\theta}_{m} - \hat{\bm{\mu}}\right\|_{\bm{\mathcal{K}}'}^2 \right) \\ 
	&\Leftrightarrow  \sum_{m=1}^{M^{*}} p_{m} \left[ \bm{\mathcal{K}}'^{-1} 
	-  \bm{\mathcal{K}}'^{-1} \left(\bm{\theta}_{m} - \bm{\mu}' \right) ^{\otimes 2} \bm{\mathcal{K}}'^{-1} \right]  = 0,  
\end{aligned}
\end{equation}
In this way, we gain the optimal policy
\begin{equation}
\small
	\hat{\bm{\mu}} =  \frac{\sum_{m=1}^{M^{*}} p_m \bm{\theta}_m}{ \sum_{m=1}^{M^{*}} p_m},\
	 \hat{\bm{\mathcal{K}}} = \frac{\sum_{m=1}^{M^{*}} p_{m} \left(\bm{\theta}_{m}-\hat{\bm{\mu}}\right)^{\otimes 2}}{\sum_{m=1}^{M^{*}} p_{m}}. 
\end{equation}
%

\section{Derivation of Accelerated Moving Averaging}\label{appdx:AMA}
According to Algorithm~\ref{alg:ASTO}, we gain
\begin{equation}\label{eq:AMA_mu0}
\small
\begin{aligned}
	&\bm{\mu}^\textit{md}_{n+1} = (1 - \alpha_n^\mu)(\bm{\mu}^\textit{md}_n + \beta_n^\mu \nabla \hat{\mathcal{R}}^\textit{md}_n) + \alpha_n^\mu (\bm{\mu}_{n-1} + \lambda_n^\mu \nabla \hat{\mathcal{R}}^\textit{md}_n) \\
	&= \bm{\mu}^\textit{md}_n + [\beta_n^\mu + \alpha_n^\mu (\lambda_n^\mu - \beta_n^\mu)] (\hat{\bm{\mu}}_n -  \bm{\mu}^\textit{md}_n) + \alpha_n^\mu(\bm{\mu}_{n\text{-}1} - \bm{\mu}^\textit{md}_n) 
\end{aligned}
\end{equation}
\begin{equation}\label{eq:AMA_covar0}
\small
	\bm{\mathcal{K}}^\textit{md}_{n+1} =  \bm{\mathcal{K}}^\textit{md}_n + [\beta_n^\kappa + \alpha_n^\kappa (\lambda_n^\kappa - \beta_n^\kappa)] (\hat{\bm{\mathcal{K}}}_n -  \bm{\mathcal{K}}^\textit{md}_n) + \alpha_n^\kappa(\bm{\mathcal{K}}_{n\text{-}1} - \bm{\mathcal{K}}^\textit{md}_n)
\end{equation}
Since $\bm{\mu}_{n} = \bm{\mu}_{n-1} + \lambda_n^\mu (\hat{\bm{\mu}}_n - \bm{\mu}^\textit{md}_n)$ and $\bm{\mu}_{0}= \bm{\mu}^\textit{md}_{1}$, we get
\begin{equation}\label{eq:mu_remain}
\small
\begin{aligned}
	\bm{\mu}_{n{-}1} - \bm{\mu}^\textit{md}_n  &= (1-\alpha_{n-1}^\mu)\left[\bm{\mu}_{n{-}2} - \bm{\mu}^\textit{md}_{n-1}\right.\\
	&\quad \left. +  (\lambda_{n-1}^\mu - \beta_{n-1}^\mu)(\hat{\bm{\mu}}_{n-1} - \bm{\mu}^\textit{md}_{n-1}) \right] \\
	& = \sum_{i=1}^{n-1} \prod_{j=i}^{n-1} (1-\alpha_j^\mu)(\lambda_i^\mu - \beta_i^\mu)(\hat{\bm{\mu}}_{i} - \bm{\mu}^\textit{md}_{i}), \\ 
\end{aligned}
\end{equation}
\begin{equation}\label{eq:covar_remain}
\small
\begin{aligned}
	\bm{\mathcal{K}}_{n{-}1} - \bm{\mathcal{K}}^\textit{md}_n  
	& = \sum_{i=1}^{n-1} \prod_{j=i}^{n-1} (1-\alpha_j^\kappa)(\lambda_i^\kappa - \beta_i^\kappa)(\hat{\bm{\mathcal{K}}}_{i} - \bm{\mathcal{K}}^\textit{md}_{i}), \\ 
\end{aligned}
\end{equation}
then substitute \eqref{eq:mu_remain} into \eqref{eq:AMA_mu0} and substitute \eqref{eq:covar_remain} into \eqref{eq:AMA_covar0} to gain the AMA method: 
\begin{equation}\label{eq:AMA}
\small
\begin{aligned} 
	\bm{\mu}^{\textit{md}}_{n+1} &= \bm{\mu}_n^\textit{md} + \alpha_n(\lambda_n-\beta) (\hat{\bm{\mu}}_n - \bm{\mu}_n^\textit{md}) \\[-4pt]
	&\quad + \alpha_{n-1}\Gamma_{n-1} \sum_{i=1}^{n-1} \Gamma_{i-1}^{-1} (\lambda_i-\beta)(\hat{\bm{\mu}}_i - \bm{\mu}_i^\textit{md}), \\
	\bm{\mathcal{K}}^{\textit{md}}_{n+1} &= \bm{\mathcal{K}}_n^\textit{md} + \alpha_n(\lambda_n-\beta) (\hat{\bm{\mathcal{K}}}_n - \bm{\mathcal{K}}_n^\textit{md}) \\[-4pt]
	&\quad + \alpha_{n-1}\Gamma_{n-1} \sum_{i=1}^{n-1} \Gamma_{i-1}^{-1} (\lambda_i-\beta)(\hat{\bm{\mathcal{K}}}_i - \bm{\mathcal{K}}_i^\textit{md}). 
\end{aligned}
\end{equation}

So when we set $\beta^\mu_n = (1-\alpha^{\mu})^n, \beta^\kappa_n = (1-\alpha^{\kappa})^n$ and $\lambda^\mu_n = 1+\beta^\mu_n, \lambda^\kappa_n = 1+\beta^\kappa_n$, we gain the qAdam method: 
\begin{equation}
\small
\begin{aligned}
	\bm{\mu}^\textit{md}_{n+1} &= \bm{\mu}^\textit{md}_n + [(1-\alpha^{\mu})^n + \alpha^{\mu}] (\hat{\bm{\mu}}_n -  \bm{\mu}^\textit{md}_n) \\
	&\quad + \alpha^\mu\sum_{i=1}^{n-1} \prod_{j=i}^{n-1} (1-\alpha^\mu)(\hat{\bm{\mu}}_{i} - \bm{\mu}^\textit{md}_{i}) \\
	&= \hat{\bm{\mu}}_n - \alpha^{\mu}\frac{(1-\alpha^{\mu})(1-(1-\alpha^{\mu})^{n-1})}{{1-(1-\alpha^{\mu})}} (\hat{\bm{\mu}}_n -  \bm{\mu}^\textit{md}_n) \\
	&\quad + \alpha^\mu\sum_{i=1}^{n-1} \prod_{j=i}^{n-1} (1-\alpha^\mu)(\hat{\bm{\mu}}_{i} - \bm{\mu}^\textit{md}_{i}) \\
	&= {\hat{\bm{\mu}}_n} + \alpha^{\mu}\sum_{i=1}^{n-1} \frac{(\hat{\bm{\mu}}_i - \bm{\mu}^{\textit{md}}_i) - (\hat{\bm{\mu}}_n - \bm{\mu}^{\textit{md}}_n)}{(1-\alpha^{\mu})^{i-n}},
\end{aligned}
\end{equation}
\begin{equation}
\small
\begin{aligned}
	\bm{\mathcal{K}}^\textit{md}_{n+1} 
	= {\hat{\bm{\mathcal{K}}}_n} + \alpha^{\kappa}\sum_{i=1}^{n-1} \frac{(\hat{\bm{\mathcal{K}}}_i - \bm{\mathcal{K}}^{\textit{md}}_i) - (\hat{\bm{\mathcal{K}}}_n - \bm{\mathcal{K}}^{\textit{md}}_n)}{(1-\alpha^{\kappa})^{i-n}}. 
\end{aligned}
\end{equation}
%

\section*{Acknowledgment}

The work of this paper is partially supported by the National Natural Science Foundation of China (No. 52175032), the Key R\&D Program of Zhejiang Province (2020C01025, 2020C01026), and Robotics Institute of Zhejiang University Grant (K11808). 

\ifCLASSOPTIONcaptionsoff
  \newpage
\fi



%

\bibliographystyle{IEEEtran}
\bibliography{AGP-STO_bib}

%

\end{document}